\documentclass{article} %
\usepackage{iclr2027_conference,times}

\author{Shayekh Bin Islam\\KAIST\\shayekh.islam@kaist.ac.kr \And
Hwanjun Song\thanks{Corresponding author.}\\KAIST\\songhwanjun@kaist.ac.kr
}

\iclrfinalcopy

\usepackage{amsmath,amsfonts,bm}

\def\eqref#1{equation~\ref{#1}}

\def\1{\bm{1}}

\DeclareMathAlphabet{\mathsfit}{\encodingdefault}{\sfdefault}{m}{sl}
\SetMathAlphabet{\mathsfit}{bold}{\encodingdefault}{\sfdefault}{bx}{n}

\usepackage{xspace}

\newcommand{\bench}{\textsc{PlaylistEval}\xspace}
\newcommand{\dataset}{\textsc{PlaylistBench}\xspace}

\newcommand{\gemininame}[1]{Gemini-#1}               %
\newcommand{\gflash}{\gemininame{3.7-Flash}\xspace}  %
\newcommand{\gpro}{\gemininame{3.1-Pro}\xspace}      %
\newcommand{\glite}{\gemininame{3.5-Flash-Lite}\xspace}  %

\newcommand{\gptt}{GPT-5.6-Terra\xspace}             %
\newcommand{\kimi}{Kimi-K2.6\xspace}                 %
\newcommand{\qmax}{Qwen-3.8-Max\xspace}              %
\newcommand{\qflash}{Qwen-3.7-Flash\xspace}          %

\newcommand{\gemmaname}[1]{Gemma-4-#1}               %
\newcommand{\gemmamoe}{\gemmaname{26B-A4B}\xspace}   %
\newcommand{\gemmaefour}{\gemmaname{E4B}\xspace}     %
\newcommand{\gemmaetwo}{\gemmaname{E2B}\xspace}      %

\newcommand{\qwenjname}[1]{Qwen-3.5-#1}              %
\newcommand{\qwennine}{\qwenjname{9B}\xspace}        %
\newcommand{\qwenfour}{\qwenjname{4B}\xspace}        %
\newcommand{\qwentwo}{\qwenjname{2B}\xspace}         %

\newcommand{\qomni}{Qwen3-Omni-30B-A3B\xspace}       %
\newcommand{\ixc}{InternLM-XComposer-2.5-Reward\xspace}  %

\newcommand{\cmuname}[1]{VideoJudge-#1}              %
\newcommand{\cmuseven}{\cmuname{7B}\xspace}          %
\newcommand{\cmuthree}{\cmuname{3B}\xspace}          %

\newcommand{\qwembname}[1]{Qwen3-VL-Embedding-#1}    %
\newcommand{\qwembeight}{\qwembname{8B}\xspace}      %
\newcommand{\qwembtwo}{\qwembname{2B}\xspace}
\newcommand{\wemm}{WeMM-Embedding-9B\xspace}         %
\newcommand{\nemo}{Omni-Embed-Nemotron-3B\xspace}    %

\newcommand{\gflashgen}{Gemini-3-Flash\xspace}       %
\newcommand{\gptfull}{GPT-5.4\xspace}                %
\newcommand{\gptmini}{GPT-5.4-mini\xspace}           %
\newcommand{\qplus}{Qwen-3.7-Plus\xspace}            %

\newcommand{\qwenvlmoe}{Qwen3-VL-30B-A3B\xspace}     %
\newcommand{\internvleight}{InternVL3.5-8B\xspace}  %
\newcommand{\asr}{Qwen3-ASR-1.7B\xspace}             %

\newcommand{\hf}[1]{\href{https://huggingface.co/#1}{\texttt{#1}}}

\newcommand{\shayekh}[1]{\textcolor{blue}{Shayekh: #1}}

\usepackage{hyperref}
\usepackage{url}

\usepackage{enumitem}
\usepackage{comment}
\usepackage{booktabs}
\usepackage{multirow}
\usepackage{graphicx}   %
\usepackage[normalem]{ulem}  %
\usepackage{tikz}

\makeatletter
\renewcommand\dashuline{\leavevmode\bgroup
  \UL@setULdepth
  \ifx\UL@on\UL@onin \advance\ULdepth2\p@\fi
  \let\ULleaders\cleaders
  \markoverwith{\kern.09em\vtop{\kern\ULdepth\hrule width .32em}\kern.09em}%
  \UL@pixel\z@
  \ULon}
\makeatother
\DeclareRobustCommand{\ico}[1]{\raisebox{-0.22ex}{\includegraphics[height=1.55ex]{images/icons/#1.pdf}}}
\DeclareRobustCommand{\icoV}{\ico{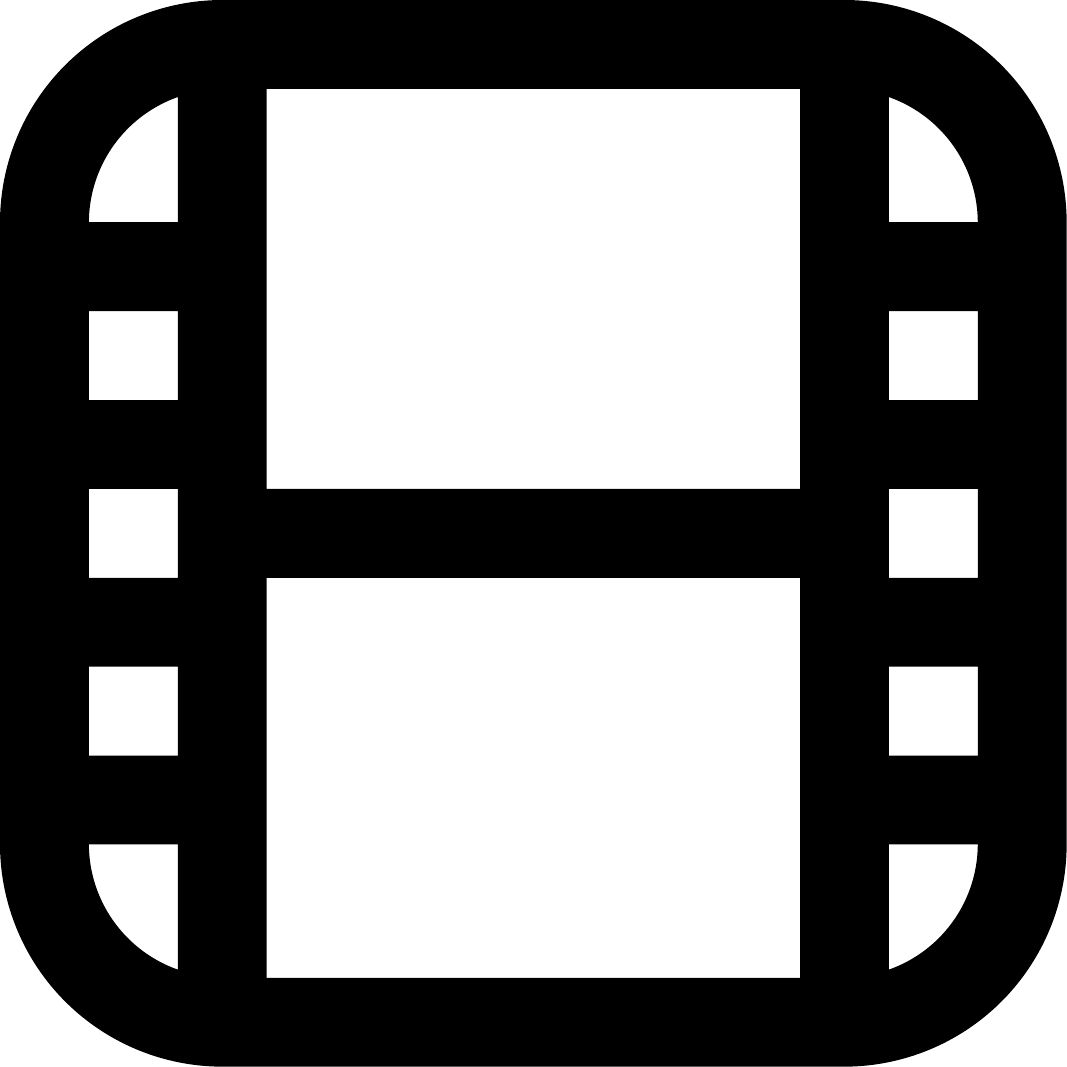}}
\DeclareRobustCommand{\icoT}{\ico{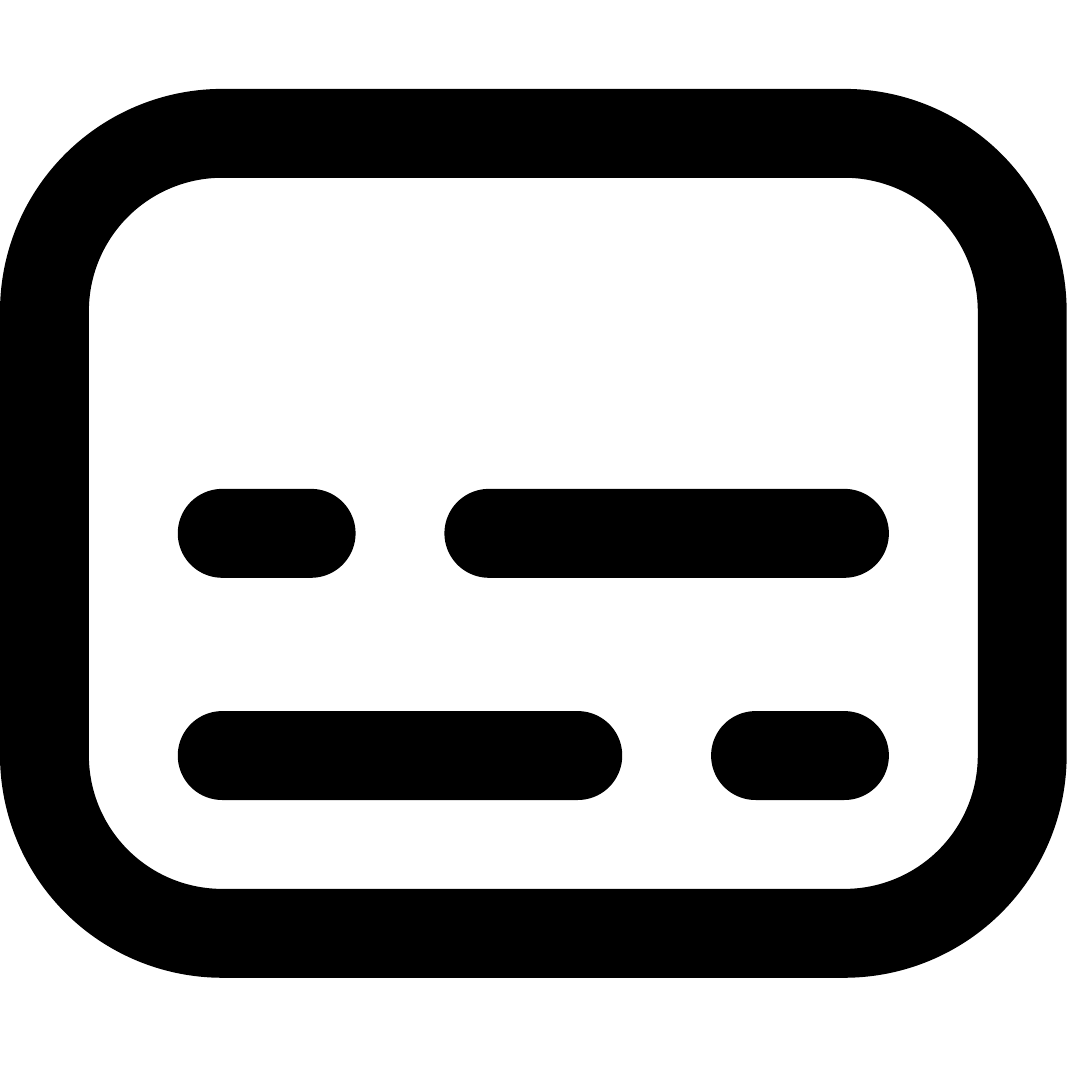}}
\DeclareRobustCommand{\icoA}{\ico{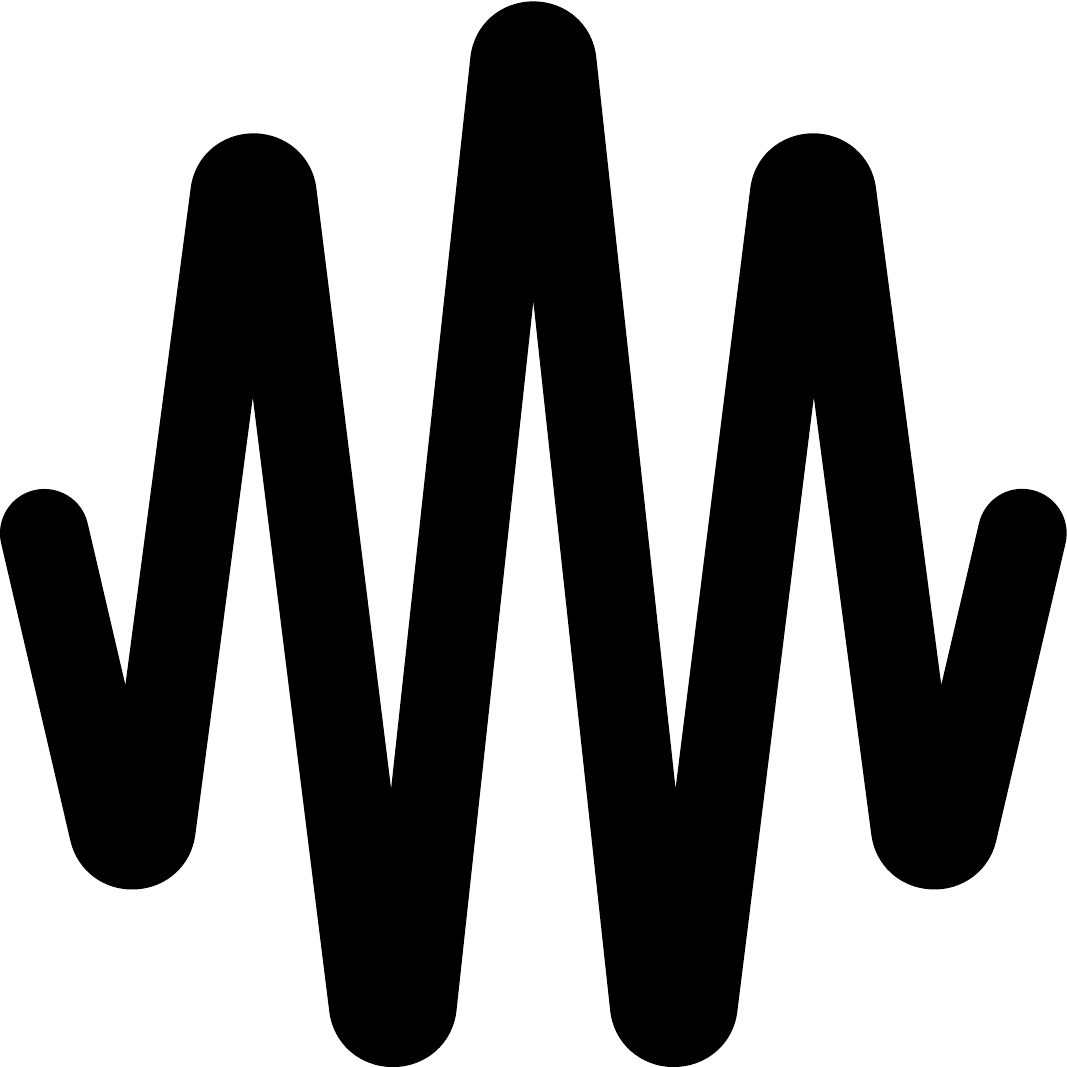}}

\DeclareRobustCommand{\ilogo}[1]{\makebox[2.3ex][l]{\raisebox{-0.3ex}{\includegraphics[height=1.75ex,width=2.1ex,keepaspectratio]{images/icons/logo_#1.pdf}}}}
\DeclareRobustCommand{\nologo}{\makebox[2.3ex][l]{}}
\DeclareRobustCommand{\rlogo}[1]{\makebox[2.6ex][l]{\raisebox{-0.3ex}{\includegraphics[height=1.75ex,width=2.3ex,keepaspectratio]{images/icons/logo_#1.pdf}}}}

\usepackage[most]{tcolorbox}
\usepackage{listings}
\usepackage{array}      %

\definecolor{promptink}{HTML}{1A1A19}
\definecolor{promptmuted}{HTML}{6B6A66}
\definecolor{promptrule}{HTML}{B9B8B2}
\definecolor{promptback}{HTML}{F8F7F4}
\definecolor{prompthead}{HTML}{3D3C39}
\definecolor{promptslot}{HTML}{1F63B5}

\newtcolorbox{promptbox}[1]{%
  enhanced, breakable,
  colback=promptback, colframe=promptrule, boxrule=0.5pt, arc=1.5pt,
  left=7pt, right=7pt, top=5pt, bottom=6pt,
  fontupper=\small, coltext=promptink,
  title={#1}, fonttitle=\small\bfseries, coltitle=white, colbacktitle=prompthead,
  toptitle=2pt, bottomtitle=2pt, lefttitle=7pt,
  before skip=8pt, after skip=12pt,
  before upper={\setlength{\parindent}{0pt}\setlength{\parskip}{3pt plus 1pt}},
}

\newtcolorbox{schemabox}[1]{%
  enhanced, breakable,
  colback=promptback, colframe=promptrule, boxrule=0.5pt, arc=1.5pt,
  left=7pt, right=7pt, top=5pt, bottom=6pt,
  fontupper=\small, coltext=promptink,
  title={#1}, fonttitle=\small\bfseries, coltitle=promptink, colbacktitle=promptrule!45!white,
  toptitle=2pt, bottomtitle=2pt, lefttitle=7pt,
  before skip=8pt, after skip=12pt,
  before upper={\setlength{\parindent}{0pt}\setlength{\parskip}{3pt plus 1pt}},
}

\newcommand{\pmeta}[1]{{\footnotesize\color{promptmuted}#1\par}\vspace{3pt}}
\newcommand{\psec}[1]{\par\vspace{5pt}{\bfseries #1}\par\vspace{1pt}}
\newcommand{\pslot}[1]{\textcolor{promptslot}{\texttt{\{\{#1\}\}}}}
\newcommand{\pfield}[1]{{\ttfamily\bfseries #1}\enspace}
\newcommand{\pdivider}[1]{\par\vspace{6pt}%
  {\footnotesize\color{promptmuted}\rule[0.55ex]{1.2em}{0.4pt}\enspace\textit{#1}\enspace
   \leaders\hrule height 0.6ex depth -0.5ex\hfill\kern0pt}\par\vspace{2pt}}

\newtcolorbox{pcond}[1]{%
  blanker, borderline west={0.6pt}{0pt}{promptrule},
  left=6pt, top=1pt, bottom=1pt, before skip=5pt, after skip=3pt,
  fontupper=\small, coltext=promptink,
  before upper={\setlength{\parindent}{0pt}\setlength{\parskip}{3pt plus 1pt}%
    {\footnotesize\color{promptmuted}\textit{#1}\par}},
}

\newenvironment{plist}{\begin{list}{\textbullet}{%
    \setlength{\leftmargin}{1.2em}\setlength{\labelwidth}{0.8em}\setlength{\labelsep}{0.4em}%
    \setlength{\itemsep}{1pt}\setlength{\parsep}{0pt}%
    \setlength{\topsep}{1pt}\setlength{\partopsep}{0pt}}}%
  {\end{list}}

\newcounter{pnumc}
\newenvironment{pnum}{\begin{list}{\arabic{pnumc}.}{\usecounter{pnumc}%
    \setlength{\leftmargin}{1.4em}\setlength{\labelwidth}{1em}\setlength{\labelsep}{0.4em}%
    \setlength{\itemsep}{1pt}\setlength{\parsep}{0pt}%
    \setlength{\topsep}{1pt}\setlength{\partopsep}{0pt}}}%
  {\end{list}}

\lstnewenvironment{promptcode}{\lstset{%
    basicstyle=\ttfamily\scriptsize, columns=fullflexible, keepspaces=true,
    breaklines=true, breakatwhitespace=false, showstringspaces=false,
    aboveskip=3pt, belowskip=3pt, xleftmargin=0.5em,
    postbreak=\mbox{\textcolor{promptmuted}{$\hookrightarrow$}\space}}}{}
\usepackage{float}      %
\usepackage{wrapfig}    %
\usepackage{needspace}  %

\title{

\bench{}: Can Video-Language Judges Be Trusted at Day Scale and Beyond?
}

\begin{document}

\maketitle

\addtocontents{toc}{\protect\setcounter{tocdepth}{-1}}

\begin{abstract}
Video-language models are increasingly used as judges of video understanding, both for evaluating model outputs and for training reward models. Whether their judgments remain reliable when the evidence is buried in day-long videos has yet to be established. Existing benchmarks cannot answer this. Their videos are typically only a few minutes long, many answer pairs can be separated from the transcript alone, and collecting human judgments does not scale to ultra-long videos. We introduce~\bench{}, an agentic framework that builds video-language judge benchmarks over $100$-hour playlist collection without human annotation. It automatically generates questions with paired answers whose differences are controlled by causal degradation, so that every pair demands retrieval across the collection. The resulting benchmark contains $630$ pairs across seven domains spanning both static and dynamic knowledge, and on a stratified subset of $152$ pairs it agrees with human judgments $93.0\%$ of the time (IAA $0.781$). Evaluating $17$ omnimodal and multimodal models from eight families reveals that frontier judges reach only $75.4\%$ pairwise accuracy, while open-source judge models perform far behind. We further show that both retrieval and final judgment depend on using multiple modalities, and that judge accuracy degrades as the playlist set grows. We release our pipeline, benchmark, and evaluation code at \href{https://playlisteval.github.io}{playlisteval.github.io}.
\end{abstract}

\section{Introduction}

Video-language judges, which score candidate responses against video evidence, now underpin both the evaluation and the training of multimodal systems \citep{zhang2025videorewardbench, waheed2026videojudge, hu2026multimodal}, and nowhere more so than for long video. On the evaluation side, long-video question answering is moving from multiple choice to long-form answers grounded in hours of video that no single reference can grade \citep{fang2024mmbench, luo2025videoautoarena}. On the training side, video MLLMs are increasingly optimized against judge-provided rewards \citep{waheed2026videojudge, wei2026video}. Since human feedback does not scale to long video, the judge is the only practical source in both roles, and a misjudged answer becomes a misranked system or a misguided update. Yet judges have rarely been tested on video beyond an hour \citep{zhang2025videorewardbench, waheed2026videojudge, wei2026video}, so we do not know whether that trust survives once the video grows to days.

Three problems in how existing judge benchmarks are built keep it that way. %
The first is \emph{length}. The videos are short. Most run under a minute and few approach an hour \citep{waheed2026videojudge, zhang2025videorewardbench, wei2026video}, so the judge sees the whole clip at once and never has to decide where to look. 
The second is \emph{grounding}. %
Answer pairs come from human labels, ground-truth answers, or text descriptions of the video, and long-video systems are often graded by text-only judges. A judge that never looks at a frame can therefore still score well \citep{wei2026video, waheed2026videojudge, ren2026videorag}.
The third is \emph{scalability}. Annotating hours of video for fine visual detail is the main cost of building long-video sets \citep{wang2025lvbench, hu2026multimodal}, which keeps existing benchmarks narrow in domain, fixed in difficulty, and impossible to rebuild over new collections.

\begin{figure}[!t]
\centering
\includegraphics[width=\textwidth]{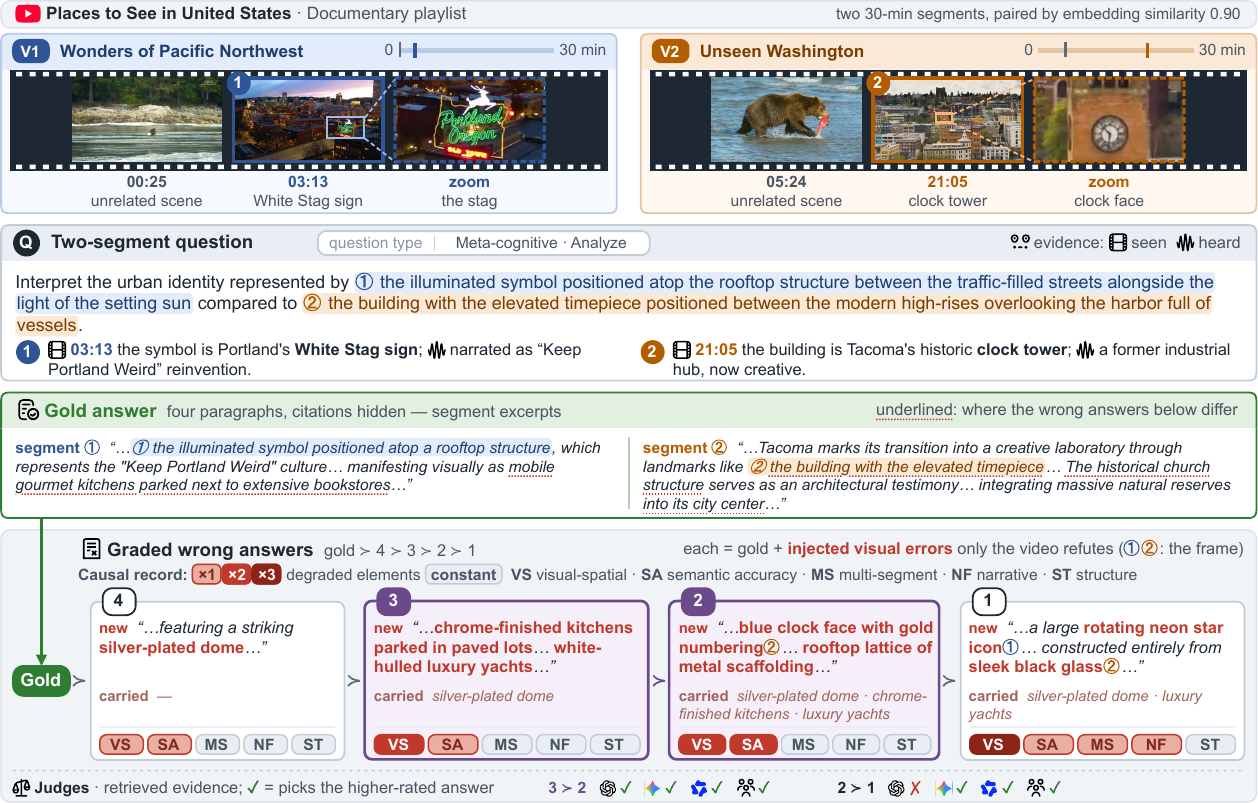}
\vspace*{-0.8cm}
\caption{\textbf{An item in \bench{}}: Two 30-minute segments of a 100-hour Documentary playlist, paired by embedding similarity, each supply one of the question's two entities. The question names neither entity and identifies each only by its surroundings, so its evidence must be found across the collection and confirmed by what is seen and heard. From the gold answer, four wrong answers are derived by injecting visual errors of graded severity (rating 4 to 1), invisible in the transcript. Any two answers form a pair, so the rating gap sets how hard each pair is.
}
\vspace*{-0.35cm}
\label{fig:example}
\end{figure}

To bridge these gaps, we introduce \bench{}, an agentic benchmark curator that tests whether a video-language judge can be trusted over a \emph{multi-day playlist}, a set of related videos totaling about 100 hours in each of seven domains (see an example in Figure \ref{fig:example}). It builds such preferences fully automatically from native video collections, matching human judgments without human annotation. Its two stages and the feedback loop around them address the three problems above in turn.

As seen in Figure \ref{fig:pipeline}, Phase I targets length. It indexes the playlist and generates a question whose evidence lies in two distant moments of it, with a gold answer that cites its supporting spans, so the judge faces a \emph{needle-in-a-haystack} search rather than a clip it can watch in full.
Phase II targets grounding. It derives four incorrect answers from the gold by injecting visual errors of varying severity, each \emph{indistinguishable} from the gold on the transcript alone, so every incorrect answer differs in what is seen rather than in what is said.
The feedback loop targets scalability. 
Validity gates at both phases return their rejection reasons to the generator, making the pipeline \emph{self-correcting}; at roughly \$1 per question, it retries until all gates are passed or the budget is exhausted, and can be rerun on any new playlist.
Applied to playlists from seven domains, this pipeline yields~\dataset{}, a benchmark of $630$ preference pairs, each pairing two answers of different error severity with the milder one as the intended preference. Human annotators confirmed these preferences in $93.0\%$ of sampled cases, with an inter-annotator agreement of $0.781$.

These design choices make \bench{} a \emph{controlled} testbed. 
\emph{(i) Playlist size}: because the gold answer is grounded in two verified spans, we can vary the playlist length without losing the answer and check whether retrieval surfaces those spans. \emph{(ii) Modality}: wrong answers are indistinguishable from the gold on the transcript alone, so we can measure the contributions of frames, transcript, and audio to judging and retrieval. \emph{(iii) Difficulty}: graded wrong answers let the rating gap control pair difficulty, from obvious errors to single-detail changes. With these controls, we evaluate $17$ general-purpose and judge-tuned models from eight families, and pair them with four retrievers to test whether retrieval helps on long playlists.
These controls uncover systematic failures across length, modality, retrieval, and answer order, with several becoming visible only at day scale and beyond, which prior benchmarks do not cover.

\vspace*{-0.2cm}
\begin{itemize}[leftmargin=*]
\item The best judge reaches $75.4\%$ against $93.0\%$ for humans, small models sit near chance, and judges fine-tuned on short video fall to or below chance.
\vspace*{-0.05cm}
\item Judge accuracy drops steadily from 1 hour to 100 hours, retrieval recovers up to $+10.5$ points, 
but the best retriever finds the right video segments only $37.9\%$ of the time. 
\vspace*{-0.05cm}
\item Frames or transcript alone costs $2$--$7$ points against using both, so neither suffices. More thinking or higher resolution barely helps, so the bottleneck is finding the evidence, not seeing it. 
\item When the two answers swap sides, weaker judges (\emph{e.g.}, \gemmamoe{} and \glite{}) reverse their verdict on roughly half of pairs, whereas stronger judges (\emph{e.g.}, \qmax{} and \gflash{}) stay largely consistent. 
\end{itemize}
\vspace*{-0.15cm}

\vspace*{-0.1cm}
\section{Related Work}
\vspace*{-0.1cm}

\textbf{Multimodal Judge Models.~~}
Using a strong model to score another model's output, as a reward model or an {LLM-as-a-judge}, has become the standard scalable proxy for human preference, since its introduction in RLHF \citep{christiano2017deep, ouyang2022training}, and this paradigm has since moved into the multimodal setting. 
On images, LLaVA-Critic \citep{xiong2025llavacritic} is trained as a generalist evaluator for both pointwise scoring and pairwise ranking, while InternLM-XComposer-2.5-Reward \citep{zang2025internlm}, Skywork-VL Reward \citep{wang2025skyworkvlrewardeffectivereward}, and MM-RLHF \citep{zhang2025mm} learn multimodal reward models to align vision--language models to human preference, with recent work hardening such rewards against spurious cues \citep{srivastava2026robust}. 
Extending judges to video is far less explored: VideoJudge \citep{waheed2026videojudge} bootstraps an MLLM-as-a-judge, and \citet{wei2026video} train dedicated video reward models. 
These judges, however, are developed and validated on clips of mostly a few minutes, leaving open whether they can supervise reasoning over ultra-long, multi-segment video, the regime \bench{} targets.

\textbf{Judge Model Evaluation.~~}
The reliability of a judge is itself measured by dedicated benchmarks, which pair each prompt with a preferred and a dispreferred response and report how often the judge agrees with human preference.
In the text-only setting, RewardBench \citep{lambert2025rewardbench} and its harder successor RewardBench 2 \citep{malik2025rewardbench} evaluate reward models across chat, reasoning, and safety, while JudgeBench \citep{tan2025judgebench} stress-tests LLM-as-a-judge on response pairs whose correctness is objectively verifiable.
For image--text inputs, VL-RewardBench \citep{li2025vlrewardbench} and Multimodal RewardBench \citep{yasunaga2025multimodal} extend this evaluation to vision--language judges, and Multimodal RewardBench 2 \citep{hu2026multimodal} broadens it to interleaved understanding and generation.
Video-language judges are assessed by VideoJudge \citep{waheed2026videojudge}, VideoRewardBench \citep{zhang2025videorewardbench}, and VURB \citep{wei2026video}. These video benchmarks, however, span clips of only a few minutes and rely on costly human annotation, which sharply limits their reach to ultra-long scenarios. In contrast, \bench{} is the first fully automated, native-video judge benchmark, built by a scalable pipeline while retaining high accuracy.

\vspace*{-0.1cm}
\section{PlaylistEval: Playlists to Judge Benchmarks}
\label{sec:playlisteval}
\vspace*{-0.1cm}

\bench{} takes a playlist collection and returns preference pairs for judge evaluation without any human annotation (see Figure \ref{fig:pipeline}). It is designed to produce pairs that are \emph{long} and \emph{video-grounded}, and, by keeping human involvement to a minimum, to \emph{scale} to any new playlist collection. The first two properties are enforced by two generation phases, one generating QA over evidence scattered across the playlist and the other generating distractors beyond what the transcript reveals, and the third by a self-correcting feedback loop that wraps around both and selects the final pairs.

\textbf{Manual Playlist Collection.~~}
Selecting playlists is the only step of \bench{} that involves a human. Following existing video benchmarks \citep{wu2024longvideobench, wang2025lvbench}, we chose seven domains---Education, Drama, Life, Art, History, Documentary, and Podcasts---that cover static factual knowledge, dynamic narrative content, and mixtures of both (Appendix~\ref{app:domains}). 
For each domain we searched YouTube playlists under varied filters to gather an initial pool of 100 playlists of diverse topics and lengths. From this pool, we curated the final set in three passes. We first discarded private or deleted video links from the playlists, then, where a playlist's order disagreed with its titles, re-sorted the videos by the episode keywords in those titles (\emph{e.g.}, Episode 4 before Episode 5), and finally trimmed or extended each domain until it covered about 100 hours. As a result, the curated collection spans $29$ playlists and $457$ videos with about 4 playlists per domain on average. 

\textbf{Automatic Indexing.~~} A $100$-hour playlist collection cannot be reliably processed as a whole by any current video-language model. Indexing it into smaller units is therefore indispensable, both for generating questions and for grounding every answer in the exact moments that support it, which is central to \bench{}. 
We split every video into $30$-second chunks, 
a length short enough to localize a single visual moment yet long enough to carry a complete utterance \citep{ren2026videorag}, transcribe each with Qwen3-ASR-1.7B \citep{shi2026qwen3}, 
and embed it into a single vector with Qwen3-VL-Embedding-8B \citep{li2026qwen3vlembedding} from its sampled frames and transcript together, so that both what is seen and said are represented. Chunk embeddings are further averaged over the $10$--$30$-minute segment, the granularity at which evidence is grounded.
Together, these chunk and segment embeddings form a searchable index over the playlist that every later stage builds on. Before it is used, we remove near-duplicate videos, since duplicated contents would make a question answerable from a second copy and repeat content across questions. We flag any pair of segments with similarity above $0.95$ and fill the gap with newer videos until each domain again covers $100$ hours. 

\newcommand{\circled}[1]{\textcircled{\scriptsize #1}}
\begin{figure}[t]
\centering
\includegraphics[width=\textwidth]{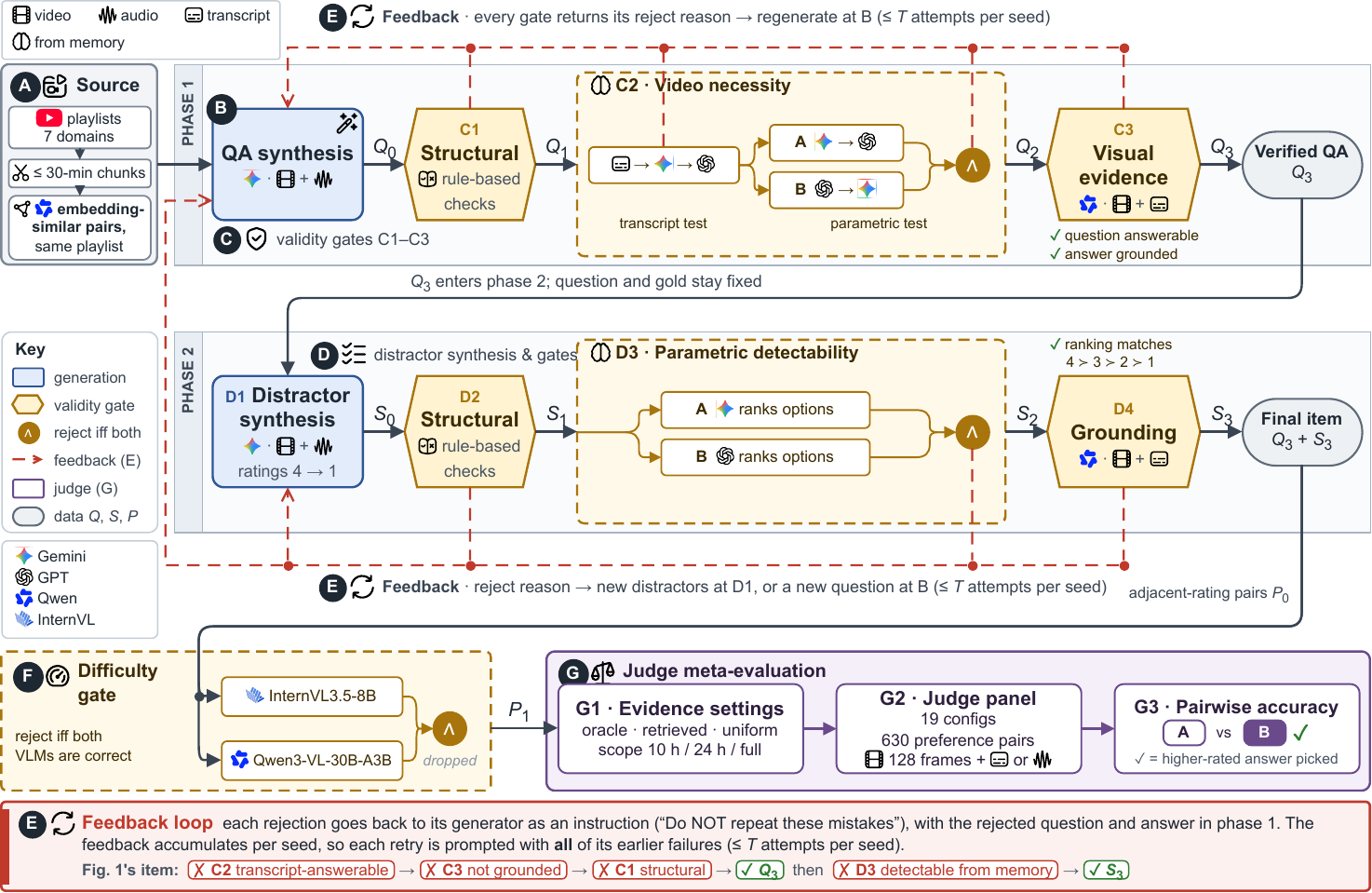}
\vspace*{-0.7cm}
\caption{
\textbf{Overview of \bench{}}: \circled{A} From indexed and paired playlist segments, \circled{B} {Phase I} generates a QA and \circled{C} verifies it through three gates, \circled{D} {Phase II} generates graded distractors and verifies them likewise, and \circled{E} the {feedback loop} returns every rejection to its generator. \circled{F} Surviving items pass a difficulty gate and \circled{G} form the preference pairs for judge meta-evaluation. 
}
\vspace*{-0.45cm}
\label{fig:pipeline}
\end{figure}

\vspace*{-0.1cm}
\subsection{Phase I: Generating QA over Scattered Evidence}
\label{sec:qa_gen}
\vspace*{-0.1cm}

Phase I turns a segment pair into a question and a gold answer that are grounded in both segments as evidence, and keeps only those that cannot be answered without watching the video.

\textbf{Evidence-Cited QA Generation.~~}
This step generates a question that cannot be answered from any single moment of the playlist. Its evidence is scattered across two distant segments of a $100$-hour playlist, so the judge must first find both and then combine them, and the gold answer cites exactly where each piece lies. In detail, each question is seeded by a pair of same-domain segments with embedding similarity in $[0.40, 0.90]$, close enough to share a question yet distinct enough to require both. \gflashgen{} receives both segments as native video ($0.5$ fps, $720$p) with audio narration, and produces a question, a gold answer, and verification metadata in one inference call.

The output is constrained so that neither the question nor the answer can be resolved without the video. The question, following \citet{wu2024longvideobench}, refers to entities only by their surroundings in the frame rather than by name, so that recognizing them requires locating the scene, and it targets the complex cells of Bloom's Knowledge Dimension Matrix \citep{ullrich2021using}, so that answering requires reasoning over both segments rather than recalling a fact. The gold answer is a $3$--$5$ paragraph response in which every claim cites its supporting span as (\texttt{video-id} @ \texttt{MM:SS}--\texttt{MM:SS}). These citations form the evidence map of the question, as exemplified in Figure~\ref{fig:example-full}.

\textbf{Validity Gates.~~}
The constraints above are imposed only at generation time, and prior work has shown that such instructions alone are insufficient \citep{nagrani2024neptune}. Synthesized multimodal QA is often answerable from the transcript or parametric knowledge alone \citep{mangalam2023egoschema, nagrani2024neptune}, and cited spans do not always support their claims. We therefore verify each QA through three sequential stages, where cheap structural and text-only checks screen out early failures before the costly video call, and the verifier never belongs to the generator's family to mitigate self-preference bias.
\vspace*{-0.15cm}
\begin{itemize}[leftmargin=*]
\item \emph{Structural Validity.} A rule-based check with no model call. The answer must contain at least $3$ paragraphs, with citations covering $2$--$15$ of the $30$-second chunks in each segment (\emph{i.e.}, $1.0$--$7.5$ minutes of evidence), ensuring that the evidence is neither insufficient nor overly diffuse. 
\vspace*{-0.1cm}
\item \emph{Video Necessity.} After the structural checks, we reject any question answerable without the video. In the transcript test, \gflashgen{} answers from transcripts alone, with the relevant transcript shuffled among windows from another video of the same playlist, and \gptmini{} rejects the question if the answer matches the gold. In the parametric test, the question is answered from model memory with no input, under two generator--verifier pairs (\gflashgen{} with \gptfull{}, and \gptmini{} with \gpro{}), and rejected only if both recover the gold. Generator and verifier always come from different families to mitigate self-preference bias.
\vspace*{-0.1cm}
\item \emph{Video Sufficiency.} Finally, we reject questions that the video itself cannot answer. \qplus{}, a third family, receives each segment's full video at $0.5$ fps with its transcript and verifies that the question is answerable from the video and that every claim in the gold answer is supported 
by its segment pairs.
As this model does not support native audio, the narration is fed as ASR transcript.
\end{itemize}
\vspace*{-0.15cm}
Questions that clear all three stages are fixed for Phase~II. Those that fail at any stage are sent back to the generator with the reason for rejection, which drives the feedback loop described in Sec. \ref{sec:final}. Appendix \ref{app:datagen-config} discusses each stage's model, inputs and acceptance rule; and Appendix \ref{app:data_prompt} the prompts.

\vspace*{-0.1cm}
\subsection{Phase II: Generating Distractors beyond the Transcript}
\label{sec:distractor}
\vspace*{-0.1cm}

Phase II turns a verified QA into a set of wrong answers that differ from the gold only in what is seen, so that a judge who reads the transcript but never watches the video cannot tell them apart.

\textbf{Graded Visual Degradation.}
This step generates four wrong answers from the gold with controlled severity, rated $4$ to $1$ with the gold as $5$, so that any two answers form a preference pair whose difficulty is set by their rating gap. \gflashgen{} receives the question's two evidence segments as native video with the fixed question and gold, and rewrites the gold by injecting visual-only errors (color, spatial layout, gesture, props, on-screen graphics) that a reader with only the transcript or world knowledge cannot detect, while preserving its length, tone, and structure. Following causal rubric prompting \citep{srivastava2026robust}, the model also records which question-specific attributes each answer degrades and through which elements, so that severity is an explicit, auditable quantity rather than an impression. This record fixes the intended order ``$\text{gold} \succ 4 \succ 3 \succ 2 \succ 1$." %

\textbf{Detectability Gates.} A degraded set is useful only if its errors are invisible in text yet visible in video, and neither property is guaranteed by the prompt. Each set therefore passes three gates, run in the same cheap-to-costly order and with the same cross-family generator--verifier assignment.
\vspace*{-0.15cm}
\begin{itemize}[leftmargin=*]
\item \emph{Structural Validity.} A rule-based check with no model call. Each degraded answer must carry a well-formed causal record with at least one visual-only degradation.
\vspace*{-0.1cm}
\item \emph{Textual Undetectability.} We reject any set whose ranking can be recovered without the video. \gpro{} and \gptfull{} each rank the gold and the four degraded answers, shuffled and identically formatted, from text alone with the question-specific attributes as rubric. The set is rejected only if both judges recover the intended order.
\vspace*{-0.1cm}
\item \emph{Visual Detectability.} Finally, we reject any set whose ranking cannot be recovered even with the video. \qplus{} receives each evidence segment's full video at $0.5$ fps with its transcript and ranks the four degraded answers, each accompanied by its causal record to verify against the frames. The set passes only if the judge reproduces the intended order exactly.
\end{itemize}
\vspace*{-0.1cm}

Sets that clear all gates form a final item with their question. Those that fail return to the distractor generator with the reason for rejection, driving the feedback loop in Section \ref{sec:final}.
Details including prompts, the model, the inputs and the acceptance rule of every stage are in Appendices~\ref{app:datagen-config} and~\ref{app:data_prompt}.

\vspace*{-0.1cm}
\subsection{Self-Correcting Loop and Pair Selection}
\label{sec:final}
\vspace*{-0.1cm}

The two phases above are not a fixed pipeline of filters but a closed loop, in which every rejection becomes an instruction for the next attempt. This is what lets \bench{} run end-to-end on a new playlist without a human deciding what to regenerate, and what keeps its cost bounded.

\begin{wrapfigure}{r}{0.40\textwidth}
\centering
\includegraphics[width=\linewidth]{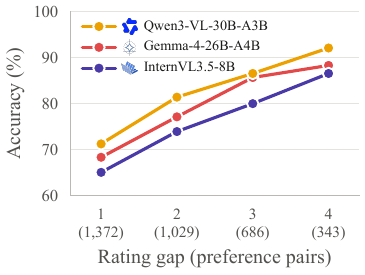}
\vspace*{-0.7cm}
\caption{
\textbf{Difficulty by rating gap.} 
}
\label{fig:rank-gap}
\vspace{-0.5\baselineskip}
\end{wrapfigure}

\textbf{Rejection as Feedback.}
Whenever a gate rejects an item, its reason and the rejected output are appended to the generator's prompt as an explicit instruction, so that the next attempt is conditioned on the exact failure. 
Feedback accumulates per seed, and a Phase~II initial rejection regenerates only the distractors up to two more times, keeping the verified question fixed so that the costly video checks of Phase~I are minimized. 
After three Phase~II rejections, the cycle restarts from Phase~I to generate a new QA with the same video segment pair, considering the previous failure history as feedback. 
To bound cost and avoid overfitting to the automated verifiers \citep{waheed2026videojudge}, we allow at most $T=6$ attempts per seed and phase, after which the segment pair is discarded.

\textbf{Controllable Pair Selection.}
Any two of the five answers of an item form a preference pair with the higher-rated answer as the intended preference, but pairs with a large rating gap are trivially easy. We thus add a difficulty gate independent of the models above. Two open-weight judges, \qwenvlmoe{} and \internvleight{}, evaluate every candidate pair, and only those that at least one of them fails are retained. Both are deliberately weaker than the judges evaluated in Sec. \ref{sec:exp}, so the gate removes pairs that even a modest judge solves without biasing the benchmark toward any evaluated model. As Figure \ref{fig:rank-gap} shows, their accuracy rises monotonically with the rating gap, confirming that our framework generates pairs of controllable difficulty.
Finally, sampling $90$ preference pairs per domain yields \textsc{PlaylistBench}, which contains $630$ pairs over $327$ unique questions and is dominated by rating gaps of $1$ and $2$. Appendix~\ref{app:data-stats} gives the statistics of the resulting benchmark.

\textbf{Cost and Fidelity.~~}
The pipeline is cheap because each gate decides whether an item proceeds, so free rule-based and cent-level text-only checks run first and only survivors reach the two native-video models that account for over $90\%$ of the bill (see Table~\ref{tab:app-cost-model}). Building the benchmark cost $352$ USD, or roughly $1$ USD per accepted question including all retries, and each question yields up to $10$ preference pairs from its five graded answers. This economy does not trade away fidelity. On a stratified subset of $152$ pairs, human annotators agreed with the intended preference in $93.0\%$ of cases with an inter-annotator agreement of $0.781$ (see Appendix~\ref{app:human-eval}).
The same comparison also shows why human annotation cannot scale to this setting. Verifying those $152$ pairs alone cost about $4.7$ USD
each, whereas our pipeline verified every pair at about $0.5$ USD 
each, roughly $\times8$ cheaper.

\vspace*{-0.15cm}
\section{Evaluation}
\label{sec:exp}
\vspace*{-0.15cm}
\begin{table}[t]
\centering
\small
\setlength{\tabcolsep}{3pt}
\setlength{\ULdepth}{1.2pt}
\caption{Judge accuracy (\%) per content domain and overall, with domain-retrieved frames versus uniform frame sampling. Each cell reports \emph{retrieved}\,/\,\emph{uniform} accuracy. 
\kimi{} is 1T-A32B and \qmax{} is 2.4T-A95B.
Per column, the best value is in \textbf{bold} and the second best has a \protect\dashuline{dashed underline}, separately for retrieved and uniform. Doc~=~Documentary, Educ~=~Education, Hist~=~History, Pod~=~Podcast. Input modalities: \icoV\ video frames, \icoT\ text transcript, \icoA\ audio track.}
\label{tab:e01-main}
\begin{tabular}{>{\raggedright\arraybackslash}p{100pt}l*{8}{c}r}
\toprule
 & & \multicolumn{7}{c}{Domain (retrieved\,/\,uniform)} & \multicolumn{2}{c}{Overall} \\
\cmidrule(lr){3-9}\cmidrule(lr){10-11}
Judge & Input & Educ & Drama & Life & Art & Hist & Doc & Pod & R\,/\,U & $\Delta$ \\
\midrule
\multicolumn{11}{c}{\emph{Hosted API models (general-purpose)}} \\
\addlinespace[1pt]
\ilogo{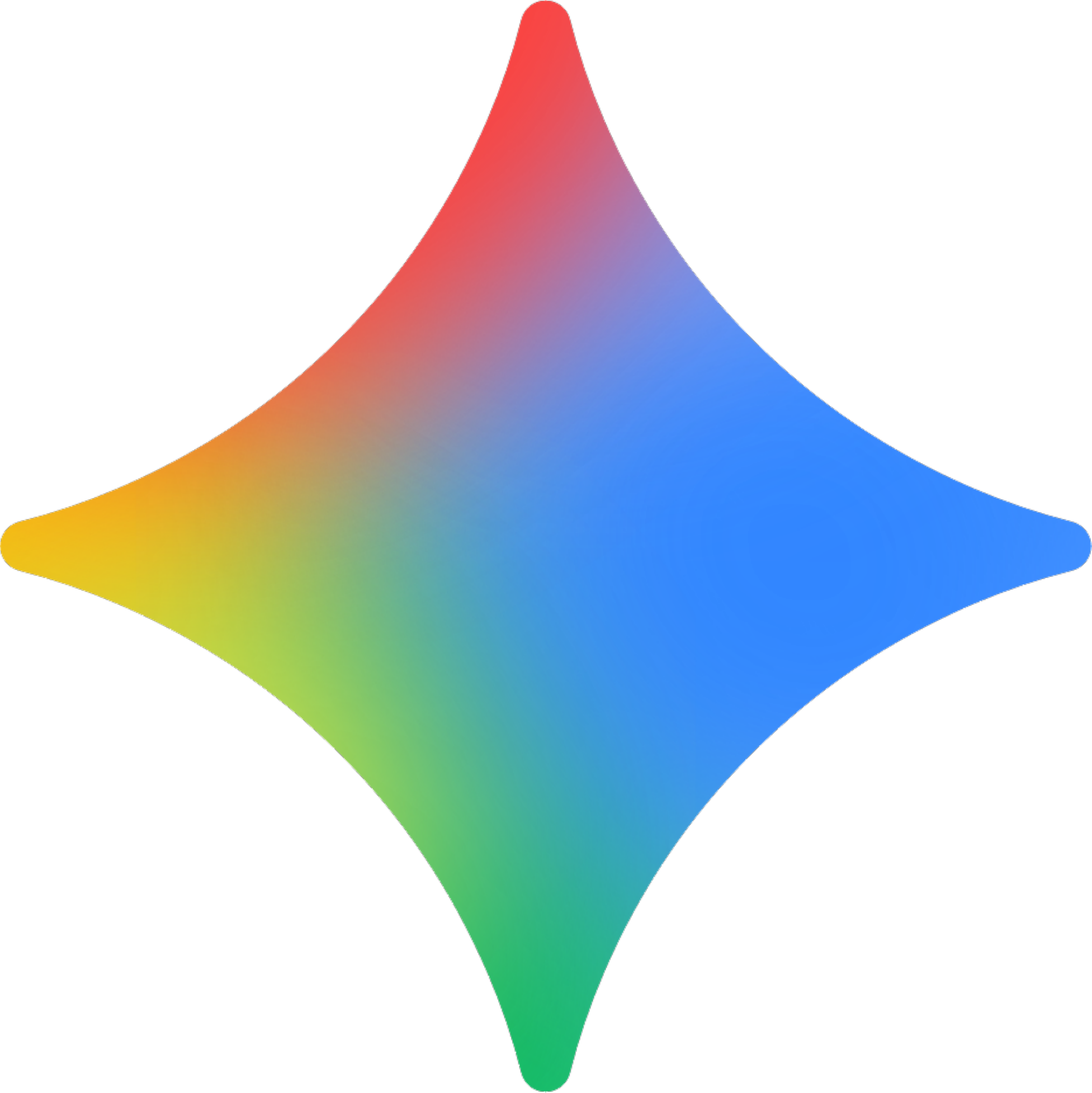}\gflash{} & \icoV\,\icoA & 68/\dashuline{61} & \textbf{72}/\textbf{69} & 76/66 & \textbf{77}/74 & \textbf{78}/\textbf{74} & \textbf{79}/\textbf{74} & \textbf{79}/\textbf{78} & \textbf{75.4}/\textbf{71.0} & $+$4.4 \\
\ilogo{gemini}\gpro{} & \icoV\,\icoA & 67/\textbf{64} & 63/60 & 73/70 & 71/70 & 69/68 & 73/\dashuline{73} & 74/\dashuline{71} & 70.2/68.1 & $+$2.1 \\
\ilogo{gemini}\glite{} & \icoV\,\icoA & 49/51 & 56/52 & 59/54 & 61/50 & 63/59 & 58/57 & 71/61 & 59.5/54.9 & $+$4.6 \\
\ilogo{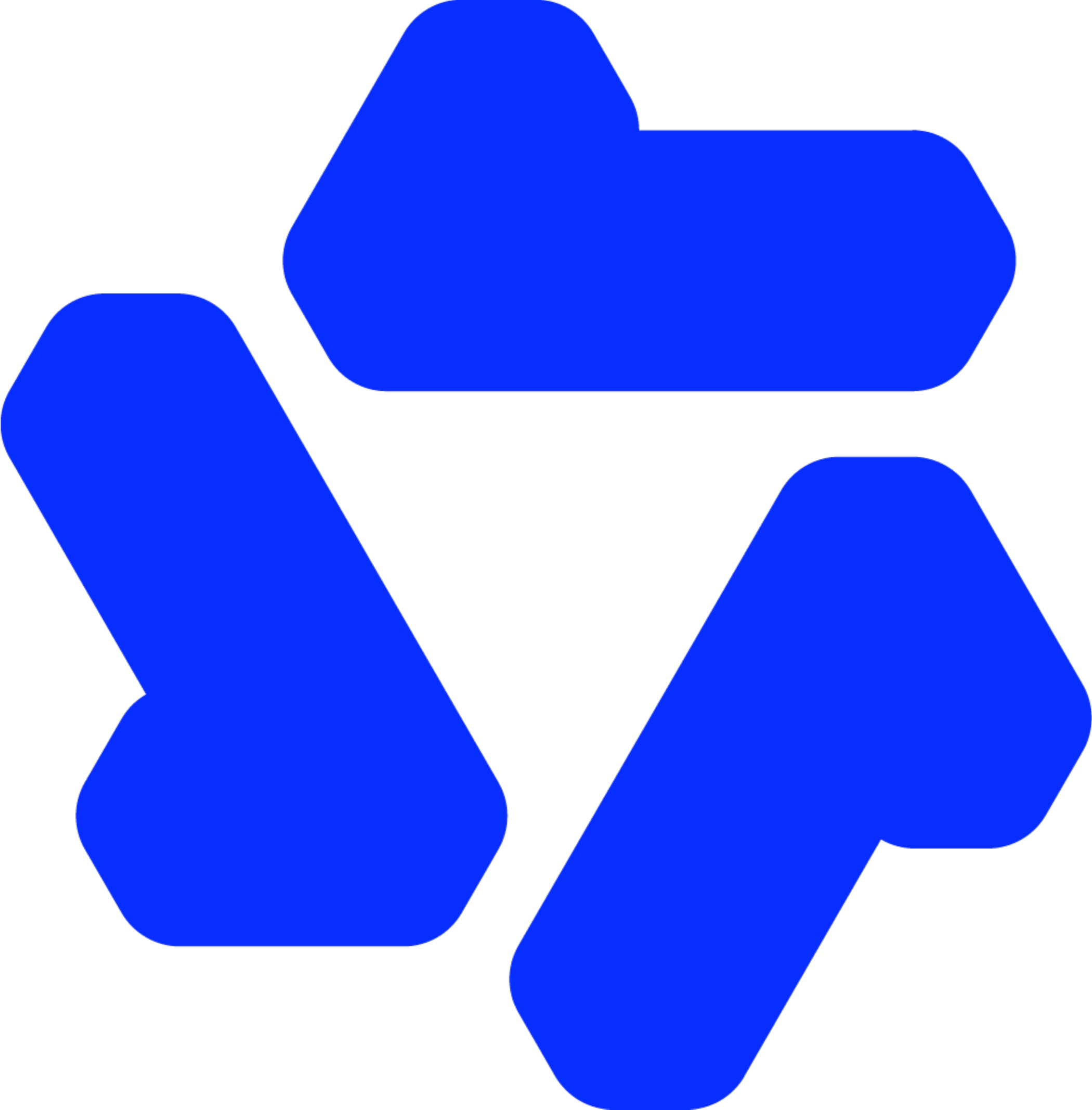}\qmax{} & \icoV\,\icoT & \textbf{76}/59 & 64/\dashuline{68} & \textbf{82}/\textbf{74} & \dashuline{73}/\textbf{78} & \dashuline{76}/\dashuline{71} & 72/68 & 74/70 & \dashuline{73.9}/\dashuline{69.7} & $+$4.2 \\
\ilogo{qwen}\qflash{} & \icoV\,\icoT & 64/\dashuline{61} & 62/62 & 73/\dashuline{72} & 71/61 & 65/63 & 68/61 & 69/51 & 67.5/61.7 & \dashuline{$+$5.7} \\
\ilogo{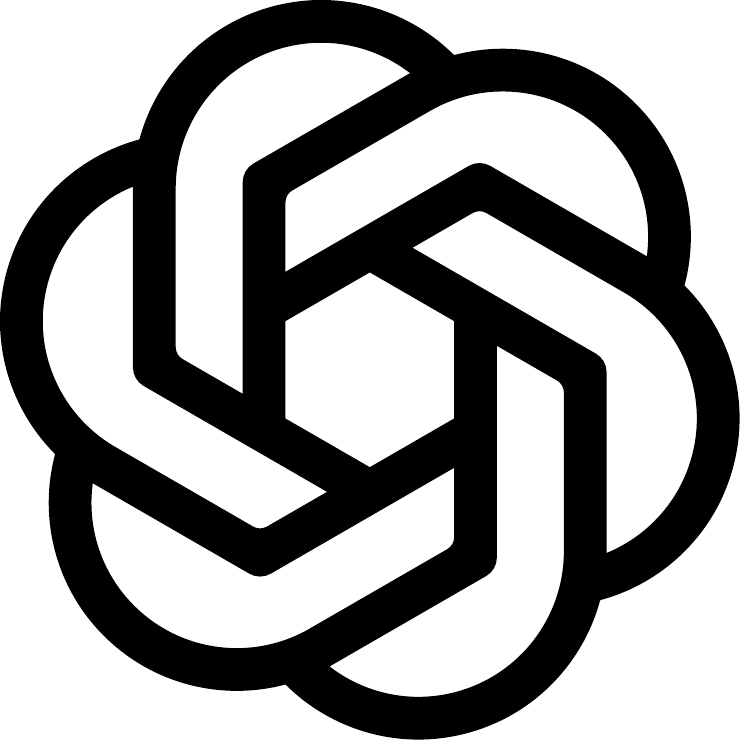}\gptt{} & \icoV\,\icoT & 66/\textbf{64} & \dashuline{68}/66 & \dashuline{81}/70 & 71/\dashuline{77} & 69/\dashuline{71} & \dashuline{77}/\dashuline{73} & \dashuline{77}/67 & 72.5/\dashuline{69.7} & $+$2.9 \\
\ilogo{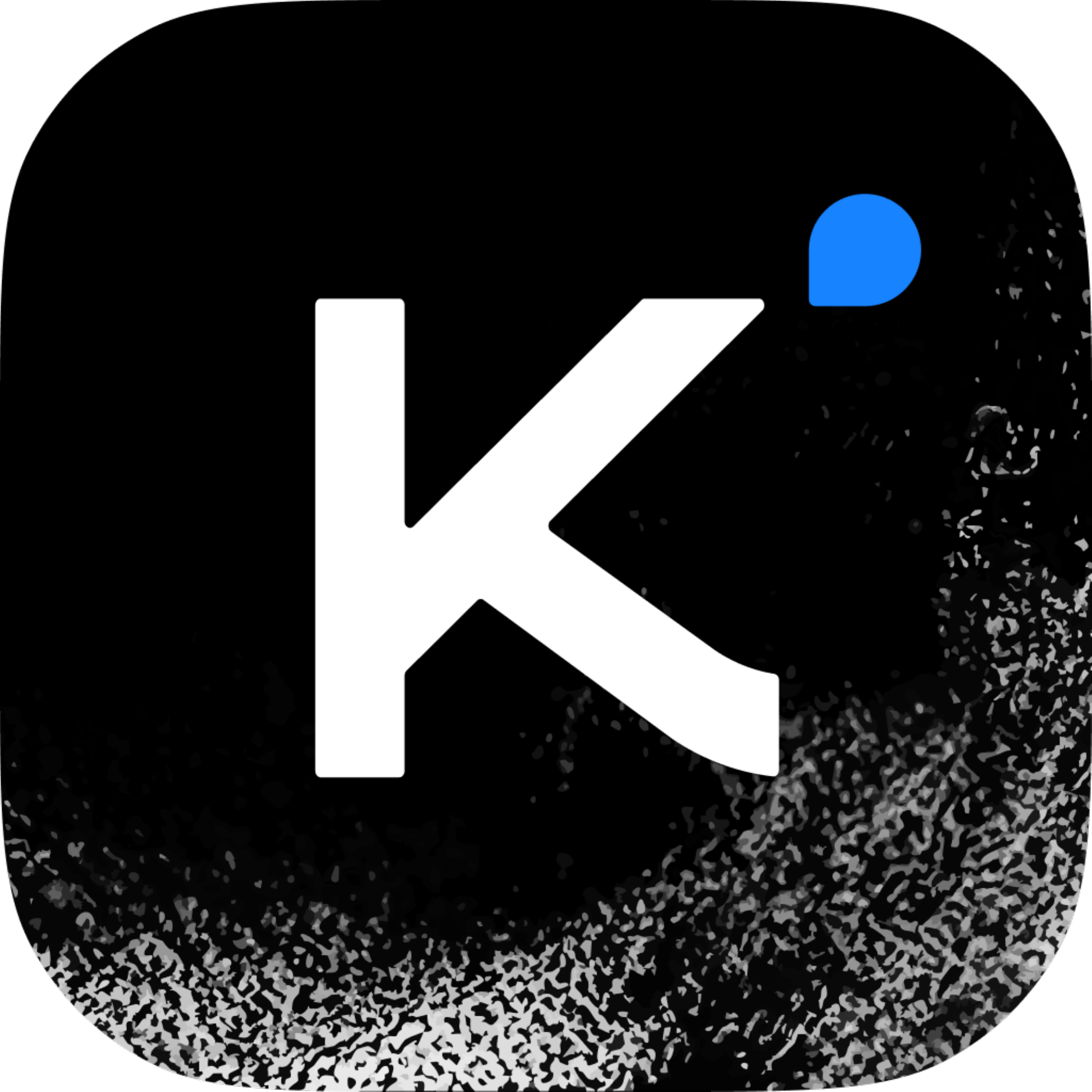}\kimi{} & \icoV\,\icoT & \dashuline{71}/60 & \dashuline{68}/61 & 77/64 & 70/67 & 73/70 & 67/69 & 75/69 & 71.4/65.7 & \dashuline{$+$5.7} \\
\midrule
\multicolumn{11}{c}{\emph{Open-weight local models (general-purpose)}} \\
\ilogo{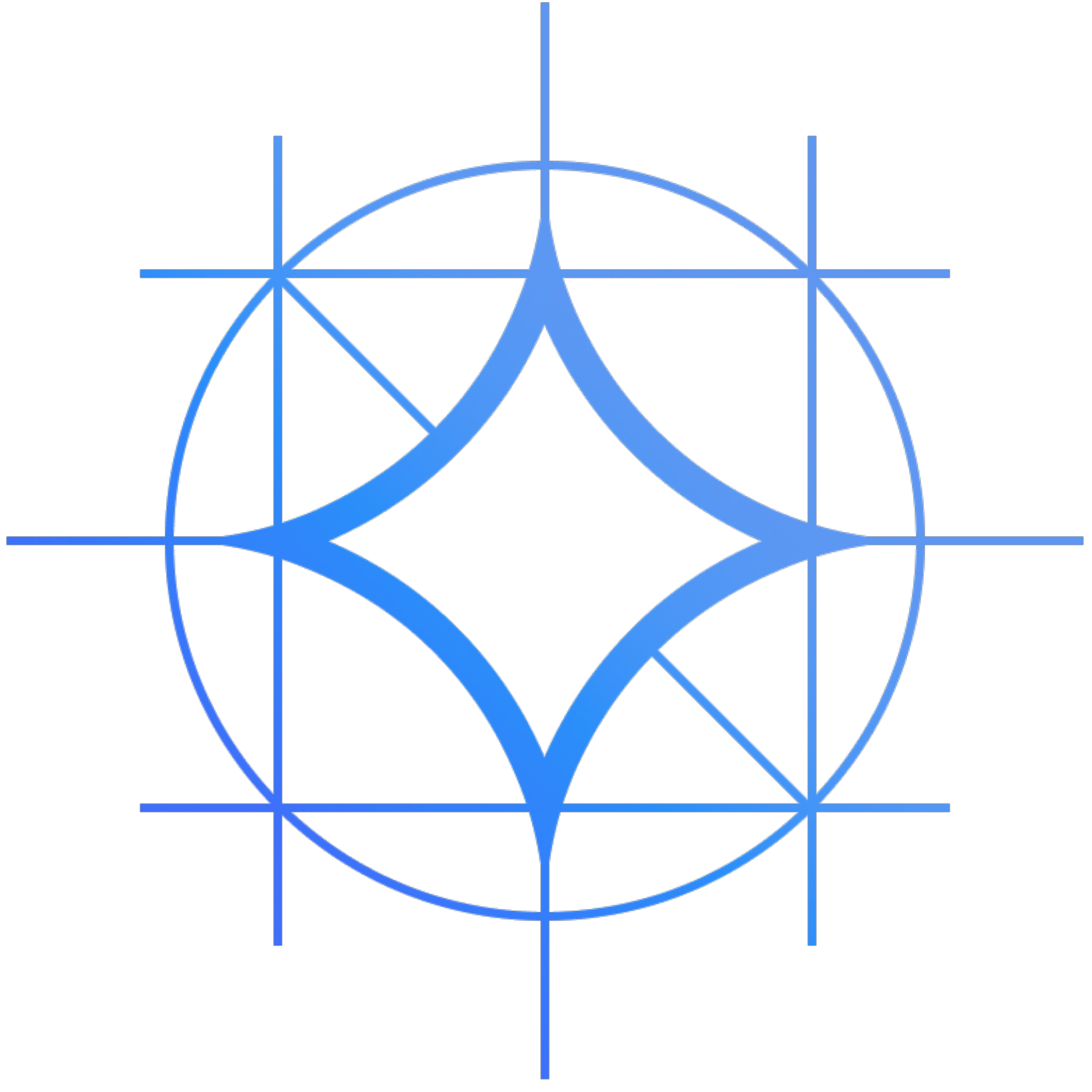}\gemmamoe{} & \icoV\,\icoT & 51/40 & 58/46 & 60/48 & 59/51 & 63/51 & 61/52 & 66/57 & 59.7/49.2 & {\boldmath$+$}\textbf{10.5} \\
\ilogo{gemma}\gemmaetwo{} & \icoV\,\icoT & 43/47 & 50/42 & 56/50 & 44/42 & 42/46 & 39/41 & 48/41 & 46.0/44.1 & $+$1.9 \\
\ilogo{gemma}\gemmaefour{} & \icoV\,\icoT & 43/43 & 42/47 & 47/52 & 38/40 & 51/41 & 41/40 & 48/43 & 44.3/43.8 & $+$0.5 \\
\ilogo{qwen}\qwennine{} & \icoV\,\icoT & 64/51 & 53/56 & 59/49 & 46/53 & 63/52 & 58/51 & 53/61 & 56.7/53.3 & $+$3.3 \\
\ilogo{qwen}\qwenfour{} & \icoV\,\icoT & 51/53 & 57/58 & 61/56 & 64/57 & 58/50 & 57/49 & 46/44 & 56.2/52.4 & $+$3.8 \\
\ilogo{qwen}\qwentwo{} & \icoV\,\icoT & 49/42 & 47/51 & 53/51 & 51/39 & 50/47 & 56/47 & 56/49 & 51.6/46.5 & $+$5.1 \\
\ilogo{qwen}\qomni{} & \icoV\,\icoA & 43/39 & 50/49 & 54/49 & 51/51 & 50/49 & 49/51 & 44/47 & 48.9/47.8 & $+$1.1 \\
\midrule
\multicolumn{11}{c}{\emph{Open-weight local models (fine-tuned as judges)}} \\
\addlinespace[1pt]
\ilogo{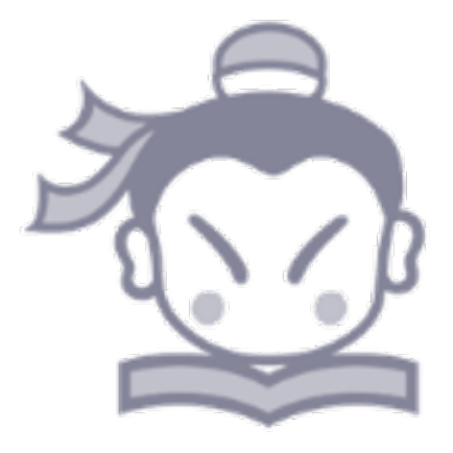}InternLM-2.5-Reward\!\!\! & \icoV\,\icoT & 58/43 & 54/59 & 48/50 & 54/57 & 47/58 & 51/47 & 57/67 & 52.7/54.3 & $-$1.6 \\
\ilogo{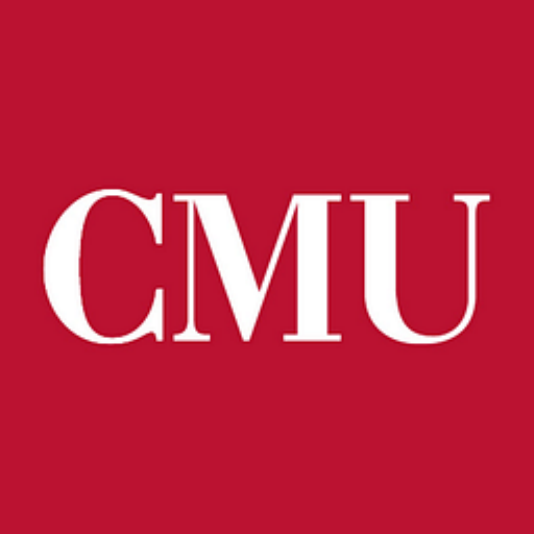}\cmuthree{} & \icoV\,\icoT & 41/44 & 42/47 & 42/40 & 50/43 & 51/49 & 53/54 & 59/62 & 48.3/48.3 & $+$0.0 \\
\ilogo{cmu}\cmuseven{} & \icoV\,\icoT & 49/51 & 49/48 & 34/36 & 48/50 & 49/50 & 48/52 & 58/57 & 47.8/49.0 & $-$1.3 \\
\bottomrule
\end{tabular}
\vspace*{-0.35cm}
\end{table}

\textbf{Setup.~~} No judge can read a $100$-hour collection at once, so each pair is judged over frames drawn from the question's domain in one of two ways: (i) \emph{Retrieval} ranks the indexed $30$-second chunks by similarity to the question and takes two frames from each of the top-$64$;\footnote{We primarily use Qwen3-VL-Embedding-8B as the retriever in video-text modality. Section \ref{sec:modality} also compares Qwen3-VL-Embedding-2B, WeMM-Embedding-9B, and Omni-Embed-Nemotron-3B over the same 30-second chunks, together with  various combinations of modalities; WeMM-Embedding-9B is also used for the embedding-width ablation in  Appendix~\ref{app:retrieval-budget}. See Appendix~\ref{app:retriever-config} for detailed retriever configurations.
} and (ii) \emph{Uniform Sampling} spaces the frames evenly over the whole domain, without looking at the question. In both, the judge receives $128$ frames of size $640\times480$, resized from 720p. Audio-capable judges additionally receive the audio of the chunks the frames come from, merged into one file with $0.6$-second pauses for the three Gemini models, whose API accepts a single audio input, and as separate turns for the locally run \qomni{}; all other judges receive those chunks' ASR transcript.

\textbf{Judge models.~~} We evaluate $17$ video-language models from eight families. Fourteen are general-purpose: seven hosted API models (Gemini-3.1-Pro, 3.5-Flash-Lite, and 3.7-Flash; GPT-5.6-Terra; Kimi-K2.6; Qwen-3.7-Flash and 3.8-Max) and seven open-weight models run on our own GPUs (Gemma-4-E2B, E4B, and 26B-A4B; Qwen-3.5-2B, 4B, and 9B; \qomni{}). Three are fine-tuned as judges: InternLM-XComposer-2.5-Reward \citep{zang2025internlm}, a reward model trained on text, image, and video preferences, and VideoJudge-3B/7B \citep{waheed2026videojudge}, trained for video understanding evaluation. 
Refer to Appendix \ref{app:judge-config} for configuration details.

\textbf{Metrics.~~} Following \citet{lambert2025rewardbench, waheed2026videojudge}, a judge is scored by ``pairwise accuracy,'' the fraction of the $630$ pairs on which it prefers the answer the benchmark designates as better; the side each answer appears on is fixed per pair by a shared seed. A retriever is scored at the $10$--$30$-minute segment level, where evidence is grounded: 
for each of the $343$ questions, it ranks the few hundred segments of the domain by the mean of their three highest chunk similarities, and we report ``Hit@$k$,'' the fraction of questions with {both} gold segments in the top $k$~\citep{xiong2021answering}, and ``nDCG@$k$,'' which further rewards ranking them near the top, with $k=10$. Requiring both segments makes Hit@$k$ strict (chance $0.11\%$ at $k=10$), since a two-segment question cannot be answered from one. Appendix \ref{app:retrieval-metrics} and Table \ref{tab:app-hit-breakdown} give the definitions and per-segment breakdown.

\vspace*{-0.1cm}
\subsection{Trust: Judges Fall Far Short of Humans}
\label{sec:key_findings}
\label{sec:trust}
\vspace*{-0.1cm}

\textbf{Even the strongest judges fall far short of humans.~~}
The best judge overall, \gflash{}, reaches only $75.4\%$ pairwise accuracy with retrieved evidence (Table~\ref{tab:e01-main}), trailing the $93.0\%$ agreement of the human annotators (Appendix~\ref{app:human-eval}).
The remaining hosted judges cluster below it, from \qmax{} ($73.9\%$), \gptt{} ($72.5\%$), and \kimi{} ($71.4\%$) down to \glite{} ($59.5\%$), showing that reliable judgment over ultra-long video is far from solved even for frontier systems.

\textbf{Smaller and open-weight judges approach chance.~~}
Below the hosted API models, the general-purpose open-weight models perform substantially worse: only \gemmamoe{} ($59.7\%$) rises clearly above the $50\%$ chance floor, while the rest fall between $44$ and $57\%$ (Table~\ref{tab:e01-main}). Accuracy broadly tracks scale, with the largest member of each family ahead of its smaller variants, indicating that the spatio-temporal reasoning our pairs demand emerges only at scale.

\textbf{Domain-specific short-video reward models do not transfer.~~}
The judges fine-tuned for video evaluation, \ixc{}, \cmuseven{}, and \cmuthree{}, perform no better than chance ($47.8$--$52.7\%$), despite being trained for video preference tasks. Retrieval, which lifts the general-purpose models, leaves them unchanged or slightly worse (\emph{e.g.}, $-1.6$ points for \ixc{}), suggesting that short-context training does not equip a judge to locate and reason over evidence scattered across a $100$-hour collection.

\textbf{Retrieval helps, but only for models that can exploit it.~~}
Supplying domain-retrieved frames in place of uniform sampling improves nearly every capable judge, by up to $+10.5$ points for \gemmamoe{} and $+4.4$ for \gflash{} (Table~\ref{tab:e01-main}); the exceptions are the fine-tuned critics, whose accuracy does not move. Delivering the right evidence is therefore necessary but not sufficient: it raises accuracy only when the judge can reason over what it receives.

\textbf{Weaker judges also show position bias.} Swapping the two answers' A/B sides flips roughly half of the verdicts for the weaker models, while stronger judges stay more consistent (Appendix~\ref{app:position-bias}).

\vspace*{-0.1cm}
\subsection{Length: Accuracy Falls as the Playlist Grows}
\label{sec:length}
\vspace*{-0.1cm}

\begin{figure}[t]
\centering
\includegraphics[width=5.33in]{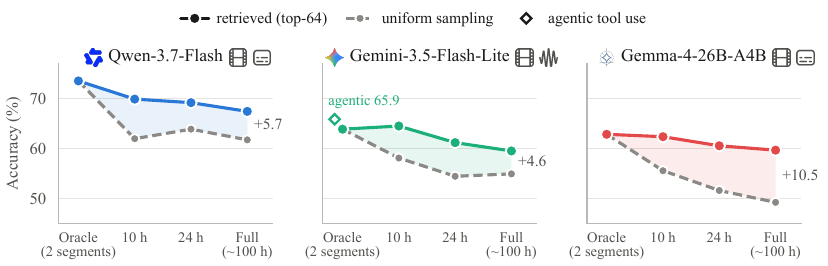}
\vspace*{-0.4cm}
\caption{
\textbf{Retrieval scope.} %
Pairwise accuracy of three judges as the haystack shrinks from the full $\sim$100-hour domain to 24- and 10-hour corpora around the gold segments, down to the oracle (the two gold segments alone). Solid: retrieved evidence; dashed: uniform sampling over the same scope; shaded: retrieval gain. $N=630$ pairs per point.
}
\vspace*{-0.2cm}
\label{fig:scope-ladder}
\end{figure}

\textbf{Judge accuracy declines as the collection grows.~~}
We sweep the retrieval haystack from the two oracle segments ($\sim$1 hour) through frozen 10- and 24-hour corpora to the full $\sim$100-hour domain (Figure~\ref{fig:scope-ladder}). Every judge is most accurate at the oracle scope and degrades as the haystack expands: with retrieval, \qflash{} falls from $73.6\%$ to $67.5\%$, \glite{} from $63.9\%$ to $59.5\%$, and \gemmamoe{} from $62.9\%$ to $59.7\%$ between the oracle and full scopes.
\textbf{Retrieval matters more as the haystack grows.}
Uniform frame sampling degrades far more steeply than retrieval over the same corpora---\gemmamoe{} loses $13.7$ points from oracle to full under uniform sampling but only $3.2$ with retrieval---so the retrieval advantage widens with collection size, reaching $+10.5$ points at the full domain. Retrieval narrows but never closes the length penalty, leaving a persistent gap from the oracle even for the strongest judge.

\vspace*{-0.1cm}
\subsection{Modality: Neither Frames nor Transcript Suffices}
\label{sec:modality}
\vspace*{-0.05cm}

\begin{figure}[t]
\centering
\includegraphics[width=5.4in]{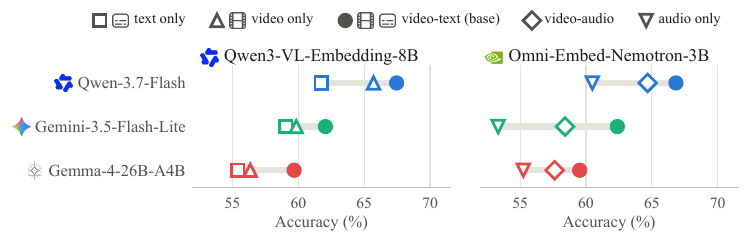}
\vspace*{-0.4cm}
\caption{
\textbf{Judge accuracy by modality.} Accuracy under vision-text (left) and vision-audio (right) retrieval, varying the judge's input stream. 
}
\vspace*{-0.3cm}
\label{fig:modality}
\end{figure}

\textbf{Both frames and transcript are needed.~~}
Judges peak with combined video-text input, and removing either the frames or the transcript costs $2$--$7$ points across judges under our default retriever \qwembeight{} (Figure~\ref{fig:modality}, left). No single modality suffices: our pairs are constructed so that the deciding evidence appears in the frames while the surrounding context is carried by the narration, and both must be exploited for optimal performance.

\Needspace{11\baselineskip}
\begin{wraptable}{r}{0.57\linewidth}
\vspace*{0.1cm}
\vspace{-\baselineskip}
\centering
\small
\setlength{\tabcolsep}{3pt}
\setlength{\ULdepth}{1.2pt}
\caption{
Retriever performance evaluation.
}
\label{tab:e04-02-retrievers}
\footnotesize
\begin{tabular}{llcc}
\toprule
Retriever & Input & Hit@10 & nDCG@10 \\
\midrule
\rlogo{qwen}\qwembeight{} & \icoV\,\icoT & \textbf{37.9} & \textbf{45.0} \\
 & \icoV & 32.9 & 37.9 \\
 & \icoT & 30.0 & 37.0 \\
\addlinespace[3pt]
\rlogo{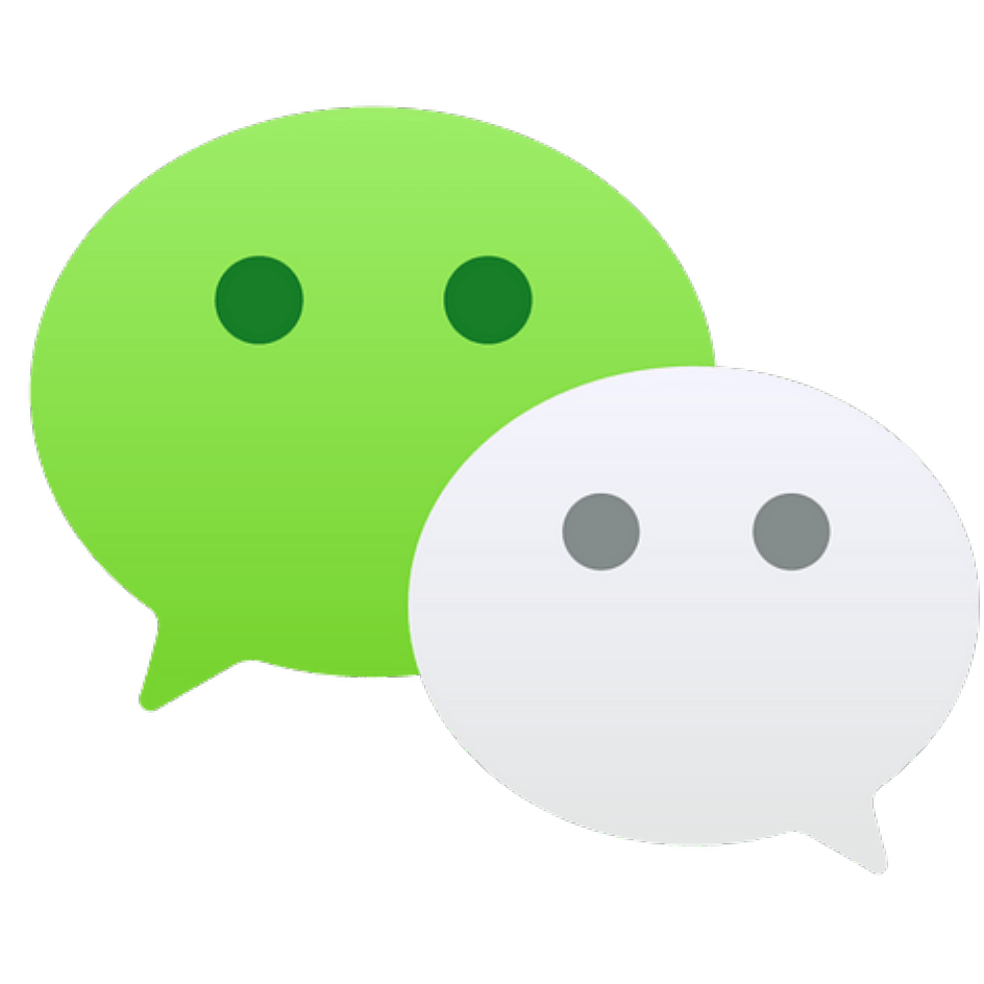}\wemm{} & \icoV\,\icoT & \dashuline{37.6} & \dashuline{42.9} \\
\addlinespace[3pt]
\rlogo{qwen}\qwembtwo{} & \icoV\,\icoT & 32.7 & 38.1 \\
\addlinespace[3pt]
\rlogo{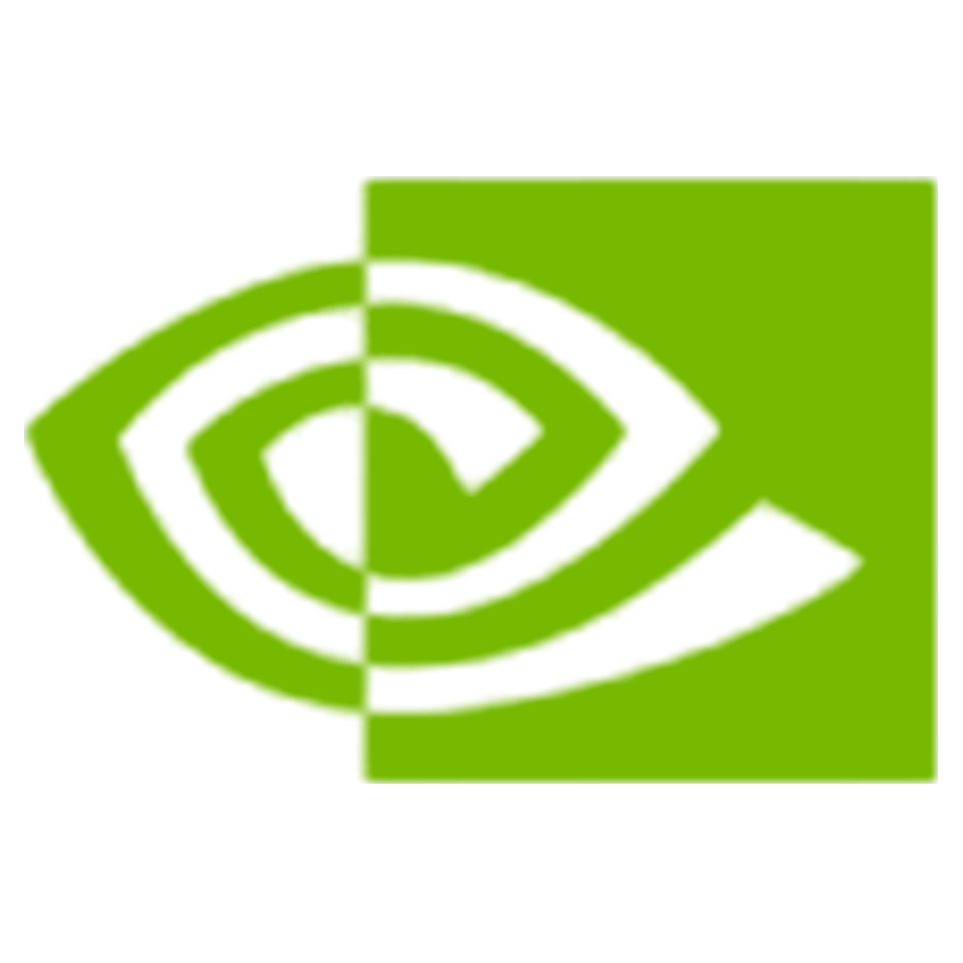}\nemo{} & \icoV\,\icoT & 30.6 & 39.9 \\
 & \icoV\,\icoA & 26.2 & 33.4 \\
 & \icoA & \phantom{0}7.3 & 14.0 \\
\bottomrule
\end{tabular}
\vspace{-0.5\baselineskip}

\end{wraptable}

\noindent\textbf{Text-space retrieval beats audio-space retrieval.}
Retrieving in video-text space consistently beats video-audio retrieval with \nemo{}, both in retriever quality: Hit@10 of $30.6\%$ (video-text) vs. $26.2\%$ (video-audio); and $7.3\%$ (audio only), as in Table \ref{tab:e04-02-retrievers}; and in the downstream judges each index feeds (Figure \ref{fig:modality}, right). We deliberately exclude questions about purely auditory properties, such as music or vocal tone, so that the two modalities are compared fairly, and under this setting text is the stronger channel for both retrieval and judgment.

\vspace*{-0.1cm}
\subsection{Bottleneck: Finding the Evidence, Not Seeing It}
\label{sec:bottleneck}
\vspace*{-0.1cm}

\begin{wraptable}{r}{0.50\linewidth}
\vspace*{-0.65cm}
\centering
\small
\setlength{\tabcolsep}{3pt}
\caption{
Effect of reasoning budget.
}
\label{tab:e07-02-reasoning-thinking-level}
\begin{tabular}{lcccc}
\toprule
 & & \multicolumn{2}{c}{Tokens} & \\
\cmidrule(lr){3-4}
Setting & Accuracy & Prompt & Output & USD \\
\midrule
minimal & 51.7 ($-$7.8) & 84k & \phantom{0,}364 & 8.26 \\
low & 59.2 ($-$0.3) & 84k & 1,035 & 8.79 \\
medium (base) & 59.5 \phantom{($-$7.8)} & 84k & 1,447 & 9.11 \\
high & \textbf{59.7} ($+$0.2) & 84k & 1,948 & 9.50 \\
\bottomrule
\end{tabular}
\vspace{-0.5\baselineskip}
\end{wraptable}

\textbf{Effect of reasoning.~~}
\emph{Reasoning helps, then saturates.} Enabling reasoning lifts \glite{} by $7.8$ points over its \emph{minimal} setting ($51.7$$\to$$59.5\%$), but further budget barely moves it: from \emph{low} to \emph{high} accuracy rises only $0.5$ points ($59.2$$\to$$59.7\%$) even as output tokens more than quintuple (Table~\ref{tab:e07-02-reasoning-thinking-level}). A modest amount of thinking is therefore worthwhile, but scaling it up does not close the gap to humans. We also tested agentic reasoning with tool use, which yielded only marginal gains (Appendix~\ref{app:agentic}).

\textbf{Effect of pixel budget.~~}
\emph{Pixel budget is not the bottleneck.} Raising the judge's per-frame token budget changes accuracy by at most $1.3$ points (\emph{high} $60.8\%$ vs.\ \emph{low} $59.5\%$) while roughly doubling API cost ($\$9.11\to\$18.83$), and \emph{medium} is no better than \emph{low} (Table~\ref{tab:e07-03-image-detail-media-resolution}). The limiting factor is thus finding the right evidence, not resolving fine visual detail once it is in view.

\textbf{Effect of audio speed.~~}
\emph{Faster audio does not hurt.} Playing the merged audio at $2\times$ halves its duration and cuts prompt tokens from $84$k to $60$k, lowering cost by $25\%$ ($\$9.11\to\$6.84$) while slightly \emph{improving} accuracy ($+1.6$ points; Table~\ref{tab:e07-04-audio-speed}). 
The shorter context helps rather than hurts, again pointing to context length, not evidence resolution, as the constraint.

\vspace*{-0.1cm}
\subsection{Error Analysis}
\label{sec:error_analysis}
\vspace*{-0.1cm}

To understand why judges made mistakes, we examined 100 pairs that the four strongest judges judged incorrectly and identified the root cause of each 
through manual inspection by the authors.
These failures fall into four types:
\textbf{1) Retrieval error} (Appendix Fig.~\ref{fig:failure-retrieval}), the dominant failure: the retriever never surfaces the deciding evidence, so it never reaches the judge, in about $61\%$ of examined cases.
\textbf{2) Frame sampling error} (Fig.~\ref{fig:failure-sampling}): the right $30$-second chunk is retrieved, but none of its sampled frames land on the deciding moment, in about $15\%$ of cases.
\textbf{3) Reasoning error} (Fig.~\ref{fig:failure-reasoning}): the evidence is served, but the judge is misled by adversarial retrieved chunks and fails to identify the correct visual cues, in about $16\%$ of cases.
\textbf{4) Perception error} (Fig.~\ref{fig:failure-perception}): the deciding frame is served and legible, but the judge misreads it, \emph{e.g.}\ handwritten text or a character's attributes, in about $6\%$ of cases (see Appendix \ref{app:failures} for all examples). 
The remaining ${\approx}2\%$ we attribute to annotation noise in the ground-truth ratings.

\vspace*{-0.1cm}
\section{Conclusion}
\vspace*{-0.1cm}

In this work, we introduce \bench{}, a fully automated framework that constructs video-language judge benchmarks over
$700$ hours of playlist collections without human annotation.
Our analysis reveals a wide gap between human and model accuracy where frontier video-language judge systems fail to retrieve and reason reliably.
Moreover, judge models fine-tuned on short clips fail to transfer, and judge accuracy degrades as the collection grows, demonstrating that ultra-long spatio-temporal verification remains largely unsolved.
We hope \bench{} plays a foundational role in the development of judge systems that are trustworthy over day-long video collections.

\subsection*{AI use statement}

In this work, we used generative AI tools to
generate synthetic data sets (Sections~\ref{sec:qa_gen}--\ref{sec:final}),
design or provide feedback on research methodology or experiments,
implement methods (e.g., for coding and verification),
assist with translation,
clean and reformat dataset,
support qualitative and thematic data analysis.
We have not used generative AI tools to 
propose or refine hypotheses, interpret results; and 
to help develop theoretical models or conceptual frameworks, 
formulate mathematical claims, 
provide critical ingredients for proving mathematical claims, 
assist in the writing of proofs
are not applicable to this work. 
Additionally, we used generative AI tools to 
create or modify scientific figures or images, 
create or edit software code,  
creation of artifacts, 
summarize or analyse existing literature (e.g., finding related benchmarks and AI models), 
brainstorming, 
sourcing/searching for information, 
edit a research paper to improve readability, 
and
identify relevant literature.
For citations, we use official bibtex entries manually exported and verified from ICLR Proceedings, NeurIPS Proceedings, PMLR, ACL Anthology, CVF, ACM Digital Library and arXiv. For web URLs, we use a consistent format.
We have reviewed all AI-assisted work. 
LLM-generated code was verified and tested for correctness by all authors.
We take responsibility for the final content of this work, including text, claims or artifacts produced with the aid of generative AI.

\subsection*{Reproducibility statement}
We provide detailed configurations for retrieval (Appendix~\ref{app:retriever-config},~\ref{app:retrieval-metrics}), judge models (Appendix~\ref{app:judge-config}), and data generation (Appendix~\ref{app:datagen-config}), along with benchmark statistics and a complete data item (Appendices~\ref{app:data-stats},~\ref{app:full-item}). All prompts are reproduced verbatim (Appendices~\ref{app:rm_prompt}--\ref{app:prompt-ground}), and cost breakdowns (Appendices~\ref{app:datagen-cost},~\ref{app:eval-cost}) and the human evaluation protocol (Appendix~\ref{app:human-eval}) are included. We will release our pipeline, benchmark, and evaluation code upon publication.

\subsection*{Ethics statement}
All videos in \textsc{PlaylistEval} are publicly available YouTube content. We release only the video URLs, timestamps, and our generated annotations; no raw video or audio files. 
YouTube videos may be removed or made private by their owners, any URL in our dataset may become inaccessible over time.
For cases where a video is no longer publicly available, we will share the corresponding files only with the explicit permission of the video owner. Our benchmark does not collect or store any personally identifiable information beyond what is visible in the public videos themselves. As the study poses minimal risk and ensures participant anonymity during crowd-sourced human annotation, we did not seek ethics board approval, in accordance with standards commonly considered exempt from review.

\newpage
\subsection*{Acknowledgements}
This work was supported by Institute of Information \& communications Technology Planning \& Evaluation (IITP) grant funded by the Korea government (MSIT) (No. RS-2024-00445087 \& RS-2025-25464461) and National IT Industry Promotion Agency(NIPA) grant funded by the Korea government(MSIT) (No. RS-2026-25621604).

\bibliography{iclr2027_conference}
\bibliographystyle{iclr2027_conference}

\appendix
\raggedbottom
\setlength{\abovetopsep}{5pt}

\addtocontents{toc}{\protect\setcounter{tocdepth}{2}}
\renewcommand{\contentsname}{Appendix Contents}
\clearpage          %
\tableofcontents

\section{
Limitations
and Future Work
}

\textbf{Generator bias.} 
\bench{} relies on \gflashgen{} for question and answer generation, with other model families reserved for verification. This asymmetry may introduce systematic biases in the generated data. As cheaper omni-modal models become available~\citep{qwen2026qwen38omni}, future work could diversify the generator pool to mitigate family-specific artifacts.

\textbf{Evaluation cost.} Because each preference pair requires serving long video to every judge, scaling the benchmark to substantially more pairs or judges increases cost. Nevertheless, automated evaluation remains an order of magnitude cheaper than manual human annotation at this video duration scale (see Appendix~\ref{app:datagen-cost} for cost breakdown).

\textbf{Language coverage.} The current playlist collection is primarily in English, so the benchmark may not capture challenges specific to multilingual or low-resource settings, where ASR, embedding and generative models may underperform at various stages of our framework.

\section{Additional Experimental Settings}

\subsection{Retriever configurations}
\label{app:retriever-config}

Every video segment is split into 30-second chunks, 
and each retriever embeds a chunk once into a single vector 
(Table~\ref{tab:app-retrievers}). 
At evaluation time the chunks of the question's
domain are ranked and the best 64 are handed to the judge in \emph{chronological order} 
in the video and \emph{video order} in the playlist collection.
We discuss this default setup in Table~\ref{tab:app-retrieval-setup}  with the settings our experiments vary around it, and 
in Table~\ref{tab:app-asr} the automated speech recognition system behind a chunk's transcript.

\begin{table}[H]
\centering
\small
\setlength{\tabcolsep}{4pt}
\caption{\textbf{Retrievers.} Every index stores one float32 vector per 30-second chunk, for
about 100k
chunks across the seven domains. Latency is the average measured time to answer a single query.
}
\label{tab:app-retrievers}
\begin{tabular}{@{}lrrr@{}}
\toprule
Retriever & Dim. & Index size & Query latency (H200 GPU) \\
\midrule
\rlogo{qwen}\qwembeight{}~\citep{li2026qwen3vlembedding} & 4{,}096 & 1.59\,GB & 13.0\,ms \\
\quad{\scriptsize\hf{Qwen/Qwen3-VL-Embedding-8B}} & & & \\
\addlinespace[2pt]
\rlogo{qwen}\qwembtwo{}~\citep{li2026qwen3vlembedding} & 2{,}048 & 0.79\,GB & 7.5\,ms \\
\quad{\scriptsize\hf{Qwen/Qwen3-VL-Embedding-2B}} & & & \\
\addlinespace[2pt]
\rlogo{wechat}\wemm{}~\citep{zhou2026wemmembedding} & 4{,}096 & 1.59\,GB & 13.4\,ms \\
\quad{\scriptsize\hf{tencent/WeMM-Embedding-9B}} & & & \\
\addlinespace[2pt]
\rlogo{nvidia}\nemo{}~\citep{xu2025omniembed} & 2{,}048 & 0.79\,GB & 7.5\,ms \\
\quad{\scriptsize\hf{nvidia/omni-embed-nemotron-3b}} & & & \\
\bottomrule
\end{tabular}
\end{table}

\begin{table}[H]
\centering
\small
\setlength{\tabcolsep}{5pt}
\caption{\textbf{Retrieval settings.} Each default applies to every experiment except the one that
varies that setting; the last column lists the values swept and where. The pixel limits, input
length, and precision are the loading arguments of \qwembeight{}, our default retriever.}
\label{tab:app-retrieval-setup}
\begin{tabular}{@{}l>{\raggedright\arraybackslash}p{0.30\linewidth}>{\raggedright\arraybackslash}p{0.34\linewidth}@{}}
\toprule
Setting & Default & Varied over \\
\midrule
\multicolumn{3}{@{}l}{\emph{Retrieval index}} \\
Indexed unit & 30-second chunk & --- \\
Chunk representation & video + transcript & video only, text only;
  video + audio, audio only (Fig.~\ref{fig:modality}) \\
Frames per chunk vector & at most 16 & 8, 4 (Fig.~\ref{fig:app-retrieval-budget} left) \\
Pixels per frame & $64\times64$ to $960\times720$ & --- \\
Pixel budget per chunk & frames $\times\,960\times720$, so every frame keeps its full size & --- \\
Input length & at most 32{,}000 tokens & --- \\
Precision & bfloat16 & --- \\
Embedding width & full & 2{,}048 to 64 dimensions (\wemm{}; Fig.~\ref{fig:app-retrieval-budget} right) \\
\addlinespace[3pt]
\multicolumn{3}{@{}l}{\emph{Search}} \\
Corpus & the question's domain ($\sim$100\,h) & 24\,h, 10\,h, oracle
  (Fig.~\ref{fig:scope-ladder}) \\
Retrieved chunks & top 64 & --- \\
Order shown to the judge & chronological & --- \\
Baseline without retrieval & uniform sampling over the same corpus & --- \\
\addlinespace[3pt]
\multicolumn{3}{@{}l}{\emph{Retriever evaluation (Table~\ref{tab:e04-02-retrievers})}} \\
Ranked unit & GT 10-30-minute segments, 247--366 per domain & --- \\
Segment score & mean of its 3 best chunk scores & --- \\
Questions; relevant segments & 343; 2 per question (random Hit@10 0.11\% for both segments,
  6.8\% for either) & --- \\
\bottomrule
\end{tabular}
\end{table}

\begin{table}[H]
\centering
\small
\setlength{\tabcolsep}{6pt}
\caption{\textbf{Speech recognition.} \asr{} produces the transcript that is indexed with every
30-second chunk.}
\label{tab:app-asr}
\begin{tabular}{@{}ll@{}}
\toprule
Setting & Value \\
\midrule
Model & {\footnotesize\hf{Qwen/Qwen3-ASR-1.7B}} \\
Forced aligner & {\footnotesize\hf{Qwen/Qwen3-ForcedAligner-0.6B}}, bfloat16 \\
Inference engine & vLLM \\
Output length & at most 4{,}096 new tokens per audio input \\
\bottomrule
\end{tabular}
\end{table}

\subsection{Retrieval metrics}
\label{app:retrieval-metrics}

We formally define the two metrics used to score retrievers in
Table~\ref{tab:e04-02-retrievers} and Table~\ref{tab:app-hit-breakdown}.

\textbf{Setup.}
The ranking unit is the ground-truth evidence segment, the $10$--$30$ minute portion of a video
named by the dataset. For a question $q$, the corpus is the set of segments in $q$'s domain
($N\approx 288$ on average), and the gold set $G_q=\{g_1,g_2\}$ holds the two segments that contain
its evidence; $|G_q|=2$ for all $343$ questions. Each segment $c$ contains $m\approx 60$
thirty-second chunks with unit-norm embeddings $\mathbf{s}_1,\dots,\mathbf{s}_m$, and the query
embedding $\mathbf{e}_q$ is unit-norm, so an inner product is a cosine similarity. We score a
segment by the mean of its three highest chunk similarities,
\begin{equation*}
\mathrm{score}(c)=\frac{1}{3}\sum_{j\in\mathcal{T}_3(c)}\langle \mathbf{e}_q,\mathbf{s}_j\rangle,
\end{equation*}
where $\mathcal{T}_3(c)$ is the set of three chunks in $c$ most similar to $\mathbf{e}_q$. We
rank all segments by $\mathrm{score}(\cdot)$ in descending order and let $\operatorname{rank}(c)$ be
the resulting $1$-indexed position. Every metric below is reported as a mean over the $343$
questions.

\textbf{Hit@$k$.}
Writing $r_1=\operatorname{rank}(g_1)$ and $r_2=\operatorname{rank}(g_2)$, the default metric
requires \emph{both} segments within the top $k$,
\begin{equation*}
\mathrm{Hit}_{\mathrm{both}}@k=\mathbf{1}\!\left[\max(r_1,r_2)\le k\right],
\end{equation*}
the two-segment analogue of top-$k$ retrieval accuracy: a two-segment question is unanswerable
from a single segment, so the evidence is delivered only when both segments arrive. For comparison
we also report the single-segment convention (at least one segment in the top $k$) and the
per-segment rates,
\begin{equation*}
\mathrm{Hit}_{\mathrm{any}}@k=\mathbf{1}\!\left[\min(r_1,r_2)\le k\right],\qquad
\mathrm{Hit}_{\mathrm{seg}\,h}@k=\mathbf{1}\!\left[r_h\le k\right],
\end{equation*}
which Table~\ref{tab:app-hit-breakdown} lists alongside $\mathrm{Hit}_{\mathrm{both}}@k$. Drawing
$k$ of $N$ segments uniformly at random with two relevant gives the chance floors
\begin{equation*}
P\!\left(\mathrm{Hit}_{\mathrm{any}}@k\right)=1-\frac{\binom{N-2}{k}}{\binom{N}{k}},\qquad
P\!\left(\mathrm{Hit}_{\mathrm{both}}@k\right)=\frac{k(k-1)}{N(N-1)};
\end{equation*}
at $N=288$ and $k=10$ these are $6.8\%$ and $0.11\%$, so a low ``both'' value is not weak in the
way its gap from the ``any'' value would suggest.

\textbf{nDCG@$k$.}
With binary relevance $\mathrm{rel}_i=\mathbf{1}\!\left[\text{the segment at rank }i\in G_q\right]$,
\begin{equation*}
\mathrm{DCG}@k=\sum_{i=1}^{k}\frac{\mathrm{rel}_i}{\log_2(i+1)},\qquad
\mathrm{IDCG}@k=\sum_{i=1}^{\min(|G_q|,\,k)}\frac{1}{\log_2(i+1)},\qquad
\mathrm{nDCG}@k=\frac{\mathrm{DCG}@k}{\mathrm{IDCG}@k}.
\end{equation*}
Unlike Hit@$k$, which is a step function of $k$, nDCG@$k$ is rank-sensitive and rewards placing
both gold segments near the top of the ranking, which matters because the judge's frame budget is
spread over the retrieved chunks in rank order. For $|G_q|=2$ and $k\ge 2$,
$\mathrm{IDCG}@k=1+1/\log_2 3\approx 1.631$. We report $k=10$.

\subsection{Video LM configurations}
\label{app:judge-config}

Table~\ref{tab:app-judges} lists the 17 judges and how each is accessed via official APIs or local GPUs. 
All of them see the same evidence: 
128 frames at $640\times480$, two from each retrieved chunk,
together with a second
stream, which is the chunks' audio for audio-capable judges and
their transcript for rest of the judges
(Table~\ref{tab:app-judge-inputs}).

\begin{table}[H]
\centering
\small
\setlength{\tabcolsep}{4pt}
\caption{\textbf{Details of the 17 judges.} 
Models are grouped by how they are served: through a hosted official API, or on local GPUs. 
The last column gives the narration stream format each judge receives alongside the video frames.
}
\label{tab:app-judges}
\begin{tabular}{@{}lll@{}}
\toprule
Judge & Serving & Stream \\
\midrule
\multicolumn{3}{@{}l}{\emph{Hosted Official APIs}} \\
\addlinespace[1pt]
\ilogo{gemini}\gflash{}~\citep{google2026gemini37flash} & API & audio \\
\ilogo{gemini}\gpro{}~\citep{google2026gemini31pro} & API & audio \\
\ilogo{gemini}\glite{}~\citep{google2026gemini35flashlite} & API & audio \\
\addlinespace[3pt]
\ilogo{openai}\gptt{}~\citep{openai2026gpt56} & API & transcript \\
\ilogo{kimi}\kimi{}~\citep{moonshot2026kimik26} & API & transcript \\
\ilogo{qwen}\qmax{}~\citep{qwen2026qwen38max} & API & transcript \\
\ilogo{qwen}\qflash{}~\citep{qwen2026qwen37flash} & API & transcript \\
\midrule
\multicolumn{3}{@{}l}{\emph{Run on Local GPUs}} \\
\addlinespace[1pt]
\ilogo{gemma}\gemmamoe{}~\citep{gemma2026gemma4} & vLLM & transcript \\
\ilogo{gemma}\gemmaefour{}~\citep{gemma2026gemma4} & vLLM & transcript \\
\ilogo{gemma}\gemmaetwo{}~\citep{gemma2026gemma4} & vLLM & transcript \\
\addlinespace[3pt]
\ilogo{qwen}\qwennine{}~\citep{qwen2026qwen35} & vLLM & transcript \\
\ilogo{qwen}\qwenfour{}~\citep{qwen2026qwen35} & vLLM & transcript \\
\ilogo{qwen}\qwentwo{}~\citep{qwen2026qwen35} & vLLM & transcript \\
\ilogo{qwen}\qomni{}~\citep{xu2025qwen3omni} & vLLM & audio \\
\addlinespace[3pt]
\ilogo{internlm}\ixc{}~\citep{zang2025internlm} & Transformers & transcript \\
\quad{\footnotesize short name: InternLM-2.5-Reward} & & \\
\ilogo{cmu}\cmuseven{}~\citep{waheed2026videojudge} & Transformers & transcript \\
\ilogo{cmu}\cmuthree{}~\citep{waheed2026videojudge} & Transformers & transcript \\
\bottomrule
\end{tabular}
\end{table}

\begin{table}[H]
\centering
\small
\setlength{\tabcolsep}{5pt}
\caption{
\textbf{Judge input.} 
Each default holds in every experiment except the one that varies that setting. 
The three settings in the last group apply only to \glite{}, 
the judge the sweeps are run on.
}
\label{tab:app-judge-inputs}
\begin{tabular}{@{}l>{\raggedright\arraybackslash}p{0.33\linewidth}>{\raggedright\arraybackslash}p{0.33\linewidth}@{}}
\toprule
Setting & Default & Varied over \\
\midrule
Retrieved chunks & 64 & --- \\
Frames per chunk & 2 & --- \\
Frames in total & 128 & --- \\
Frame size & $640\times480$ & $480\times360$, $720\times540$ \\
Transcript & ASR of the retrieved chunks & removed (video only) \\
Audio (audio-capable judges) & the chunks' audio, one track & follows the retrieval scope \\
Frames & kept & removed (text/audio only) \\
Side of the chosen answer & fixed per pair by a shared seed & both sides (position bias) \\
\addlinespace[3pt]
Thinking level & medium & minimal, low, high \\
Media resolution & low & medium, high \\
Audio speed & $1.0\times$ & $1.5\times$, $2.0\times$ \\
\bottomrule
\end{tabular}

\vspace{4pt}
\begin{minipage}{0.92\linewidth}\footnotesize
For audio judges the 64 chunks' audio is merged into one track with 0.6\,s pauses for the
Gemini models, which accept a single audio input, and sent as separate turns to \qomni{}.
\gemmamoe{} pools every frame to the same number of visual tokens 
(\texttt{max\_soft\_tokens}$=280$) by default.
\end{minipage}
\end{table}

\subsection{Data generation configurations}
\label{app:datagen-config}

The remaining tables cover the data pipeline of 
Figure~\ref{fig:pipeline}: the model behind each
block with its reasoning setting (Table~\ref{tab:app-datagen-models}), 
how video and transcripts are supplied (Table~\ref{tab:app-datagen-inputs}), 
and the acceptance rule of every gate (Table~\ref{tab:app-datagen-gates}). 
The prompts and JSON schemas themselves are in Appendix~\ref{app:data_prompt}.

\begin{table}[H]
\centering
\small
\setlength{\tabcolsep}{5pt}
\caption{\textbf{Models used in the data pipeline,} by block of Figure~\ref{fig:pipeline}. 
Gemini models run with their default thinking mode enabled, GPT models with reasoning effort \emph{high}, and \qplus{} with the provider's default. 
We override no sampling parameter (temperature, nucleus size, or output length) for any model.
}
\label{tab:app-datagen-models}
\begin{tabular}{@{}llll@{}}
\toprule
Block & Step & Model & API identifier \\
\midrule
B  & QA generation & \ilogo{gemini}\gflashgen{}~\citep{google2025gemini3flash} & {\scriptsize\texttt{gemini-3-flash-preview}} \\
\addlinespace[3pt]
C2 & transcript-only answer & \ilogo{gemini}\gflashgen{} & {\scriptsize\texttt{gemini-3-flash-preview}} \\
   & verification & \ilogo{openai}\gptmini{}~\citep{openai2026gpt54mini} & {\scriptsize\texttt{gpt-5.4-mini}} \\
   & parametric answer, A & \ilogo{gemini}\gflashgen{} & {\scriptsize\texttt{gemini-3-flash-preview}} \\
   & verification, A & \ilogo{openai}\gptfull{}~\citep{openai2026gpt54} & {\scriptsize\texttt{gpt-5.4}} \\
   & parametric answer, B & \ilogo{openai}\gptmini{} & {\scriptsize\texttt{gpt-5.4-mini}} \\
   & verification, B & \ilogo{gemini}\gpro{}~\citep{google2026gemini31pro} & {\scriptsize\texttt{gemini-3.1-pro-preview}} \\
C3 & video grounding & \ilogo{qwen}\qplus{}~\citep{qwen2026qwen37plus} & {\scriptsize\texttt{qwen3.7-plus}} \\
\addlinespace[3pt]
D1 & graded wrong answers & \ilogo{gemini}\gflashgen{} & {\scriptsize\texttt{gemini-3-flash-preview}} \\
D3 & ranking without the video & \ilogo{gemini}\gpro{} & {\scriptsize\texttt{gemini-3.1-pro-preview}} \\
   &  & \ilogo{openai}\gptfull{} & {\scriptsize\texttt{gpt-5.4}} \\
D4 & ranking with the video & \ilogo{qwen}\qplus{} & {\scriptsize\texttt{qwen3.7-plus}} \\
\bottomrule
\end{tabular}

\vspace{4pt}
\begin{minipage}{0.92\linewidth}\footnotesize
\texttt{qwen3.7-plus} resolves to the snapshot \texttt{qwen3.7-plus-2026-05-26}. Blocks C1 and D2 are rule-based and call no AI model.
\end{minipage}
\end{table}

\begin{table}[H]
\centering
\small
\setlength{\tabcolsep}{5pt}
\caption{\textbf{Inputs to the data pipeline.} A clip is a single segment of at most 30 minutes, and
each question is built from two clips.}
\label{tab:app-datagen-inputs}
\begin{tabular}{@{}lll@{}}
\toprule
Blocks & Setting & Value \\
\midrule
B, D1 & video & the two segments as native video, each introduced by a text marker \\
      & sampling rate & 0.5 frames per second \\
      & media resolution & medium \\
      & transcript (B only) & the transcript of both clips, appended to the prompt \\
\addlinespace[3pt]
C2 & transcript windows & each segment's 30-minute transcript plus one window on either side: \\
   &  & three per segment, six in total, in an order shuffled per item\\
   & source of the extra windows & a different video of the same playlist \\
C2, D3 & parametric input & text only: the question (C2), the shuffled answers (D3) \\
\addlinespace[3pt]
C3, D4 & video & each segment's full 30-minute source video, audio removed \\
       & frame size & resized to fit $640\times480$, aspect ratio kept \\
       & sampling rate & 0.5 frames per second, from a file stored at 2 frames per second \\
       & transcript & the segment's full 30-minute transcript, after its video \\
\addlinespace[3pt]
all & response format & JSON under a schema (Gemini), strict JSON schema (GPT, Qwen) \\
\bottomrule
\end{tabular}
\end{table}

\begin{table}[H]
\centering
\small
\setlength{\tabcolsep}{5pt}
\caption{
\textbf{Acceptance rule for each gate} of Figure~\ref{fig:pipeline}. 
The gates run in the order listed, 
so an item reaches the costly video steps only after it has passed the cheaper ones.
}
\label{tab:app-datagen-gates}
\begin{tabular}{@{}lp{0.8\linewidth}@{}}
\toprule
Block & An item is kept when \\
\midrule
C1 & the answer has at least 3 paragraphs and its citations, 
mapped onto 30-second chunks, cover at least 2 chunks per segment and 4 to 30 chunks in total (2 to 15 minutes of evidence). \\
C2 & \emph{transcript test:} the verifier judges the transcript-only answer \emph{not} to match the gold answer. \\
   & \emph{parametric test:} at least one of the two combinations judges the memory-only answer not to match. Combination B runs only for items that combination A flagged. \\
C3 & both verdicts are \emph{Yes}: the question is answerable from the videos, and the gold answer is grounded in them. \\
\addlinespace[3pt]
D2 & each of the four degraded answers carries at least one attribute-degradation entry. \\
D3 & at least one of the two judges fails to recover the intended order without the video. The
     second judge runs only for sets that the first one ordered correctly. \\
D4 & the judge that watches the video reproduces the intended order exactly. \\
\addlinespace[3pt]
E  & A rejected item is regenerated with the rejection reason, for at most 6 attempts per item in
     each phase and seed video segment pair. \\
\bottomrule
\end{tabular}
\end{table}

\subsection{Benchmark statistics}
\label{app:data-stats}

Table~\ref{tab:app-data-stats} summarizes the resulting benchmark produced by the
pipeline of Figure~\ref{fig:pipeline}, reporting the number of playlists processed,
the questions and pairs retained, and the length distributions of the questions
and of the two answers in each pair.

\begin{table}[t]
\centering
\small
\setlength{\tabcolsep}{5pt}
\caption{
Benchmark statistics. 
}
\label{tab:app-data-stats}
\begin{tabular}{@{}lr@{}}
\toprule
Statistic & Value \\
\midrule
\multicolumn{2}{@{}l}{\emph{Corpus}} \\
Domains & 7 \\
Playlists & 29 \\
Source videos & 457 \\
\addlinespace[3pt]
\multicolumn{2}{@{}l}{\emph{Benchmark}} \\
Accepted Questions & 376 \\
Sampled Questions (before pair sampling) & 343 \\
Sampled Questions (after pair sampling) & 327 \\
Preference pairs & 630 (90 per domain $\times$ 7 domains) \\
\addlinespace[3pt]
\multicolumn{2}{@{}l}{\emph{Length in words (mean / max)}} \\
Question & 64.0 / 127 \\
Gold answer & 234.6 / 477 \\
Degraded answers & 245.1 / 359 \\
\addlinespace[3pt]
\multicolumn{2}{@{}l}{\emph{Per pair}} \\
Chosen minus rejected, words & $-$2.2 \\
Longer answer wins (\%) & 41.7 \\
\bottomrule
\end{tabular}
\end{table}

\subsection{Domain descriptions}
\label{app:domains}

Table~\ref{tab:domains} lists the seven domains in \bench{} with representative topics and their knowledge type. \emph{Static} domains test largely fixed factual knowledge (\emph{e.g.}, historical events, species biology), \emph{dynamic} domains test evolving narrative or personal content (\emph{e.g.}, TV plot lines, daily vlogs), and \emph{mixed} domains contain both.

\begin{table}[H]
\centering
\small
\caption{Domains in \bench{} with example topics and knowledge type.}
\label{tab:domains}
\begin{tabular}{lll}
\toprule
Domain & Example topics & Knowledge \\
\midrule
Education    & Academic lectures, software tutorials        & Static  \\
History      & Ancient civilizations, military strategies    & Static  \\
Art          & Painter biographies, classical music          & Static  \\
Documentary  & Marine wildlife, space exploration            & Static  \\
Drama        & Scripted TV series                            & Dynamic \\
Life         & Cooking, daily vlogs                          & Dynamic \\
Podcasts     & Interviews, sports analysis                   & Mixed   \\
\bottomrule
\end{tabular}
\end{table}

\subsection{A complete~\dataset{} item}
\label{app:full-item}

\begin{figure}[p]
\centering
\includegraphics[width=\textwidth]{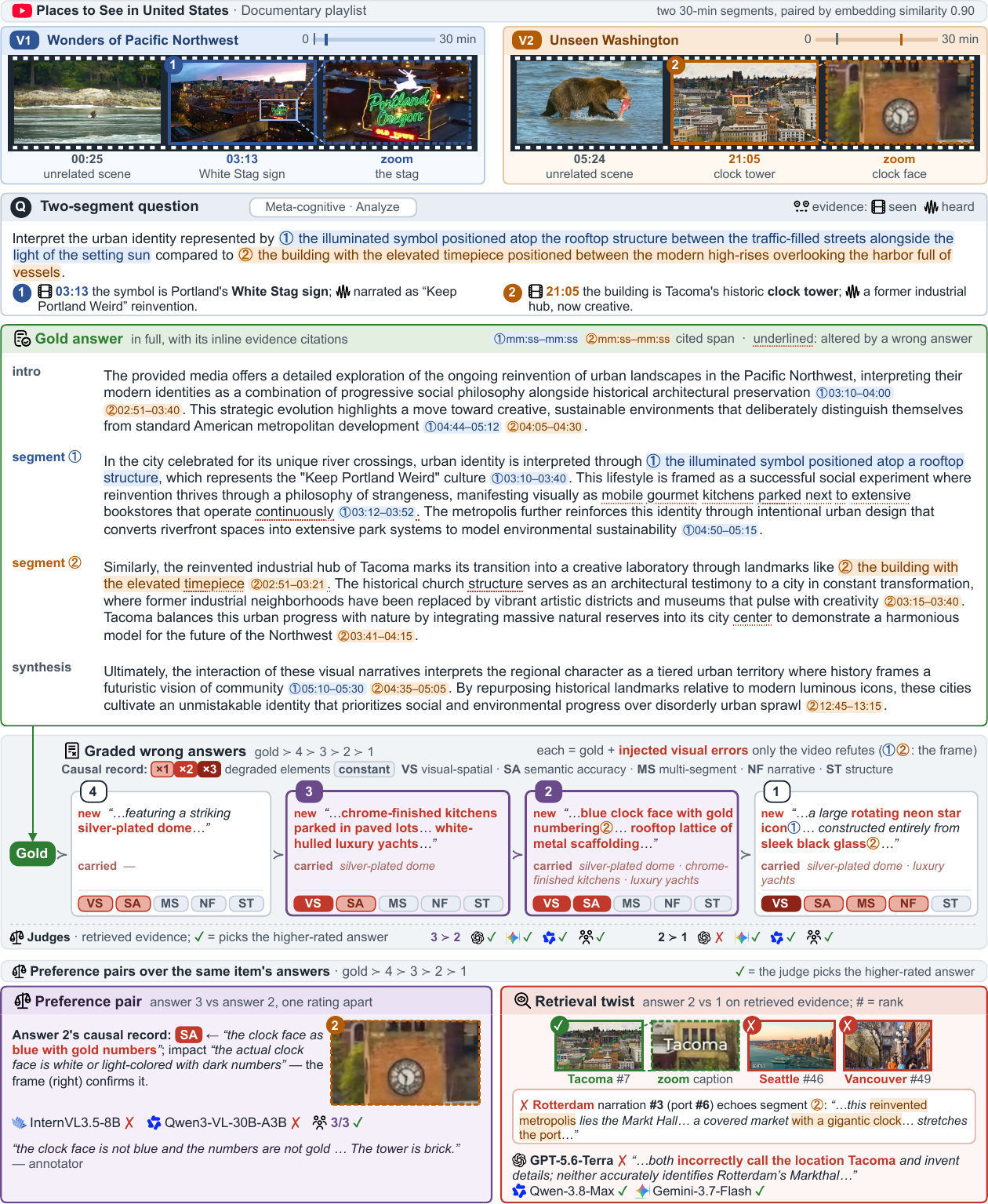}
\caption{\textbf{A detailed \dataset{} example,} the item of Figure~\ref{fig:example} in full. 
Two 30-minute segments of a Documentary playlist, paired by embedding similarity, each supply one of the question's two entities,
with the gold answer, its four graded wrong answers, and 
two preference pairs scored over them.}
\label{fig:example-full}
\end{figure}

Figure~\ref{fig:example-full} shows the item of Figure~\ref{fig:example} in full. In the gold
answer, every claim cites its supporting span inline as a chip giving the segment (\circled{1} or
\circled{2}) and the time range, and a dotted underline marks every span that some wrong answer
alters, found by diffing the gold against each rating. 

Each graded wrong answer is the gold with
visual errors injected (\emph{new}) or inherited from the answer above (\emph{carried}); its pills
show the attributes its causal record degrades, darker for more elements. Two pairs run over
these answers: on answers 3 vs.\ 2 both difficulty-gate VLMs fail while all three annotators prefer
answer 3, and on answers 2 vs.\ 1 with retrieved evidence, a Rotterdam narration that echoes
segment~\circled{2} outranks the true tower and misleads \gptt{}.

\section{Additional Results and Analysis}

\subsection{Position bias}
\label{app:position-bias}

Running the benchmark twice with the two answers' A/B sides exchanged, strong judges track the answer rather than the slot: when the sides swap, they switch letters accordingly, giving low letter agreement ($14.8\%$ for \qmax{}, $21.6\%$ for \gflash{}). Weaker judges instead tend to pick the same slot regardless of content, with letter agreement rising to $47.5\%$ for \glite{} and $49.1\%$ for \gemmamoe{} (Table~\ref{tab:e09-01-position-bias}). Position bias thus tracks judge quality and is another axis on which the weaker models prove unreliable.

\begin{table}[t]
\centering
\small
\setlength{\tabcolsep}{3pt}
\caption{
Sensitivity to answer position by the judge models.
}
\label{tab:e09-01-position-bias}
\begin{tabular}{lccccc}
\toprule
 & \multicolumn{3}{c}{\emph{Consistent ($<25\%$)}} & \multicolumn{2}{c}{\emph{Inconsistent ($\geq25\%$)}} \\
\cmidrule(lr){2-4}\cmidrule(lr){5-6}
 & \ilogo{qwen}Qwen & \ilogo{gemini}Gemini & \ilogo{qwen}Qwen & \ilogo{gemini}Gemini & \ilogo{gemma}Gemma \\
 & 3.8-Max & 3.7-Flash & 3.7-Flash & 3.5-Flash-Lite & 4-26B-A4B \\
\midrule
Input & \icoV\,\icoT & \icoV\,\icoA & \icoV\,\icoT & \icoV\,\icoA & \icoV\,\icoT \\
Accuracy, forward (\%) & 74.4 & 75.4 & 67.9 & 59.5 & 59.6 \\
Accuracy, reversed (\%) & 75.9 & 76.7 & 68.7 & 62.5 & 59.8 \\
Picked A, forward (\%) & 49.5 & 57.9 & 48.5 & 67.8 & 70.6 \\
Picked A, reversed (\%) & 58.8 & 61.4 & 52.3 & 70.8 & 68.4 \\
\textbf{Letter agreement} (\%) & 14.8 & 21.6 & 22.3 & 47.5 & 49.1 \\
Solved in both runs (\%) & 67.8 & 65.2 & 57.2 & 37.3 & 35.1 \\
Wrong in both runs (\%) & 17.4 & 13.2 & 20.5 & 15.2 & 15.7 \\
\bottomrule
\end{tabular}
\vspace*{-0.6cm}
\end{table}

\subsection{Effect of agentic tool use}
\label{app:agentic}

At the oracle scope of Figure~\ref{fig:scope-ladder}, where the judge receives only the two gold segments, \glite{} with agentic tool use~\citep{agenticvision2026gemini}, which fetches its own evidence from the video instead of receiving a fixed sample of frames, reaches $65.9\%$. That is only $2.0$ points above the same model with fixed frames at the same scope ($63.9\%$), and $27.1$ points below human agreement ($93.0\%$). Even without retrieval, letting the judge search for itself does not close the gap: finding and reading the deciding moment within a 10--30 minute segment remains hard.

\subsection{Correlation with long-video understanding}
\label{app:lvbench}

\begin{figure}[H]
\centering
\includegraphics[height=2.05in]{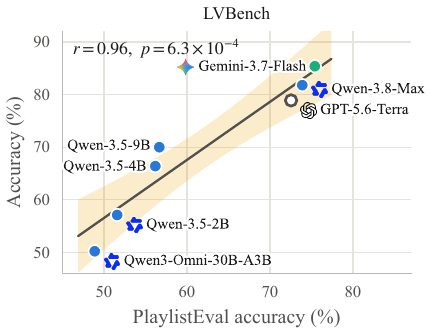}
\caption{
Accuracy on~\bench{} vs. accuracy on LVBench.
}
\label{fig:downstream-lvbench}
\end{figure}

A judge's accuracy on~\bench{} tracks its long-video question answering ability. 
For the seven judges with a published LVBench score~\citep{wang2025lvbench}, the two accuracies are strongly correlated (Pearson $r = 0.96$, $p = 6.3\times10^{-4}$; Figure~\ref{fig:downstream-lvbench}). This suggests that our fully automated generation pipeline recovers the same model ranking as human-annotated benchmarks, while remaining far from saturated (best judge system $75.4\%$ vs.\ $93.0\%$ human).

\subsection{Retrieval by segment}
\label{app:hit-breakdown}

\begin{table}[H]
\centering
\small
\setlength{\tabcolsep}{3pt}
\setlength{\ULdepth}{1.2pt}
\caption{\textbf{Hit@10 by segment}: the breakdown behind the Hit@10 column of Table~\ref{tab:e04-02-retrievers}; the \emph{both}, \emph{either}, and per-segment rates are defined in Appendix~\ref{app:retrieval-metrics}. All values in \% over the 343 questions. Best per column in \textbf{bold}, second best \protect\dashuline{dashed}. Glyphs: \icoV\ video frames, \icoT\ transcript, \icoA\ audio track.}
\label{tab:app-hit-breakdown}
\footnotesize
\begin{tabular}{llcccc}
\toprule
Retriever & Input & Both segments & Either segment & Segment 1 & Segment 2 \\
\midrule
\rlogo{qwen}\qwembeight{} & \icoV\,\icoT & \textbf{37.9} & \textbf{79.0} & \textbf{56.9} & \dashuline{60.1} \\
 & \icoV & 32.9 & 70.3 & 51.6 & 51.6 \\
 & \icoT & 30.0 & 69.7 & 46.4 & 53.4 \\
\addlinespace[3pt]
\rlogo{wechat}\wemm{} & \icoV\,\icoT & \dashuline{37.6} & \dashuline{76.7} & \dashuline{52.5} & \textbf{61.8} \\
\addlinespace[3pt]
\rlogo{qwen}\qwembtwo{} & \icoV\,\icoT & 32.7 & 72.0 & 47.5 & 57.1 \\
\addlinespace[3pt]
\rlogo{nvidia}\nemo{} & \icoV\,\icoT & 30.6 & 74.9 & 51.3 & 54.2 \\
 & \icoV\,\icoA & 26.2 & 67.9 & 47.5 & 46.6 \\
 & \icoA & \phantom{0}7.3 & 37.3 & 22.2 & 22.4 \\
\bottomrule
\end{tabular}
\end{table}

Table~\ref{tab:app-hit-breakdown} splits the Hit@10 of Table~\ref{tab:e04-02-retrievers} into the rate at which each segment is retrieved on its own, and the rate at which at least one of the two arrives. The either-segment rate is the number usually reported as Hit@10; both segments together, which is what a two-segment question needs, is roughly half of it.

\subsection{Frame budget and embedding width of the retriever}
\label{app:retrieval-budget}

\begin{figure}[H]
\centering
\includegraphics[width=\textwidth]{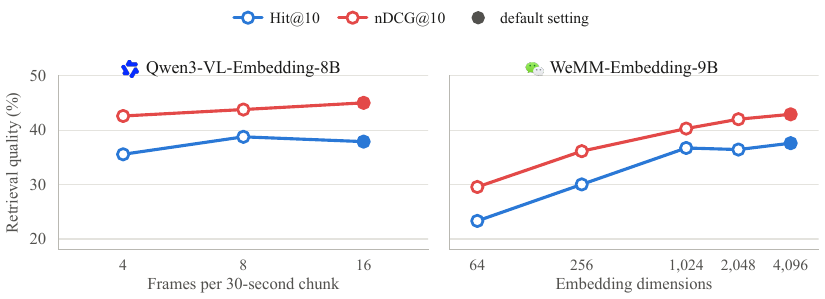}
\caption{\textbf{Frame budget and embedding width of the retriever.} \emph{Left:} \qwembeight{}
embeds each 30-second chunk from 4, 8, or 16 frames (16 is the default). \emph{Right:} \wemm{}
truncated along its matryoshka dimensions from 4,096 down to 64. Filled markers are each
retriever's default; every point retrieves over the question's whole domain for the same 343
questions (Hit@10 and nDCG@10 as defined in Appendix~\ref{app:retrieval-metrics}).}
\label{fig:app-retrieval-budget}
\end{figure}

Figure~\ref{fig:app-retrieval-budget} varies two costs of the retriever while holding the rest of
the setup fixed. Embedding a chunk from more frames barely helps: going from 4 to 16 frames moves
Hit@10 from 35.6 to 37.9 (it peaks at 38.8 with 8 frames) and nDCG@10 from 42.6 to 45.0. The
embedding width can be cut much further before it matters. Truncating \wemm{} from 4,096 to 1,024
dimensions costs 0.9 points of Hit@10 at a quarter of the storage; below that the loss grows fast,
and at 64 dimensions Hit@10 has fallen by 14.3 points, to 23.3. The full truncation shrinks the
float32 vector stored per chunk from 16\,KB to 256\,B.

\subsection{Pixel budget and audio speed of the judge}
\label{app:pixel-local}
\label{app:judge-budget}

\begin{table}[H]
\begin{minipage}[t]{0.49\linewidth}
\renewenvironment{table}[1][]{}{}%
\begin{table}[t]
\centering
\small
\setlength{\tabcolsep}{3pt}
\caption{
Effect of judge's image resolution.
}
\label{tab:e07-03-image-detail-media-resolution}
\footnotesize
\begin{tabular}{lcccc}
\toprule
 & & \multicolumn{2}{c}{Tokens} & \\
\cmidrule(lr){3-4}
Setting & Accuracy & Prompt & Output & USD \\
\midrule
low (base) & 59.5 \phantom{($-$1.1)} & \phantom{0}84k & 1,447 & \phantom{0}9.11 \\
medium & 58.4 ($-$1.1) & 119k & 1,514 & 12.48 \\
high & \textbf{60.8} ($+$1.3) & 186k & 1,528 & 18.83 \\
\bottomrule
\end{tabular}
\end{table}

\end{minipage}\hfill
\begin{minipage}[t]{0.49\linewidth}
\renewenvironment{table}[1][]{}{}%
\begin{table}[t]
\centering
\small
\setlength{\tabcolsep}{3pt}
\caption{
Effect of judge's audio playback speed.
}
\label{tab:e07-04-audio-speed}
\footnotesize
\begin{tabular}{lcccc}
\toprule
 & & \multicolumn{2}{c}{Tokens} & \\
\cmidrule(lr){3-4}
Setting & Accuracy & Prompt & Output & USD \\
\midrule
1.0$\times$ (base) & 59.5 \phantom{($+$1.1)} & 84k & 1,447 & 9.11 \\
1.5$\times$ & 60.6 ($+$1.1) & 68k & 1,445 & 7.60 \\
2.0$\times$ & \textbf{61.1} ($+$1.6) & 60k & 1,441 & 6.84 \\
\bottomrule
\end{tabular}
\end{table}

\end{minipage}
\end{table}

\begin{wraptable}{r}{0.46\linewidth}
\vspace*{-0.4cm}
\vspace{-\baselineskip}
\centering
\small
\setlength{\tabcolsep}{4pt}
\caption{\textbf{Pixel-budget on Gemma-4-26B-A4B}. Accuracy (\%), with the paired difference from the low default in parentheses (percentage points). Best accuracy in \textbf{bold}.}
\label{tab:e07-03-image-detail-media-resolution-local}
\begin{tabular}{lcc}
\toprule
Setting & Max soft tokens & Accuracy \\
\midrule
low (base) & \phantom{0,}280 & \textbf{59.7} \phantom{($-$1.8)} \\
medium & \phantom{0,}560 & 57.8 ($-$1.8) \\
high & 1,120 & 57.6 ($-$2.1) \\
\bottomrule
\end{tabular}
\vspace{-0.5\baselineskip}
\end{wraptable}

Table~\ref{tab:e07-03-image-detail-media-resolution-local} repeats the pixel-budget sweep of Table~\ref{tab:e07-03-image-detail-media-resolution} on \gemmamoe{}, which we run on our own GPUs (NVIDIA H200). It pools every frame to the same number of visual tokens (Table~\ref{tab:app-judge-inputs}), so a larger budget only sharpens what it sees before pooling; its accuracy does not improve. 
\WFclear %

\section{Failure Cases}
\label{app:failures}

\begin{figure}[!ht]
\centering
\includegraphics[width=\textwidth]{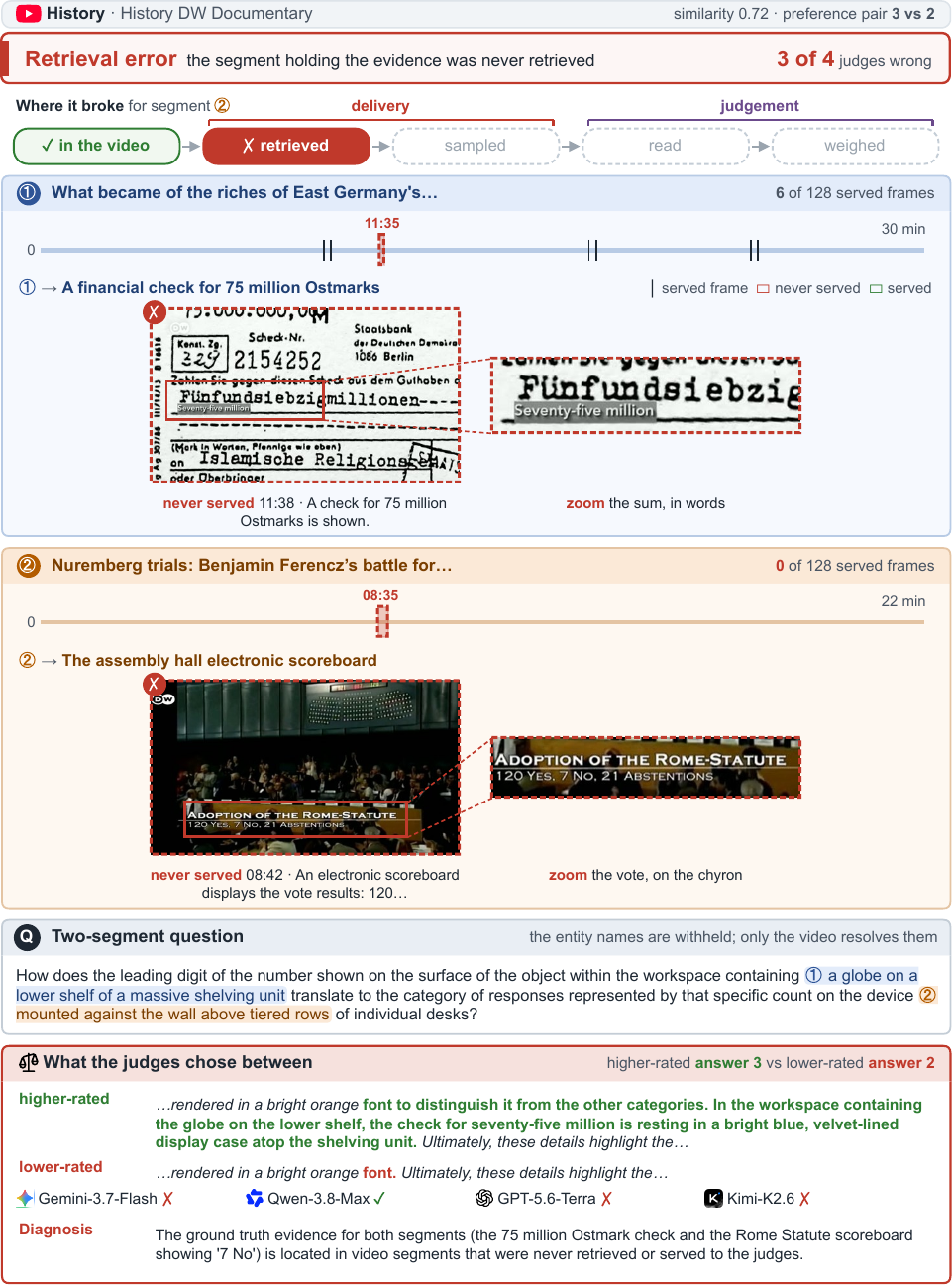}
\caption{\textbf{Retrieval error.} A History pair, answers 3 versus 2, that three of the four
judges lost. The question bridges the leading 7 of a 75-million-Ostmark cheque in segment~\circled{1}
to the 7 \emph{no} votes on the Rome Statute scoreboard in segment~\circled{2}. Segment~\circled{2}
was never retrieved, and none of the six frames served from segment~\circled{1} lands on the cheque,
so neither number reached the judges; only \qmax{} chose the higher-rated answer.}
\label{fig:failure-retrieval}
\end{figure}

\begin{figure}[p]
\centering
\includegraphics[width=\textwidth]{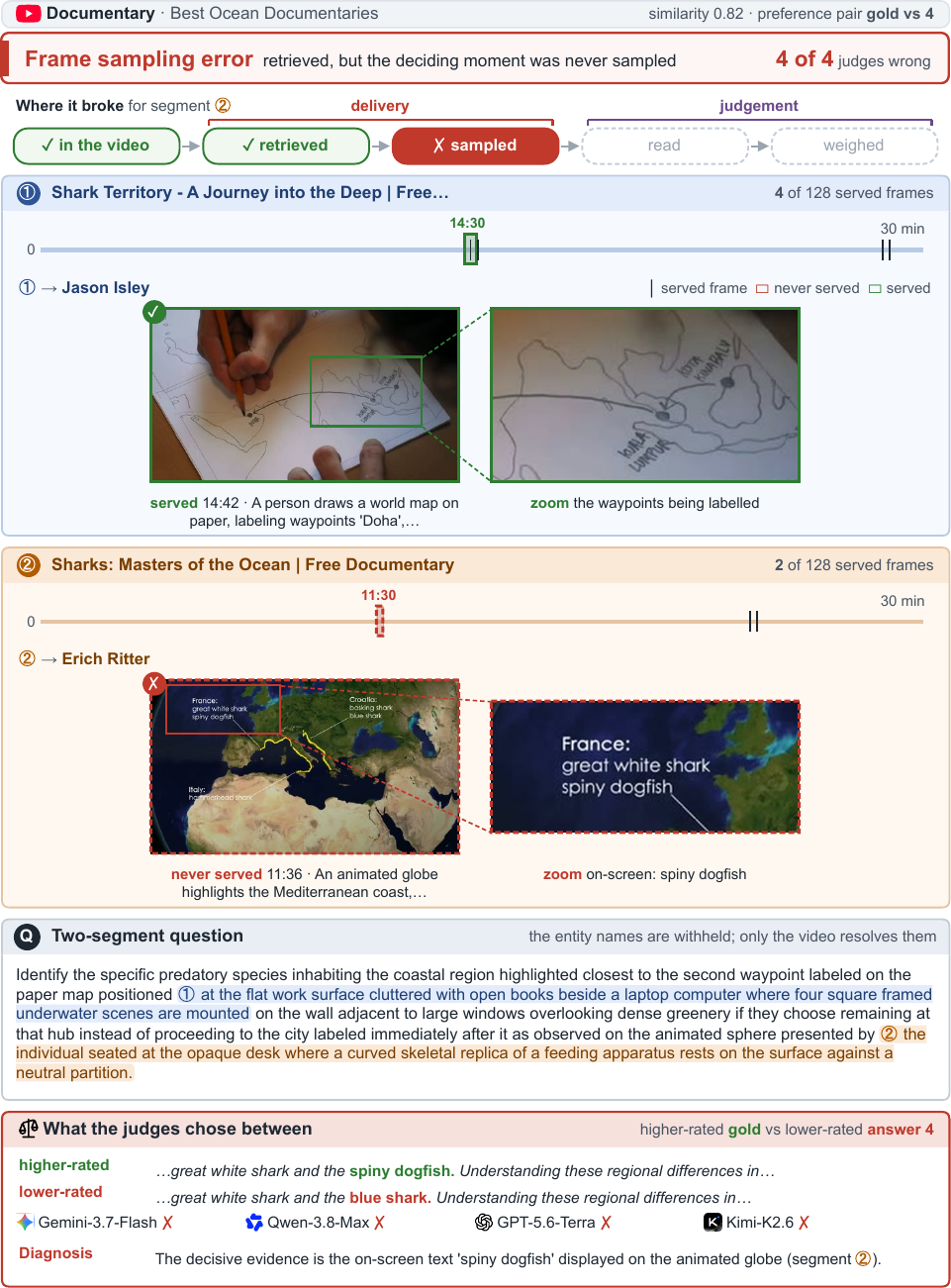}
\caption{\textbf{Frame sampling error.} A Documentary pair, the gold answer versus answer 4,
that all four judges lost. Both segments were retrieved, but none of the 128 sampled frames lands
on the deciding moment --- the on-screen label \emph{spiny dogfish} at 11:36 in segment~\circled{2}
--- so the judges fell back on the transcript, which names a different species in an unrelated passage.}
\label{fig:failure-sampling}
\end{figure}

\begin{figure}[p]
\centering
\includegraphics[width=\textwidth]{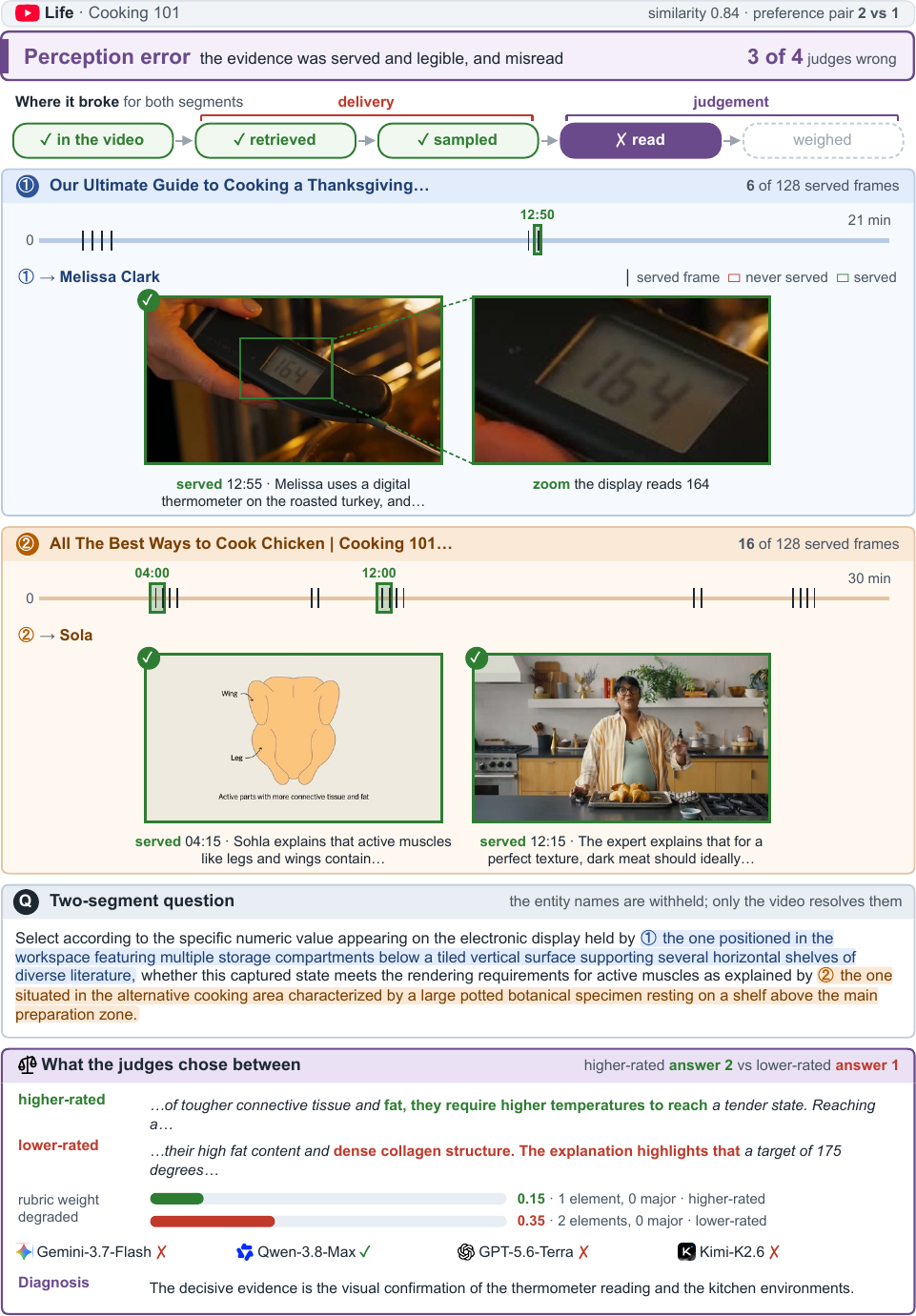}
\caption{\textbf{Perception error.} A Life pair, answers 2 versus 1, that three of the four
judges lost. The deciding evidence is present and legible --- a served frame at 12:55 shows a
handheld probe thermometer reading 164 over the roasted bird --- yet the judges misread the
display, and three preferred the answer whose account of it is wrong.}
\label{fig:failure-perception}
\end{figure}

\begin{figure}[p]
\centering
\includegraphics[width=\textwidth]{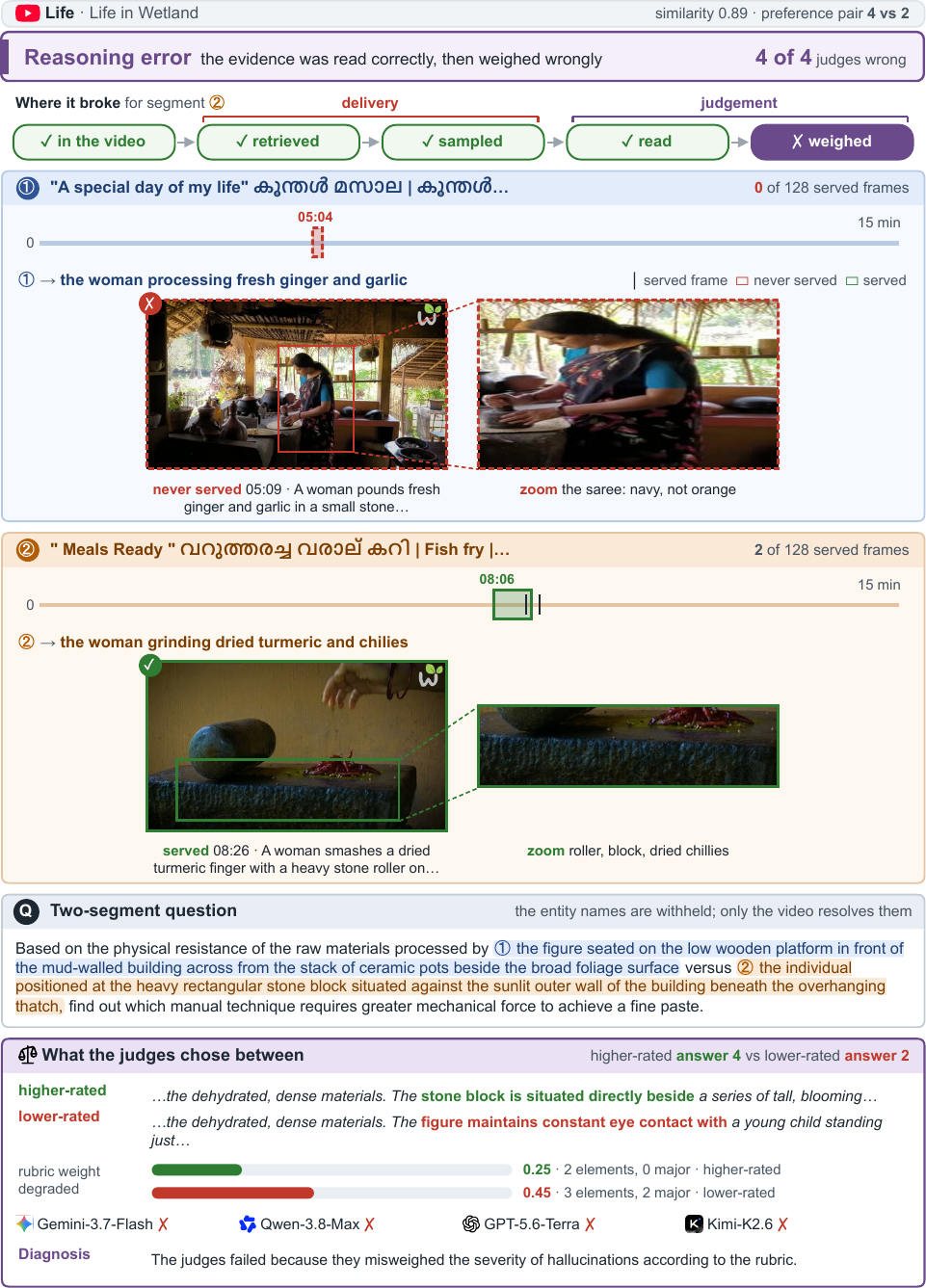}
\caption{\textbf{Reasoning error.} A Life pair, answers 4 versus 2, that all four judges lost.
A served frame lands on the grinding sequence in segment~\circled{2}, which shows neither the purple
flowering plant nor the child that the lower-rated answer describes --- its two major fabrications.
By the generator's own rubric that answer is degraded on 0.45 of the weight and the higher-rated one
on 0.25 with no major element, yet all four judges preferred the lower-rated answer.}
\label{fig:failure-reasoning}
\end{figure}

Section~\ref{sec:error_analysis} counts how often each kind of error decides a pair.
Figures~\ref{fig:failure-retrieval} to~\ref{fig:failure-reasoning} show one pair of each kind in full. Each panel shares the following layout.

The header names the domain, playlist, segment similarity, and the pair's rating gap. A banner below it labels the error type and how many of the four strongest judges it affected. A five-step chain (\emph{in the video}, \emph{retrieved}, \emph{sampled}, \emph{read}, \emph{weighed}) marks the first step at which the deciding evidence was lost: retrieval and sampling are delivery steps, while reading and weighing are the judge's own reasoning.

The two segment blocks each show a timeline with ticks for every served frame, the deciding moment highlighted in green (served) or red (missed), the entity each segment identifies, and zoomed thumbnails of the key evidence. Below them, the two-segment question is shown with entity names withheld, as the judge must resolve them from the video alone.

The bottom panel shows the span where the two answers differ over the deciding evidence, 
the four judges' verdicts, and a diagnosis of the failure.

\clearpage

\section{Cost Analysis}

As long-video input is quite costly, we analyze the cost of our data generation process with details of each pipeline stage including the retries.

\subsection{Data generation cost}
\label{app:datagen-cost}

Building the benchmark cost \$352.45 in API calls, or \$0.94 per accepted question.
That figure is the whole bill divided by the questions that survived: retries and the
candidates a gate rejected are paid for too, and are included here.
Table~\ref{tab:app-cost-domain} breaks it down by domain and
Table~\ref{tab:app-cost-model} by model. Two models carry almost all of it, and both
are the ones that read native video: question and wrong-answer generation with
\gflashgen{}, and the video-grounded checks with \qplus{}. 
The text-only verifiers,
which see a transcript or nothing at all, together account for under a tenth.

In Table~\ref{tab:app-cost-domain}, \emph{pooled} divides a domain's spend by its accepted
questions, while \emph{mean} averages the per-question cost across its runs (batches), counting a small run
the same as a large one. In Table~\ref{tab:app-cost-model}, \emph{tokens} counts prompt and output
together and \emph{per question} divides a model's spend by all 376 accepted questions; the rounded
per-model figures sum to a cent below the \$352.45 total.

\begin{table}[t]
\centering
\small
\setlength{\tabcolsep}{5pt}
\caption{\textbf{Cost to generate a question, by domain.} \emph{Spend} is the domain's total API cost, and \emph{pooled} and \emph{mean} are two per-question averages of it (defined in the text). All figures are in US dollars.}
\label{tab:app-cost-domain}
\begin{tabular}{lrrrrcc}
\toprule
 & & & & \multicolumn{3}{c}{Per question (\$)} \\
\cmidrule(lr){5-7}
Domain & Runs & Questions & Spend (\$) & Pooled & Mean $\pm$ sd & Range \\
\midrule
Art & 5 & 56 & 57.37 & 1.02 & 1.05 $\pm$ 0.16 & 0.80--1.22 \\
Documentary & 4 & 47 & 47.04 & 1.00 & 1.02 $\pm$ 0.21 & 0.83--1.29 \\
Drama & 4 & 49 & 39.01 & 0.80 & 0.83 $\pm$ 0.13 & 0.71--1.00 \\
Education & 4 & 62 & 60.54 & 0.98 & 0.97 $\pm$ 0.09 & 0.90--1.09 \\
History & 5 & 47 & 49.25 & 1.05 & 1.17 $\pm$ 0.30 & 0.75--1.54 \\
Life & 4 & 65 & 46.20 & 0.71 & 0.64 $\pm$ 0.13 & 0.52--0.79 \\
Podcast & 4 & 50 & 53.04 & 1.06 & 1.10 $\pm$ 0.11 & 0.99--1.25 \\
\midrule
\textbf{All} & 30 & 376 & 352.45 & \textbf{0.94} &  &  \\
\bottomrule
\end{tabular}
\end{table}

\begin{table}[t]
\centering
\small
\setlength{\tabcolsep}{6pt}
\caption{\textbf{Cost to generate a question, by model.} \emph{Tokens} counts prompt and output together, and \emph{per question} divides a model's spend by all 376 accepted questions. All figures are in US dollars.}
\label{tab:app-cost-model}
\begin{tabular}{lrrrr}
\toprule
Model & Tokens & Spend (\$) & Share (\%) & Per question (\$) \\
\midrule
\ilogo{gemini}\gflashgen{} & 518M & 165.34 & 46.9 & 0.440 \\
\ilogo{qwen}\qplus{} & 189M & 154.75 & 43.9 & 0.412 \\
\ilogo{gemini}\gpro{} & \phantom{00}4M & 16.61 & 4.7 & 0.044 \\
\ilogo{openai}\gptfull{} & \phantom{00}3M & 12.84 & 3.6 & 0.034 \\
\ilogo{openai}\gptmini{} & \phantom{00}3M & 2.90 & 0.8 & 0.008 \\
\midrule
\nologo{}\textbf{All} & 717M & 352.44 & 100.0 & \textbf{0.937} \\
\bottomrule
\end{tabular}
\end{table}

\subsection{Downstream evaluation cost}
\label{app:eval-cost}

Scoring the benchmark once with all seventeen judges cost \$340.16:
\$245.39 for the hosted models and \$94.77 of GPU time, 23.7 NVIDIA H200 GPU-hours, for the ones we
run ourselves locally. Table~\ref{tab:app-cost-eval} gives it per judge, in the groups and order of
Table~\ref{tab:e01-main}, with each of the 17 judges scoring all 630 pairs on retrieved evidence.
Only this main run is counted here and each ablation is a further pass of the same kind.

\begin{table}[t]
\centering
\small
\setlength{\tabcolsep}{6pt}
\caption{\textbf{Cost of one pass over the benchmark,} per judge. A hosted judge is billed for the tokens it reads and writes, and a locally served judge for the GPU time it occupies (at \$4 per NVIDIA H200 GPU-hour), so each row fills one column or the other. 
Spend is in US dollars.
}
\label{tab:app-cost-eval}
\begin{tabular}{lrrr}
\toprule
Judge & Tokens & GPU-hours & Spend (\$) \\
\midrule
\multicolumn{4}{c}{\emph{Hosted API models (general-purpose)}} \\
\addlinespace[1pt]
\ilogo{gemini}\gflash{} & 53.9M & --- & 21.44 \\
\ilogo{gemini}\gpro{} & 67.5M & --- & 82.21 \\
\ilogo{gemini}\glite{} & 54.0M & --- & \phantom{0}9.11 \\
\addlinespace[3pt]
\ilogo{qwen}\qmax{} & 32.4M & --- & 33.92 \\
\ilogo{qwen}\qflash{} & 34.5M & --- & \phantom{0}4.32 \\
\addlinespace[3pt]
\ilogo{openai}\gptt{} & 23.5M & --- & 25.79 \\
\addlinespace[3pt]
\ilogo{kimi}\kimi{} & 46.2M & --- & 68.60 \\
\midrule
\multicolumn{4}{c}{\emph{Open-weight local models (general-purpose)}} \\
\addlinespace[1pt]
\ilogo{gemma}\gemmamoe{} & --- & 3.9 & 15.50 \\
\ilogo{gemma}\gemmaetwo{} & --- & 0.9 & \phantom{0}3.61 \\
\ilogo{gemma}\gemmaefour{} & --- & 0.4 & \phantom{0}1.69 \\
\addlinespace[3pt]
\ilogo{qwen}\qwennine{} & --- & 6.2 & 24.61 \\
\ilogo{qwen}\qwenfour{} & --- & 4.9 & 19.56 \\
\ilogo{qwen}\qwentwo{} & --- & 2.6 & 10.51 \\
\ilogo{qwen}\qomni{} & --- & 1.2 & \phantom{0}4.83 \\
\midrule
\multicolumn{4}{c}{\emph{Open-weight local models (fine-tuned as judges)}} \\
\addlinespace[1pt]
\ilogo{internlm}\ixc{} & --- & 0.6 & \phantom{0}2.55 \\
\addlinespace[3pt]
\ilogo{cmu}\cmuthree{} & --- & 0.2 & \phantom{0}0.97 \\
\ilogo{cmu}\cmuseven{} & --- & 2.7 & 10.92 \\
\midrule
\nologo{}\textbf{All} & 312.1M & 23.7 & \textbf{340.16} \\
\bottomrule
\end{tabular}
\end{table}

\section{Human evaluation details}
\label{app:human-eval}

To reduce the burden of watching long videos, we retrieve the key moments corresponding to each part of the question and each answer sentence using \qplus{} from both video segments in a preference pair; Appendix~\ref{app:prompt-ground} gives the prompt it is asked this with and the schema of its response. Moreover, we underline the differences between two answers in each preference pair to spot subtle changes in text, with the option to toggle by the annotators.

We have collected 87 hours of human annotation with median time spent 7 to 11 minutes across workers. The annotators received 8.23 USD per hour on average, which is higher than the federal minimum wage in the US.

We restrict the task to workers who meet four Amazon Mechanical Turk qualifications: a lifetime HIT approval rate of at least $95\%$, at least $1{,}000$ approved HITs, residence in a majority-English-speaking country (Australia, Canada, New Zealand, the United Kingdom, or the United States), and the Masters qualification granted by Amazon. To further improve the annotation quality, we include 10\% attention check samples, trivial questions with straightforward answers, and we discard the ratings of any worker who fails it.

We summarize the study in Table~\ref{tab:app-human-eval}.
The three annotators of a pair agree with
each other at $\alpha = 0.781$, and $93.0\%$ of the individual ratings fall on the answer
\bench{} intends to win. Taking the majority of a pair's ratings, $147$ of the $152$ pairs
side with the benchmark and five overturn it. Neither slot is favoured, so the
position preference the judges show (Table~\ref{tab:e09-01-position-bias}) is a property of the
judges rather than of the task.

\begin{table}[t]
\centering
\small
\setlength{\tabcolsep}{6pt}
\caption{\textbf{Summary of human evaluation.} Each pair was a forced A/B choice shown to three annotators, with the benchmark's intended answer placed in a random slot. The middle block reports agreement among the annotators, the last block agreement with the intended answer (per rating and per pair majority), and the final row the share of ratings per slot.}
\label{tab:app-human-eval}
\begin{tabular}{@{}lr@{}}
\toprule
Statistic & Value \\
\midrule
\multicolumn{2}{@{}l}{\emph{Study}} \\
Pairs & 152 \\
Valid ratings & 456 \\
Attention checks passed & 40/40 \\
\addlinespace[3pt]
\multicolumn{2}{@{}l}{\emph{Agreement between annotators}} \\
Krippendorff's $\alpha$~\citep{Krippendorff2011ComputingKA} & 0.781 \\
Unanimous pairs (3 of 3) & 126 \\
\addlinespace[3pt]
\multicolumn{2}{@{}l}{\emph{Agreement with the intended preference}} \\
Ratings that agree (\%) & 93.0 \\
Majority agrees & 147 (96.7\%) \\
Majority overturns & 5 (3.3\%) \\
\addlinespace[3pt]
\multicolumn{2}{@{}l}{\emph{Answer position}} \\
Preferred slot A\,/\,B (\%) & 49.8\,/\,50.2 \\
\bottomrule
\end{tabular}
\end{table}

\begin{figure}[p]
\centering
\fbox{\includegraphics[width=0.93\textwidth]{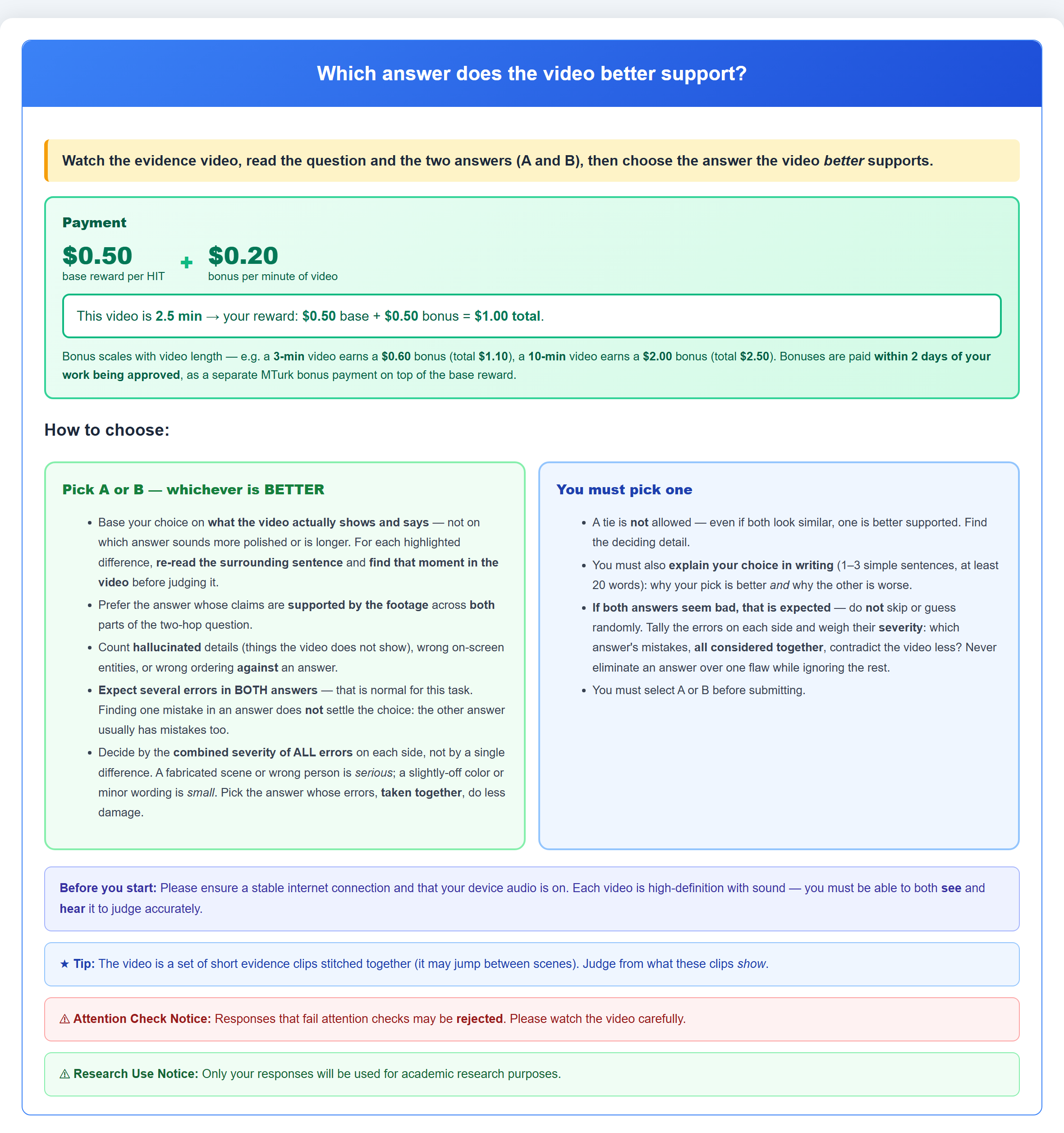}}
\caption{\textbf{Annotation task on Amazon Mechanical Turk, part 1 of 3:} what the worker is
asked to do, 
what it pays, 
and how to choose between the two answers.}
\label{fig:mturk-instructions}
\end{figure}

\begin{figure}[p]
\centering
\fbox{\includegraphics[width=0.93\textwidth]{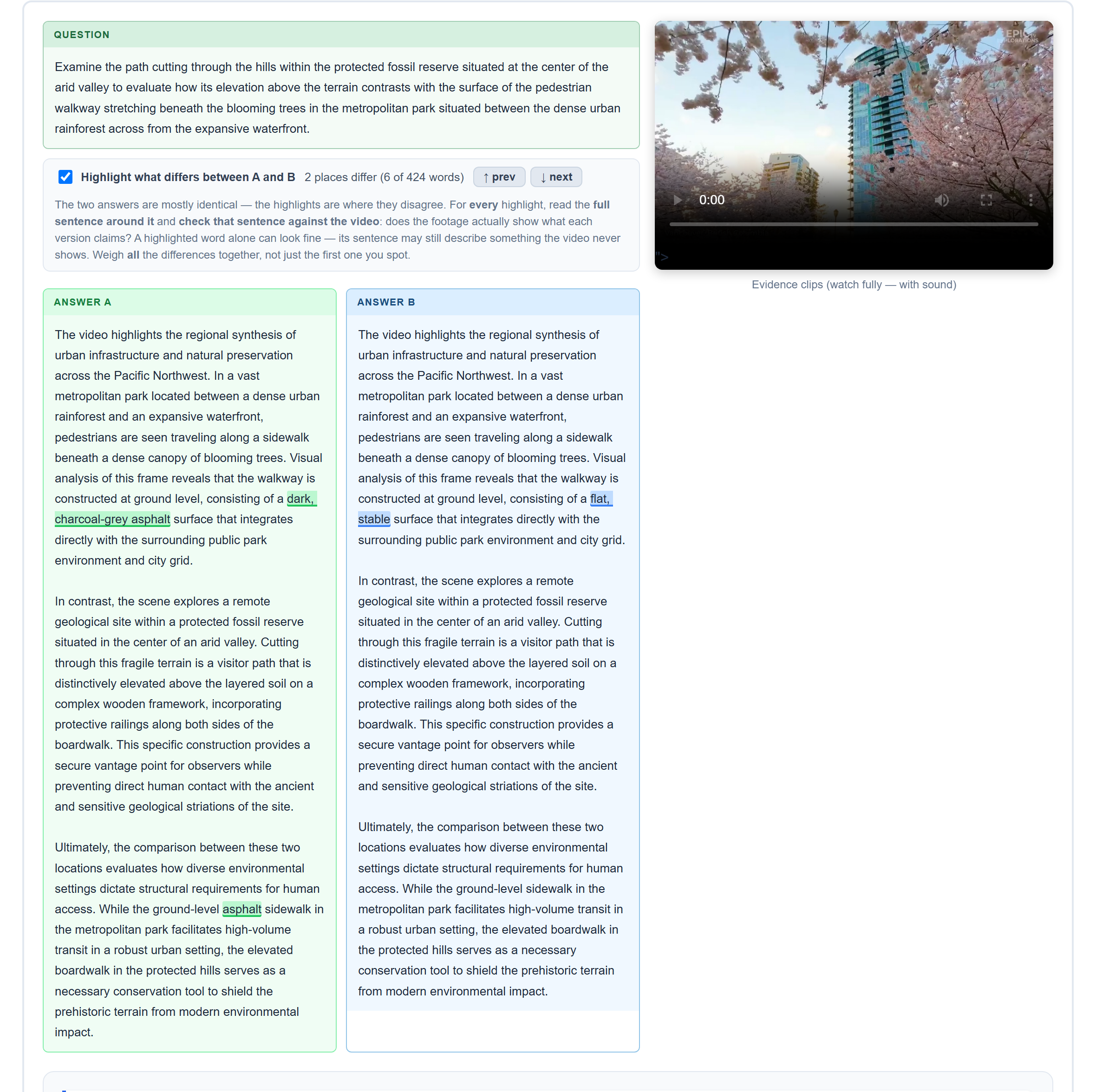}}
\caption{\textbf{Part 2 of 3:} the two-segment question with the evidence video beside it, and the two
answers. Differences between the answers are highlighted and can be stepped through, so a worker
compares only the few places where they disagree rather than re-reading both in full.}
\label{fig:mturk-task}
\end{figure}

\begin{figure}[p]
\centering
\fbox{\includegraphics[width=0.93\textwidth]{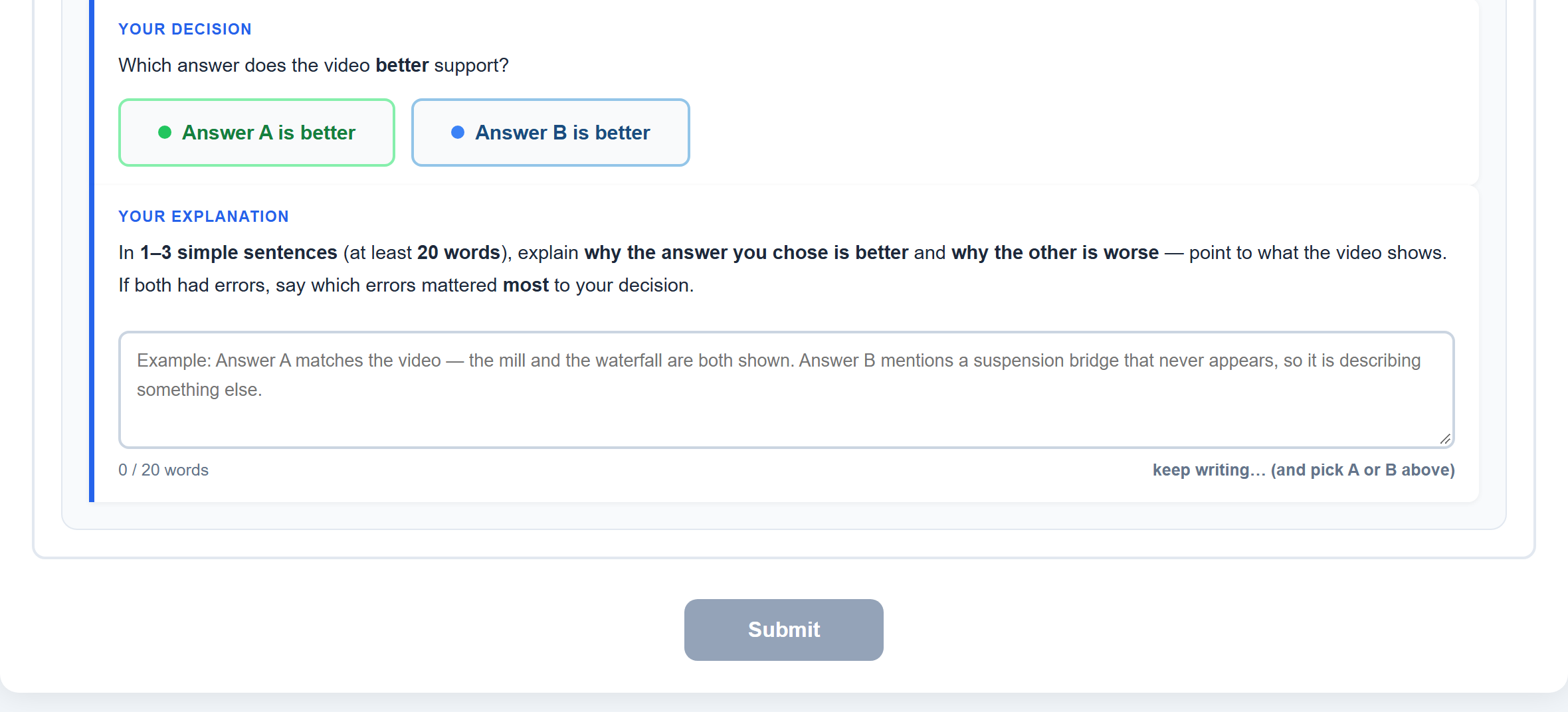}}
\caption{\textbf{Part 3 of 3:} the forced choice between the two answers, and the written
explanation of at least 20 words that a worker must give before submitting.}
\label{fig:mturk-decision}
\end{figure}

The task as a worker sees it is shown in Figures~\ref{fig:mturk-instructions},
\ref{fig:mturk-task} and~\ref{fig:mturk-decision}: the instructions and payment terms,
the question with its evidence video and the two answers, and the decision and written
explanation the worker submits.

\clearpage

\section{Prompts}

\subsection{Reward model evaluation prompt}
\label{app:rm_prompt}

Every judge in the meta-evaluation (block~G of Figure~\ref{fig:pipeline}) receives the prompt
below, and the difficulty gate (block~F) runs \internvleight{} and \qwenvlmoe{} on the same prompt.
Following~\citet{luo2025videoautoarena,lambert2025rewardbench}, the judge is shown two answers to one question, one ranked above the other in the intended order,
and must pick the better one; a tie is not accepted. The frames are attached to the request. The
transcript block is sent only when the judge input includes the transcript, as it does by default
for every judge that does not receive audio (Table~\ref{tab:app-judge-inputs}). The verdict is the
\texttt{[[A]]} or \texttt{[[B]]} in the reply. In the meta-evaluation, which answer appears as
Model~A is drawn once per pair with a shared seed, so every judge sees a given pair the same
way round, and a reply without a verdict, whether unparsed output or a provider's refusal, is
credited at chance.

\begin{promptbox}{Pairwise judging}
\pmeta{Blocks F and G \textperiodcentered{} \internvleight{} and \qwenvlmoe{} (F), every judge of
  Table~\ref{tab:app-judges} (G) \textperiodcentered{} input: frames, with the transcript or the
  audio of their segments}

You are an expert video understanding evaluator. You are shown frames sampled from a set of short
video segments retrieved from a video library --- they are NOT one clip per hop, they are not in a
guaranteed order of relevance, and some may be irrelevant to the question. Together they may or
may not contain what is needed to answer BOTH hops. You are given the two-hop question and two
candidate answers to it (from Model A and Model B). Act as an impartial judge and decide which
answer is better, grounding EVERY judgement in what the frames actually show and say --- not in
surface plausibility. If the frames do not show something an answer asserts, treat that assertion
as unsupported rather than assuming a segment you were not shown covers it.

\psec{Two-hop question:}
\texttt{<question>}\\
\pslot{question}\\
\texttt{</question>}

\begin{pcond}{sent only when the judge input includes the transcript}
\psec{Transcript (dialogue from the retrieved segments):}
\texttt{<transcript>}\\
\pslot{transcript}\\
\texttt{</transcript>}
\end{pcond}

\psec{Model A's answer:}
\texttt{<answer\_model\_a>}\\
\pslot{answer\_a}\\
\texttt{</answer\_model\_a>}

\psec{Model B's answer:}
\texttt{<answer\_model\_b>}\\
\pslot{answer\_b}\\
\texttt{</answer\_model\_b>}

Evaluate both answers against these standards:
\begin{pnum}
\item{} [Instruction Following]: The answer closely follows the question and directly addresses
  the specified two-hop task.
\item{} [Accuracy]: The answer uses the frames faithfully --- correct events, on-screen entities,
  and the order in which they appear across BOTH hops and the bridge between them; no
  hallucinated visual/audio details, no actions attributed to the wrong subject or phase;
  contextually coherent with precise terminology.
\item{} [Relevance]: The answer is comprehensive and on-topic, covering both hops and their
  connecting bridge without straying, and offering the detail the question calls for.
\item{} [Helpfulness]: The answer gives clear, valuable information that actually resolves the
  question, avoiding vague or irrelevant content.
\end{pnum}

\textbf{You MUST choose one answer. A tie is not an option.} Even when the two answers look similar
in quality --- both strong, both weak, or both partly unsupported by the frames --- one of them is
still better on the standards above. Find the discriminating detail and commit to it: a claim one
answer grounds in the frames and the other does not, a hop one covers and the other skips, a
hallucinated entity, a wrong ordering across the bridge. Do NOT say both are good, do NOT say
neither is good, and do NOT decline to choose.

Avoid any position biases and ensure that the order in which the answers were presented does not
influence your decision. Do not allow the length of an answer to influence your evaluation --- a
longer answer is not a better one, and extra detail that the frames do not support counts against
it, not for it. Be as objective as possible.

\textbf{Follow these steps for your judgement:}
\begin{plist}
\item Step 1: Analyze which answer is better on [Instruction Following].
\item Step 2: Analyze which answer is better on [Accuracy].
\item Step 3: Analyze which answer is better on [Relevance].
\item Step 4: Analyze which answer is better on [Helpfulness].
\item Step 5: From Steps 1-4, determine the overall winner. If the four standards are split, weigh
  [Accuracy] highest --- grounding in the frames is what this task is testing. The outcome is
  either Model A or Model B; there is no third option. Emit it as [[A]] or [[B]].
\end{plist}

Respond strictly in the following format:
\begin{promptcode}
```[Instruction Following]
[Your Analysis]
```
```[Accuracy]
[Your Analysis]
```
```[Relevance]
[Your Analysis]
```
```[Helpfulness]
[Your Analysis]
```
```[Overall Judge]
[[A]] if assistant A is better, [[B]] if assistant B is better.
```
\end{promptcode}

\pdivider{slot formats}
\pslot{answer\_a} and \pslot{answer\_b} are the two answers with their citations removed.

\pslot{transcript} has one line per segment, in the order of the frames; $i$ numbers the source
videos by first appearance, and \texttt{mm:ss} is where the segment starts in its video:
\begin{promptcode}
[Video i @ mm:ss] <transcript of the segment>
\end{promptcode}
\end{promptbox}

\subsection{Data generation prompts}
\label{app:data_prompt}

This section gives every prompt of the data pipeline, in the order of
Figure~\ref{fig:pipeline}, each followed by the schema of the response it must return.
Table~\ref{tab:app-prompt-stages} maps each block of the figure to its model and prompt. The
wording inside
each prompt box is the prompt as sent; Markdown emphasis, headings and bullets in the templates
are rendered as typography. Slots such as \pslot{question} are filled for each request, and
marked dividers show text added to a template at run time. Every call returns JSON constrained
by a response schema, passed to Gemini models as a JSON schema and to the GPT and Qwen models as
a strict JSON schema. Each schema box shows the typed field tree and then the field
descriptions verbatim, because the model reads those descriptions too, and for question
generation and wrong-answer generation they carry much of the specification.

\begin{table}[h]
\centering
\small
\setlength{\tabcolsep}{5pt}
\caption{\textbf{Steps of the data pipeline,} by block of Figure~\ref{fig:pipeline}, with the model
that runs each step and its prompt. Blocks C1 and D2 are rule-based and call no model.}
\label{tab:app-prompt-stages}
\begin{tabular}{@{}llll@{}}
\toprule
Block & Step & Model & Prompt \\
\midrule
B  & Two-segment QA generation & \gflashgen{} & \ref{app:prompt-s01} \\
\addlinespace[3pt]
C1 & Structural checks & --- & --- \\
C2 & Transcript test: transcript-only answer & \gflashgen{} & \ref{app:prompt-s03} \\
   & \quad verification of that answer & \gptmini{} & \ref{app:prompt-verify} \\
   & Parametric test: answer, combination A & \gflashgen{} & \ref{app:prompt-s05} \\
   & \quad verification, combination A & \gptfull{} & \ref{app:prompt-verify} \\
   & Parametric test: answer, combination B & \gptmini{} & \ref{app:prompt-s05} \\
   & \quad verification, combination B & \gpro{} & \ref{app:prompt-verify} \\
C3 & Video grounding & \qplus{} & \ref{app:prompt-s09} \\
\addlinespace[3pt]
D1 & Graded wrong answers & \gflashgen{} & \ref{app:prompt-s21} \\
D2 & Structural checks & --- & --- \\
D3 & Ranking without the video & \gpro{} and \gptfull{} & \ref{app:prompt-s23} \\
D4 & Ranking with the video & \qplus{} & \ref{app:prompt-s25} \\
\addlinespace[3pt]
E  & Regeneration feedback & appended to the B and D1 prompts & \ref{app:prompt-feedback} \\
\bottomrule
\end{tabular}
\end{table}

\subsubsection{Two-segment QA generation (B)}
\label{app:prompt-s01}

\gflashgen{} receives the two paired 30-minute clips as native video sampled at 0.5\,fps, each
introduced by a marker (\texttt{---Here is Clip 1:---}, \texttt{---Here is Clip 2:---}),
followed by the prompt below. The model answers in a structured format whose field descriptions
carry most of the question and answer specification; the schema box after the prompt gives them
in full. The keyword \pslot{question\_type\_word} is drawn per request, uniformly over
the Understand, Apply, Analyze and Evaluate columns of the knowledge dimension matrix.

\begin{promptbox}{Two-segment QA generation --- prompt}
\pmeta{Block B \textperiodcentered{} \gflashgen{} \textperiodcentered{} input: two clips as video}

\psec{Role}
You are an expert video analyst and educational content designer specializing in multi-hop
reasoning.

\psec{Task}
Generate one diverse, complex, two-hop question that require synthesizing information from two
distinct video segments. These video segments are the part of a large video database on a
specific topic.

\psec{Output Requirements}
\begin{plist}
\item \textbf{JSON Format:} Must be a strictly valid JSON object.
\item \textbf{Timestamps:} Provide \texttt{start\_time} and \texttt{end\_time} in seconds. This
  duration must contain necessary information to understand the question context and the answer
  to the corresponding question.
\item \textbf{Clip IDs:} Use 1-indexed video indices.
\end{plist}

\vspace{3pt}
{\footnotesize\setlength{\tabcolsep}{3pt}\renewcommand{\arraystretch}{1.1}
\begin{tabular}{@{}>{\raggedright\arraybackslash}p{0.17\linewidth}
                *{5}{>{\raggedright\arraybackslash}p{0.143\linewidth}}@{}}
\toprule
\textbf{The Knowledge Dimension} & \textbf{1 Remember} & \textbf{2 Understand} & \textbf{3 Apply}
  & \textbf{4 Analyze} & \textbf{5 Evaluate} \\
\midrule
\textbf{A Factual} & name, list, define, label & restate, order & state, determine
  & distinguish, classify & select according to \\
\textbf{B Conceptual} & identify, locate & describe, explain & illustrate, show
  & examine, analyze & rank, compare \\
\textbf{C Procedural} & tell, describe & summarize, translate & solve, demonstrate
  & deduct, diagram & conclude, choose \\
\textbf{D Meta Cognitive} & --- & interpret, paraphrase & find out, use
  & infer, examine & jutsify, judge \\
\bottomrule
\end{tabular}}
\vspace{3pt}

The keywords in the Knowledge Dimension Matrix indicate the level of complexity within a search
query. While simpler questions are located at the top left, more complex questions are
positioned on the bottom right of the table. Please formulate diverse and more complex questions
requiring multi-hop reasoning.

\psec{Contextual Identification Protocol}
\begin{plist}
\item \textbf{Zero Nominal Reference:} Do not use pronouns, names, or formal titles.
\item \textbf{Environmental Anchoring:} Refer to participants only by their position relative
  to fixed landmarks (e.g., ``the one positioned between the flickering lamp and the open
  doorway'').
\item \textbf{Attribute Exclusion:} Strictly avoid describing what an entity looks like, wears,
  is doing, is acting or is made of. The viewer must deduce ``who'' or ``what'' based solely on
  \emph{where} they are and within the shot. For example, you MUST avoid the \emph{COLOR}.
\item \textbf{Anti-Generic Detail:} Use hyper-specific environmental cues that exist only in
  this specific sequence to ensure the video is the only key to the description.
\item \textbf{Process:} Find a scene, observe the people, entities or objects in this scene.
  Describe/outline the scene information in the question stem.
  \begin{plist}
  \item Example Question stem: On the red wooden table, there is an iron grid rack with a glass
    bowl containing four rolls of food. In the frame, there is a brush covered with yellow liquid
    decorating them.
  \end{plist}
\end{plist}

\pdivider{appended at run time: the transcript of both clips}

\psec{Audio transcript of the clips (everything SPOKEN in the two videos)}
\pslot{transcript}

\psec{CRITICAL --- visual grounding requirement}
The generated question AND its answer MUST require WATCHING the video. The transcript above is
the complete spoken audio. Your question+answer must NOT be answerable from this transcript
alone, nor from general world knowledge --- the answer must depend on VISUAL details shown on
screen but NOT stated in the transcript (e.g.\ colours, spatial layout, gestures, on-screen
objects/text, who or what appears). Do not merely restate what is said.
\end{promptbox}

\begin{schemabox}{Two-segment QA generation --- response schema}
\pmeta{Structured output \texttt{Recipe}, enforced as a JSON schema; its field descriptions,
  below the tree, are read by the model together with the prompt}

\begin{promptcode}
Recipe
  qa_pairs : list of QA_Pair
    question         : string
    answer           : string
    question_type    : { row    : "Factual" | "Conceptual" | "Procedural" | "Meta Cognitive",
                         column : "Remember" | "Understand" | "Apply" | "Analyze" | "Evaluate" }
    bridge           : string
    hop_1            : string
    hop_2            : string
    scene_references : list of { scene_description : string, actual_entity_name : string }
    clip_ranges      : list of { clip_id : integer, start_time : integer,
                                 end_time : integer, qa_reference : string }
\end{promptcode}

\pdivider{field descriptions}

\pfield{qa\_pairs} List of QA pairs.
\begin{plist}
\item \textbf{Linguistic Architecture}
  \begin{plist}
  \item \textbf{Entity Diversity:} Each QA pair must focus on different objects, people, or
    concepts to ensure zero repetition.
  \end{plist}
\item \textbf{Unified Media Perspective}
  \begin{plist}
  \item \textbf{Single Entity Rule:} Treat all provided clips as one single video.
  \item \textbf{Terminology:} Use only ``the video,'' ``the scene,'' ``the frame,'' or ``the
    shot.''
  \item \textbf{Prohibited References:} Never use ``Video 1,'' ``first lecture,'' ``second
    clip,'' or any term implying the media is split.
  \end{plist}
\end{plist}

\psec{Each QA pair}

\pfield{question} Question stem sentence(s), followed by a multihop question.
\begin{plist}
\item \textbf{Forbidden Terms:} You must not use the word ``and'' in any question.
\item \textbf{Forbidden Syntax:} Avoid semicolons or comma-splices used to mimic the word
  ``and.''
\item \textbf{Question Specifications}
  \begin{plist}
  \item Draft as a single, grammatically correct sentence without the word ``and.''
  \end{plist}
\item \textbf{Temporal Distance:} The two segments used for each question must be at least 4
  minutes apart.
\item \textbf{The ``Two-Hop'' Logic:}
\item \textbf{Hop 1:} Extract a specific fact or concept from Video A.
\item \textbf{Bridge:} Connect that fact to a related concept in Video B.
\item \textbf{Hop 2:} Derive the final answer based on the interaction of both facts.
\item \textbf{Complexity:} Target the bottom-right of the \textbf{Knowledge Dimension Matrix}
  (Analyze, Evaluate, Meta-Cognitive).
\end{plist}
\textit{Multi-Hop QA Generation Tasks}
\begin{plist}
\item \textbf{Question 1: The Justification Query}
  \begin{plist}
  \item \textbf{Keyword Requirement:} The question must include the word
    ``\pslot{question\_type\_word}''.
  \end{plist}
\item \textbf{Multi-Hop Logic:}
  \begin{plist}
  \item \textbf{Hop 1:} Information must originate from the Clip 1.
  \item \textbf{Hop 2:} Information must originate from the Clip 2.
  \end{plist}
\item \textbf{Anti-Generic Detail:} Use hyper-specific environmental cues that exist only in
  this specific sequence to ensure the video is the only key to the description.
\item \textbf{Detailed Scene Description:} Give detailed description of the scene surrounding
  the entity. Avoid very short and vague descriptions. Use at least 20 words per entity.
\item \textbf{Attribute Exclusion:} Strictly avoid describing what an entity looks like, wears,
  is doing, is acting or is made of. The viewer must deduce ``who'' or ``what'' based solely on
  \emph{where} they are and within the shot. For example, you MUST avoid the \emph{color},
  \emph{organs} or \emph{any direct physcial attribute}.
\end{plist}

\pfield{answer} \textit{Task: Comprehensive Video Answer Synthesis.}
\textbf{Objective:} Generate a detailed, long-form summary of the provided video content,
organized into distinct thematic paragraphs.
\begin{pnum}
\item \textit{Citation Formatting Constraints.} You MUST cite the video source for every claim
  made. Use the following strict format for timestamp ranges:
  \begin{plist}
  \item \textbf{Format:} \texttt{(video-id @ MM:SS - MM:SS)} --- the 1-indexed clip id, then
    \texttt{@}, then a \texttt{start - end} range in \texttt{MM:SS}.
  \item \textbf{Multiple ranges:} comma-separate them inside one bracket, repeating
    \texttt{video-id @} whenever the clip changes, e.g.\
    \texttt{(1 @ 04:20 - 05:15, 2 @ 10:02 - 13:32)}.
  \item \textbf{One clip per range:} each \texttt{start - end} range belongs to a SINGLE clip.
    NEVER mix two clips in one range (do NOT write \texttt{(1 @ 04:20 - 2 @ 05:15)}).
  \item \textbf{Placement:} Brackets must be placed only at the end of the sentence or section
    they support.
  \item \textbf{Example:} ``The speaker argues that renewable energy costs have plummeted
    significantly (1 @ 04:20 - 05:15).''
  \end{plist}
\item \textit{Structural Requirements.}
  \begin{plist}
  \item \textbf{Multi-Paragraph Format:} Do not use bullet points for the main body. Use 3-5
    distinct paragraphs to group related concepts.
  \item \textbf{Introduction:} Briefly state the primary topic and the speaker's core thesis.
  \item \textbf{Deep Dive:} Summarize the specific evidence, data, or narrative sequences
    presented in the footage.
  \item \textbf{Conclusion:} Summarize the final takeaways or calls to action provided at the
    end of the video.
  \end{plist}
\item \textit{Content Accuracy.} Summarize only the information retrieved from the video. Do not
  add outside information. Ensure the summary is cohesive and maintains the original context of
  the discussion.
\end{pnum}

\pfield{question\_type} Type of question as per the Knowledge Dimension Matrix
(\texttt{row}: Factual, Conceptual, Procedural or Meta Cognitive; \texttt{column}: Remember,
Understand, Apply, Analyze or Evaluate).

\pfield{bridge} Logical bridge between hops. \quad
\pfield{hop\_1} First hop. \quad
\pfield{hop\_2} Second hop.

\pfield{scene\_references} Scene references where entity has been encoded with the
scene-referred surrounding description in the QA pair with decoded identity. Each gives
\texttt{scene\_description} (the scence-referred hint or description used in the question) and
\texttt{actual\_entity\_name} (the real identity of the entity).

\pfield{clip\_ranges} Comprehensive list of clip ranges for covering all the references to the
question context and answer. Each gives \texttt{clip\_id} (1-indexed video index),
\texttt{start\_time} (absolute start time (adding start offset) in seconds where any reference
to the question context and/or answer begins), \texttt{end\_time} (absolute end time (adding
start offset) in seconds covering a particular reference to the question context and answer
from the start time) and \texttt{qa\_reference} (the reference to the question context and/or
answer).
\end{schemabox}

\subsubsection{Transcript-only answer (C2)}
\label{app:prompt-s03}

The transcript-shortcut check asks whether a question can be answered without the video.
\gflashgen{} answers from transcripts alone; \gptmini{} then compares that answer with the gold
answer using the verification prompt of Appendix~\ref{app:prompt-verify}. To make the relevant
passage harder to locate, each segment's own 30-minute transcript is placed among two further
30-minute windows from a different video of the playlist, in an order shuffled per item.

\begin{promptbox}{Transcript-only answer}
\pmeta{Block C2 \textperiodcentered{} \gflashgen{} \textperiodcentered{} input: text only}

\psec{Task: Comprehensive Answer Synthesis (Transcript-Grounded)}
\textbf{Objective:} Answer the given question using ONLY the provided transcript(s). Do not use
outside knowledge and do not infer visual details that are not stated in the transcript.
Organize the answer into distinct thematic paragraphs.

\psec{Structural Requirements}
\begin{plist}
\item \textbf{Multi-Paragraph Format:} Do not use bullet points for the main body. Use 3-5
  distinct paragraphs to group related concepts.
\item \textbf{Transcript Grounding:} Base every claim on the transcript text. If the transcript
  does not contain the information needed to answer, state that plainly rather than guessing.
\end{plist}

\psec{Output Fields (return JSON)}
\begin{plist}
\item \textbf{explanation:} FIRST, briefly reason about the question --- identify the two hops
  it spans and how the transcript connects them. This is your scratchpad; it is not part of the
  answer.
\item \textbf{answer:} THEN write the transcript-grounded answer as described above (3-5
  paragraphs). Do not restate the explanation or add any prelude/follow-up.
\end{plist}

\psec{Transcript}
\pslot{transcript}

\psec{Question}
\pslot{question}
\end{promptbox}

\begin{schemabox}{Transcript-only answer --- response schema}
\pmeta{Structured output \texttt{GeneratedAnswer}, enforced as a JSON schema; the explanation
  comes first so the model reasons before it answers}

\begin{promptcode}
GeneratedAnswer
  explanation : string
  answer      : string
\end{promptcode}

\pdivider{field descriptions}

\pfield{explanation} Reasoning: identify the two hops the question spans and how they connect,
before writing the answer. Not shown to the verifier.

\pfield{answer} The detailed, long-form answer: 3-5 distinct thematic paragraphs, no bullet
points.
\end{schemabox}

\subsubsection{Parametric answer (C2)}
\label{app:prompt-s05}

The parametric-recall check asks whether a question can be answered from memory, with neither
the video nor its transcript. It runs in two combinations with the generator and verifier
swapped: \gflashgen{} answers and \gptfull{} verifies (combination A), and \gptmini{} answers and
\gpro{} verifies (combination B). A question is rejected only when both combinations judge the
memory-only answer correct. Both generators receive the same prompt.

\begin{promptbox}{Parametric answer}
\pmeta{Block C2 \textperiodcentered{} combination A: \gflashgen{} \textperiodcentered{}
  combination B: \gptmini{} \textperiodcentered{} input: the question only}

\psec{Task: Comprehensive Answer Synthesis}
\textbf{Objective:} Generate a detailed, long-form answer to the given question, organized into
distinct thematic paragraphs, using your own parametric knowledge.

\psec{Structural Requirements}
\begin{plist}
\item \textbf{Multi-Paragraph Format:} Do not use bullet points for the main body. Use 3-5
  distinct paragraphs to group related concepts.
\item \textbf{Introduction:} Briefly state the primary topic and the core thesis.
\item \textbf{Deep Dive:} Summarize the specific evidence, data, or narrative.
\item \textbf{Conclusion:} Summarize the final takeaways or calls to action.
\end{plist}

\psec{Output Fields (return JSON)}
\begin{plist}
\item \textbf{explanation:} FIRST, briefly reason about the question --- identify the two hops
  it spans and how they connect. This is your scratchpad; it is not part of the answer.
\item \textbf{answer:} THEN write the answer as described above (3-5 paragraphs). Do not restate
  the explanation or add any prelude/follow-up.
\end{plist}

Question:\\
\pslot{question}
\end{promptbox}

\begin{schemabox}{Parametric answer --- response schema}
\pmeta{Structured output \texttt{GeneratedAnswer}, the same as for the transcript-only answer
  (Appendix~\ref{app:prompt-s03}); only its \texttt{answer} field reaches the verifier}

\begin{promptcode}
GeneratedAnswer
  explanation : string
  answer      : string
\end{promptcode}

\pdivider{field descriptions}

\pfield{explanation} Reasoning: identify the two hops the question spans and how they connect,
before writing the answer. Not shown to the verifier.

\pfield{answer} The detailed, long-form answer: 3-5 distinct thematic paragraphs, no bullet
points.
\end{schemabox}

\subsubsection{Answer verification (C2)}
\label{app:prompt-verify}

One verification prompt serves all three shortcut checks. It compares the answer produced
without the video, from the transcript (Appendix~\ref{app:prompt-s03}) or from memory
(Appendix~\ref{app:prompt-s05}), with the gold answer. A \emph{Yes} means the shortcut
succeeded. In the transcript check a single \emph{Yes} rejects the question; the parametric check
rejects it only when both combinations return \emph{Yes}.

\begin{promptbox}{Answer verification}
\pmeta{Block C2 \textperiodcentered{} transcript test: \gptmini{} \textperiodcentered{} parametric
  test: \gptfull{} (combination A), \gpro{} (combination B) \textperiodcentered{} input: text only}

You are an expert evaluator assessing the accuracy of an AI's predicted answer against a
ground-truth reference answer for a complex multi-hop question.

Both the Reference Answer and the Predicted Answer may be long-form text. Your task is to
extract the core factual claims and determine if the predicted text successfully resolves the
multi-hop logic without introducing fatal contradictions.

Evaluation Criteria:
\begin{pnum}
\item Deconstruct the Truth: Analyze the Reference Answer and identify the core factual
  conclusion necessary to answer the multi-hop question.
\item Scan the Prediction: Read through the long-form Predicted Answer to locate where (or if)
  it addresses those core facts.
\item Judge the Context:
  \begin{plist}
  \item Match (Yes): The predicted answer explicitly states the core factual truth found in the
    reference. Ignore extra verbosity, tangential information, or conversational filler,
    provided the core truth is present and unequivocally supported by the text.
  \item Mismatch (No): The predicted answer fails to include the core truth, completely misses
    one of the necessary logical ``hops'', or includes a direct contradiction that negates the
    correct information.
  \end{plist}
\end{pnum}

\psec{Input}
\textit{Question}\\ \pslot{question}

\textit{Reference Answer}\\ \pslot{reference\_answer}

\textit{Predicted Answer}\\ \pslot{predicted\_answer}
\end{promptbox}

\begin{schemabox}{Answer verification --- response schema}
\pmeta{Structured output \texttt{QueryJudgement}, enforced as a JSON schema; the explanation
  precedes the verdict}

\begin{promptcode}
QueryJudgement
  explanation : string
  judgement   : "Yes" | "No"
\end{promptcode}

\pdivider{field descriptions}

\pfield{explanation} Reasoning identifying the core facts in the reference, mapping them to the
long-form prediction, and noting any contradictions.

\pfield{judgement} Whether the prediction is factually consistent with the reference answer for
the question.
\end{schemabox}

\subsubsection{Video grounding (C3)}
\label{app:prompt-s09}

\qplus{} receives the prompt first and then, for each segment, the full 30-minute source video (resized
to fit $640\times480$, audio removed, sampled at 0.5\,fps) followed by its complete transcript,
labelled as shown at the end of the box. The question's own two-segment decomposition, written by the
generator in Appendix~\ref{app:prompt-s01}, is added as a checklist to verify rather than to
trust. An item is kept only when both verdicts are \emph{Yes}.

\begin{promptbox}{Video grounding}
\pmeta{Block C3 \textperiodcentered{} \qplus{} \textperiodcentered{} input: two full videos and
  their transcripts}

You are a video content analyst. You will receive video segments with their audio transcripts.

You are given a two-hop QUERY and its proposed ANSWER. Perform TWO assessments. An item is
accepted ONLY when BOTH are ``Yes'', so judge each carefully and independently.

\psec{Assessment 1 --- Is the QUERY answerable from the videos?
  $\rightarrow$ \texttt{query\_grounded}}
Determine if the following two-hop query can be answered using the information present in the
provided videos.\\
Rules for Two-Hop Evaluation:
\begin{plist}
\item Primary Goal (The Facts): Both distinct pieces of foundational information (the ``hops'')
  required to answer the query MUST be explicitly present in the videos or transcripts.
\item Secondary Goal (The Reasoning): If both hops are explicitly present, you may apply
  commonsense reasoning to connect them and draw the final conclusion. The final conclusion or
  relation itself does NOT need to be explicitly stated in the videos.
\item ``Yes'' ONLY if both foundational facts are explicitly present and the logical connection
  between them can be safely drawn.
\item ``No'' if either of the required foundational hops is missing, or if the connection
  requires specialized outside knowledge beyond basic commonsense.
\end{plist}

\psec{Assessment 2 --- Is the proposed ANSWER grounded in the videos?
  $\rightarrow$ \texttt{answer\_grounded}}
Determine whether the specific claims made in the proposed ANSWER are actually SUPPORTED BY
(grounded in) the video content and transcripts --- not merely plausible or answerable in
principle.
\begin{plist}
\item ``Yes'' ONLY if every substantive claim in the ANSWER is directly supported by what is
  shown on screen or said in the provided videos/transcripts.
\item ``No'' if the ANSWER asserts details that are absent, contradicted, hallucinated, or that
  rely on outside knowledge beyond the videos.
\end{plist}

For BOTH assessments, provide video references by their IDs in your explanations, indicating
which video contains which hop / supports which claim.

Query: \pslot{question}

Proposed Answer: \pslot{answer}

Respond in this exact JSON format:
\begin{promptcode}
{
    "query_grounding_explanation": "<1-2 sentence reasoning identifying where the two hops are found and how they connect>",
    "query_grounded": "<Yes or No>",
    "answer_grounding_explanation": "<1-2 sentence reasoning on whether the proposed answer's specific claims are supported by the videos/transcripts>",
    "answer_grounded": "<Yes or No>"
}
\end{promptcode}

\pdivider{appended: the generator's own decomposition}
INTENDED TWO-HOP DECOMPOSITION (authored with the question; the deliberately-obfuscated wording
above encodes exactly these claims). Your task is to VERIFY each part is actually supported by
the corresponding video AND its transcript --- confirm it, do not assume it:
\begin{plist}
\item Hop 1 --- should be grounded in Video 1 (+ Transcript 1): \pslot{hop\_1}
\item Hop 2 --- should be grounded in Video 2 (+ Transcript 2): \pslot{hop\_2}
\item Bridge --- the entity/reasoning linking hop 1 to hop 2: \pslot{bridge}
\end{plist}

\pdivider{appended: how the input is laid out}
You are given, for each hop, the FULL $\sim$30-minute source video FOLLOWED BY its full
spoken-audio transcript: Video 1 (hop 1) then Transcript 1, Video 2 (hop 2) then Transcript 2.
These are the WHOLE source videos (not pre-selected clips), so the relevant moment may be
anywhere within --- search the full transcript for the spoken facts and the video for the
visuals. Judge grounding from BOTH.

\pdivider{content parts that follow the prompt, once per segment $k \in \{1, 2\}$}
Video $k$ (hop $k$) --- the FULL $\sim$30-minute source video (\pslot{duration}s):
\quad\textit{[video]}\\
Transcript for Video $k$ (the full spoken audio, which the video frames alone do not convey):\\
\pslot{transcript}
\end{promptbox}

\begin{schemabox}{Video grounding --- response schema}
\pmeta{Structured output \texttt{VideoQueryJudgement}, enforced as a strict JSON schema; each
  verdict is preceded by its explanation so the model reasons first}

\begin{promptcode}
VideoQueryJudgement
  query_grounding_explanation  : string
  query_grounded               : "Yes" | "No"
  answer_grounding_explanation : string
  answer_grounded              : "Yes" | "No"
\end{promptcode}

\pdivider{field descriptions}

\pfield{query\_grounding\_explanation} 1-2 sentence reasoning identifying where the two hops are
found and how they connect.

\pfield{query\_grounded} Whether the two-hop QUERY is answerable from the provided
videos/transcripts.

\pfield{answer\_grounding\_explanation} 1-2 sentence reasoning on whether the proposed answer's
specific claims are supported by (grounded in) the provided video content and transcripts.

\pfield{answer\_grounded} Whether the proposed ANSWER's claims are actually grounded in the
videos/transcripts (not merely plausible or answerable in principle).
\end{schemabox}

\subsubsection{Graded wrong answers (D1)}
\label{app:prompt-s21}

With a question and its gold answer fixed, \gflashgen{} writes four degraded answers rated 4 to 1
while watching the same two clips, supplied as in Appendix~\ref{app:prompt-s01}. The seven
visual categories are listed in a fresh random order for every request, so no category is
anchored to a fixed position. The response schema again carries much of the specification: the
constraint every degraded answer must meet, and a severity clause that differs by rating.

\begin{promptbox}{Graded wrong answers --- prompt}
\pmeta{Block D1 \textperiodcentered{} \gflashgen{} \textperiodcentered{} input: two clips as
  video}

You are provided with two long video clips from a large database of multiple videos, a gold
standard \textbf{long-form response} rated 5 (perfectly accurate, highest quality, comprehensive
synthesis), and a corresponding \textbf{multi-hop question} for a \textbf{long-video
understanding task}. These video segments are part of a large video database on a specific
topic. This task requires detailed narrative generation, complex event synthesis, temporal
reasoning, or comprehensive summarization over extended video durations.

\psec{Citation Formatting Constraints}
The gold standard \textbf{long-form response} also contains citations from the videos. For
every claim made, the video source has been cited using the following strict format for
timestamp ranges:
\begin{plist}
\item \textbf{Format:} \texttt{(video-id @ MM:SS - MM:SS)} --- the 1-indexed clip id, then
  \texttt{@}, then a \texttt{start - end} range in \texttt{MM:SS}. Comma-separate multiple
  ranges inside one bracket when needed, repeating \texttt{video-id @} whenever the clip
  changes.
\item \textbf{Placement:} Brackets are placed only at the end of the sentence or section they
  support.
\item \textbf{Example:} ``The speaker argues that renewable energy costs have plummeted
  significantly (1 @ 04:20 - 05:15).''
\end{plist}

Your task is to generate four additional \textbf{long-form responses} that simulate
progressively lower-quality outputs for the same \textbf{multi-hop question}. Each generated
\textbf{response} should correspond to a quality rating from \textbf{4} to \textbf{1}, where
\textbf{Rating 5} is the provided gold standard and \textbf{Ratings 4 through 1} represent
decreasing quality.

As the rating decreases, the \textbf{responses} should reflect increasing levels of degradation
specific to challenges in long-video processing. These degradations should include
\textbf{temporal hallucinations (mixing up the timeline)}, \textbf{omission of entire key
segments}, \textbf{loss of narrative coherence}, and so on.

\textbf{Crucial Length Constraint:} All generated \textbf{responses} must remain similar in tone,
length and structural depth (e.g., multi-paragraph) to the gold standard. Do \textbf{not} simply
truncate the gold response to lower its quality---simulate \textbf{realistic, meaningful
degradation} and narrative drift while maintaining the long-form format. Use the provided
\textbf{videos} to ground the correctness of the response content.

\textbf{Hard Negative Constraint:} The degradations introduced in the lower-rated responses
(particularly Ratings 4 and 3) must act as \textbf{hard negatives}. Avoid obvious gibberish,
blatant self-contradictions, or sudden, jarring shifts to unrelated topics that make the errors
easily detectable. Instead:
\begin{plist}
\item Weave in plausible but subtly incorrect details, possibly grounded on the video.
\item Make realistic-sounding visual swaps (e.g., attributing an action to the wrong on-screen
  subject).
\item Maintain an authoritative, fluent, and confident tone while presenting flawed reasoning.
\end{plist}

The errors should require careful reading and deep comparison with the provided videos to spot,
forcing the evaluator to actively verify the logic and timeline rather than relying on
surface-level structural flaws.

\psec{Visual Modality Taxonomy}
Degradations should come from purely visual-only cues --- such as, but not limited to, the
following categories. You may also use other purely-visual details not listed here, as long as
they stay invisible to a transcript-only or world-knowledge reader:
\begin{plist}
\item COLOR/APPEARANCE: Object color, texture, material, clothing color
\item SPATIAL: Left/right positioning, foreground/background, proximity
\item GESTURE/POSTURE: Body language, hand position, eye contact direction
\item PROPS/OBJECTS: What object is held/used (when not named in audio)
\item ENVIRONMENT: Background setting details, room layout, visible signage
\item ON-SCREEN GRAPHICS: Slide content, diagrams, charts shown (if not read aloud)
\item CROWD/PRESENCE: Number of people visible, audience reactions
\end{plist}

DO NOT degrade using:
\begin{plist}
\item Speaker identity, names, or roles (text-verifiable)
\item Numerical claims, statistics (text-verifiable)
\item Event sequence or temporal order (text-verifiable)
\item Causal relationships between events (text-verifiable)
\item What is explicitly said or described verbally (text-verifiable)
\end{plist}

\psec{Output Format}
{\raggedright %
Return a valid JSON object matching the provided schema (\texttt{causal\_attributes},
\texttt{gold\_standard\_analysis}, and \texttt{rating\_4} \ldots{} \texttt{rating\_1}). Each
rating contains the degraded \textbf{long-form response} plus its causal-attribute analysis.
\textbf{EXCLUDE all timestamps and citations from the degraded answers.} Do not include any
commentary outside the JSON object.\par}

\psec{Input}
\textbf{multi-hop question:}\\ \pslot{question}

\textbf{Gold Standard Long-Form Response (Rating 5) containing timestamp references:}\\
\pslot{gold\_standard\_response}
\end{promptbox}

\begin{schemabox}{Graded wrong answers --- response schema}
\pmeta{Structured output \texttt{VideoResponseDegradation}, enforced as a JSON schema; its field
  descriptions, below the tree, are read by the model together with the prompt}

\begin{promptcode}
VideoResponseDegradation
  causal_attributes      : list of CausalAttribute
    attribute_name         : string
    importance_score       : number
    description            : string
  gold_standard_analysis : GoldStandardAnalysis
    attribute_analysis     : list of GoldAttributes
      attribute_name         : string
      quality_by_elements    : list of { element : string, impact : string }
  rating_4, rating_3, rating_2, rating_1 : DegradedResponse
    response               : string
    attribute_degradations : list of AttributeDegradations
      attribute_name          : string
      degradation_by_elements : list of { element : string, impact : string,
                                  modality : "visual_only" | "audio_visual" | "text_verifiable" }
    constant_attributes    : list of ConstantAttributes
      attribute_name          : string
      consistency_by_elements : list of { element : string, impact : string }
    upgraded_attributes    : list of UpgradedAttributes
      attribute_name          : string
      improvement_by_elements : list of { element : string, impact : string }
\end{promptcode}

\pdivider{field descriptions}

\pfield{causal\_attributes} As a reward model, rate answers for the given question across
multiple attributes. First identify these attributes and give an importance score between 0 and
1 for each, based on how important they are for rating a response to that question. The
importance scores should sum to 1.

Provide 5 \textbf{mutually exclusive} and important attributes required to rate an answer
holistically, along with their importance score. These attributes should be independent of each
other and depend largely on the given Question. Each gives \texttt{attribute\_name} (the name of
the holistic evaluation attribute), \texttt{importance\_score} (ranging from 0 to 1) and
\texttt{description} (what this attribute measures in the context of the question).

\pfield{gold\_standard\_analysis} Analysis of how the gold standard response (Rating 5)
satisfies the identified causal attributes. Try to mention all five
\texttt{causal\_attributes} in \texttt{GoldAttributes}. For each attribute, the specific causal
elements that make the gold standard response high quality, with the direct causal impact of
each on the attribute's high rating.

{\ttfamily\bfseries rating\_4, rating\_3, rating\_2, rating\_1}\enspace Each carries the same
constraint text, differing only in clause 3:

\hspace*{1em}\begin{minipage}{\dimexpr\linewidth-1em}
CRITICAL CONSTRAINTS:
\begin{pnum}
\item Structural Equivalence: Maintain the exact length, tone, and 2-hop QA structural depth of
  the gold standard.
\item Strict Exclusions: No timestamps, no meta-commentary, and strictly use positive phrasing
  only.
\item Visual-Dependent Degradation: \emph{(clause for this rating, below)}
\end{pnum}
Try to mention all \texttt{causal\_attributes} in AttributeDegradations, ConstantAttributes,
and UpgradedAttributes as a whole; the same attribute can be repeated across these three.
\end{minipage}

\begin{plist}
\item \textbf{Rating 4:} The text must read as perfectly logical and plausible; the error must
  be purely visual and undetectable relying solely on the transcript or world knowledge (minor
  visual alteration).
\item \textbf{Rating 3:} The text must NOT be disjointed or logically broken; it must read
  smoothly. The hallucination must rely entirely on inventing visual elements undetectable
  without video playback (moderate visual substitution).
\item \textbf{Rating 2:} Despite severe factual drift from the video, the text MUST remain
  cohesive and plausible. Do not ramble. The error must be purely visual and impossible to
  detect by reading the text or transcript (major visual event/subject change).
\item \textbf{Rating 1:} The text MUST NOT be incoherent, repetitive, or structurally broken. It
  must read flawlessly as a highly plausible long-form answer to a different video --- the total
  fabrication is 100\% visual and undetectable without watching the video (completely fabricated
  visual sequence).
\end{plist}

\psec{Each rating's fields}

\pfield{response}
\textit{Gold-Standard Response Constraints.} The gold standard response was generated with the
following constraints:
\begin{plist}
\item \textit{Structural Requirements}
  \begin{plist}
  \item \textbf{Multi-Paragraph Format:} Do not use bullet points for the main body. Use 3-5
    distinct paragraphs to group related concepts.
  \item \textbf{Introduction:} Briefly state the primary topic and the speaker's core thesis.
  \item \textbf{Deep Dive:} Summarize the specific evidence, data, or narrative sequences
    presented in the footage.
  \item \textbf{Conclusion:} Summarize the final takeaways or calls to action provided at the
    end of the video.
  \end{plist}
\item \textit{Content Accuracy.} Summarize only the information retrieved from the video. Do not
  add outside information. Ensure the summary is cohesive and maintains the original context of
  the discussion.
\end{plist}

Based on the provided Gold Standard Response and Video, generate a degraded response that
strictly adheres to the following constraints:
\begin{pnum}
\item Structural Equivalence: Maintain the exact length, tone, and structural depth of the gold
  standard.
\item 2-Hop QA Format: strictly follow the 2-hop question-answering structure used in the gold
  standard.
\item No Timestamps: Do not include or reference any video timestamps or citations.
\item No Meta-Commentary: Do not mention, hint, or imply that the response is fabricated,
  altered, or artificial.
\item Positive Phrasing Only: Strictly avoid Negative Polarity sentences (e.g., do not use
  phrasing like ``There is no real connection\ldots'', ``The video does not show\ldots'', or
  ``Unlike\ldots'').
\item Visual-Dependent Degradation (CRITICAL): The error or hallucination introduced must rely
  entirely on visual elements of the video. A reviewer reading only the text transcript/audio or
  relying on general parametric world knowledge must not be able to detect the degradation. It
  must read as perfectly logical and plausible unless compared directly against the actual video
  playback.
\end{pnum}

Generate a degraded response that passes the following adversarial test:

TEST: Give only the audio transcript of the video (no visuals) to a strong AI judge and ask it
to rate this response. The judge MUST rate this response as high quality (4-5/5) because the
degradation is invisible in the transcript or parametric knowledge.

To pass this test, degradations must ONLY target:
\begin{plist}
\item Visual appearance details (colors, clothing, object appearance)
\item Spatial/positional details (where things are placed on screen)
\item Gestural/body language details (what gesture accompanies speech)
\item Environmental/background details (what is visible in the scene)
\item Unnarrated on-screen graphics content
\end{plist}
Degradations MUST NOT target anything a transcript reveals:
\begin{plist}
\item What is said, claimed, or argued
\item The sequence or timing of verbal events
\item Names, titles, statistics, or quoted content
\item Causal logic stated in the narration
\end{plist}

SELF-CHECK before finalizing: Read your response alongside only the audio transcript. If a
fact-checker with only the transcript could flag your error, revise it. The error must survive
transcript-only verification as plausible.

\pfield{attribute\_degradations} As an expert in causal reasoning and response evaluation,
identify generalizable causal elements that directly affect the strength of each attribute
(CausalAttribute) in the degraded response.
\begin{plist}
\item Identify a list of causal elements that impact each attribute.
\item Each element must have a clear role in decreasing the attribute; explain its direct causal
  impact.
\item Do not include any non-causal heuristics.
\end{plist}
Each element gives \texttt{element}, \texttt{impact} and \texttt{modality}, which must be
\texttt{visual\_only} for valid degradation (the alternatives are \texttt{audio\_visual} and
\texttt{text\_verifiable}).

\pfield{constant\_attributes} Identify causal elements explaining how certain attributes
remained consistent with the gold standard.
\begin{plist}
\item Each element must have a clear role in maintaining stability.
\item Explain the direct causal impact on keeping the attribute stable.
\end{plist}

\pfield{upgraded\_attributes} Identify causal elements explaining how certain attributes were
improved compared to the gold standard (if any).
\begin{plist}
\item Each element must have a clear role in the enhancement.
\item Explain the direct causal impact on the improvement.
\end{plist}
\end{schemabox}

\subsubsection{Ranking without the video (D3)}
\label{app:prompt-s23}

This check asks whether the degradations can be spotted without watching. Two judges, \gpro{} and
\gptfull{}, each rank the gold answer together with its four degraded answers, shuffled, with
neither the video nor the transcript. Every candidate is formatted identically, so the gold
answer cannot be picked out by its text structure. The rubric is the set of attributes that
Appendix~\ref{app:prompt-s21} produced for this question. A set is rejected only when both
judges recover the intended order.

\begin{promptbox}{Ranking without the video}
\pmeta{Block D3 \textperiodcentered{} two judges: \gpro{} and \gptfull{} \textperiodcentered{}
  input: text only}

You are an expert evaluator. You are given a multi-hop question about long videos and several
long-form response candidates that each attempt to answer it.

You are \textbf{NOT} given the videos or their transcripts. Using ONLY your own reasoning about
internal consistency, plausibility, coherence, and general world knowledge, rank the candidates
from best (rank 1) to worst.

\psec{Multi-hop Question:}
\pslot{question}

\psec{Response Candidates:}
\pslot{candidates}

\psec{Evaluation Criteria:}
\begin{plist}
\item \textbf{Accuracy / Plausibility:} Which candidate reads as the most accurate, internally
  consistent account?
\item \textbf{Temporal Consistency:} Which maintains a coherent timeline of events?
\item \textbf{Hallucination:} Which weaves in implausible or self-contradictory details?
\item \textbf{Coherence:} Which flows most logically across paragraphs?
\end{plist}

\psec{Question-Specific Attributes (weigh the candidates on these):}
\pslot{rubric}

Rank ALL candidates from best to worst. Give particular problems in each response, not vague
differences. Return the full ranking plus a brief overall summary.

\pdivider{slot formats}
\pslot{candidates}, one block per candidate $i$:
\begin{promptcode}
### Response Candidate i:

<candidate_response>
...
</candidate_response>
\end{promptcode}
\pslot{rubric}, one line per attribute:
\begin{promptcode}
- **<attribute_name>** (importance <importance_score>): <description>
\end{promptcode}
\end{promptbox}

\begin{schemabox}{Ranking without the video --- response schema}
\pmeta{Structured output \texttt{WrongAnswerRanking}, enforced as a JSON schema; the gate compares
  the returned order with the intended one}

\begin{promptcode}
WrongAnswerRanking
  rankings        : list of RankEntry
    rank            : integer
    candidate_index : integer
    reasoning       : string
  overall_summary : string
\end{promptcode}

\pdivider{field descriptions}

\pfield{rankings} A full ranking of ALL candidates, from best (rank 1) to worst.
\begin{plist}
\item \texttt{rank}: 1 = best. Increasing rank = lower quality.
\item \texttt{candidate\_index}: The 1-indexed candidate placed at this rank.
\item \texttt{reasoning}: Specific problems or strengths that justify this placement. Name
  particular issues in the response; do not give vague differences.
\end{plist}

\pfield{overall\_summary} A brief summary of the key factors that discriminated the candidates.
\end{schemabox}

\subsubsection{Ranking with the video (D4)}
\label{app:prompt-s25}

The complementary check asks whether the degradations are visible once the video is available.
\qplus{} receives the prompt followed by the two full videos and transcripts, laid out as in
Appendix~\ref{app:prompt-s09}, and ranks the four degraded answers. The gold answer is left
out here, so each candidate can carry the degradation analysis that
Appendix~\ref{app:prompt-s21} wrote for it without making any one of them stand out. A set
passes only if the judge reproduces the intended order exactly.

\begin{promptbox}{Ranking with the video}
\pmeta{Block D4 \textperiodcentered{} \qplus{} \textperiodcentered{} input: two full videos and
  their transcripts}

You are an expert video evaluator. You are given video clips (as sampled frames) together with
their audio transcripts, a multi-hop question, and several long-form response candidates that
each attempt to answer it.

Your task is to rank these response candidates from best (rank 1) to worst based on their
accuracy, temporal consistency, and alignment with the actual VIDEO evidence --- not merely on
surface plausibility. Ground every judgement in what the clips actually show and say.

\psec{Multi-hop Question:}
\pslot{question}

\psec{Response Candidates:}
\pslot{candidates}

\psec{Question-Specific Attributes (weigh the candidates on these):}
\pslot{rubric}

\psec{Evaluation Criteria:}
\begin{plist}
\item \textbf{Accuracy:} Does the response correctly identify events and details actually shown
  in the video?
\item \textbf{Temporal Consistency:} Does it maintain the correct timeline of events as they
  appear in the clips?
\item \textbf{Hallucination:} Does it invent visual details or attribute actions to the wrong
  on-screen subject/phase?
\item \textbf{Coherence:} Does the narrative flow logically across paragraphs?
\end{plist}

Each candidate is accompanied by its intended degradation analysis for internal verification.
Verify those degradations against the video, then rank ALL candidates from best to worst. Give
particular problems grounded in the video, not vague differences. Return the full ranking plus a
brief overall summary.

\pdivider{slot format}
\pslot{candidates}, one block per candidate $i$; \pslot{rubric} as in
Appendix~\ref{app:prompt-s23}:
\begin{promptcode}
### Response Candidate i:

<candidate_response>
...
</candidate_response>

#### Intended Degradation Analysis (for internal verification):
<rubrics>
[
  {
    "attribute_name": "...",
    "degradation_by_elements": [
      {
        "element": "...",
        "impact": "...",
        "modality": "visual_only"
      }
    ]
  }
]
</rubrics>
\end{promptcode}
\end{promptbox}

\begin{schemabox}{Ranking with the video --- response schema}
\pmeta{Structured output \texttt{WrongAnswerRanking}, enforced as a strict JSON schema; the same
  schema and field descriptions as for the ranking without the video
  (Appendix~\ref{app:prompt-s23})}

\begin{promptcode}
WrongAnswerRanking
  rankings        : list of RankEntry
    rank            : integer
    candidate_index : integer
    reasoning       : string
  overall_summary : string
\end{promptcode}
\end{schemabox}

\subsubsection{Regeneration feedback (E)}
\label{app:prompt-feedback}

When an item fails a gate, the reason is fed back and the item is generated again, up to the
attempt limit. The block below is appended to the question-generation prompt
(Appendix~\ref{app:prompt-s01}) or, if only the wrong answers were rejected, to the
wrong-answer prompt (Appendix~\ref{app:prompt-s21}), listing every earlier failed attempt for
that item. \pslot{reason} is replaced by the matching guidance for the gate that rejected it.

\begin{promptbox}{Regeneration feedback --- question and answer}
\pmeta{Block E \textperiodcentered{} appended to the question-generation prompt (B) on retry}

\psec{Feedback --- regenerate a BETTER two-hop QA}
Your previous attempt(s) were REJECTED. Do NOT repeat these mistakes:
\begin{plist}
\item Previous question: \pslot{question}\\
  Previous answer: \pslot{answer}\\
  Rejected because: \pslot{reason}
\end{plist}
Generate a NEW question+answer that REQUIRES watching the video: not answerable from the
transcript/audio alone, nor from world knowledge, with both hops explicitly grounded in the
clips.

\pdivider{\pslot{reason}, by the gate that rejected the item}
\begin{plist}
\item \textbf{Structural checks (C1):} the answer had a structural/format issue --- use 3-5
  paragraphs with proper (id @ MM:SS - MM:SS) citations covering at least two 30s segments per
  hop.
\item \textbf{Transcript test (C2):} it could be answered from the transcript/audio ALONE (no
  video needed) --- make the answer depend on VISUAL details that appear only in the video.
\item \textbf{Parametric test (C2):} it could be answered from general world knowledge ---
  make it depend on specific content unique to THESE clips, not common knowledge.
\item \textbf{Video grounding (C3):} the answer was not actually shown in the video --- ensure
  BOTH hops are explicitly present in the clips.
\end{plist}
\end{promptbox}

\begin{promptbox}{Regeneration feedback --- wrong answers}
\pmeta{Block E \textperiodcentered{} appended to the wrong-answer prompt (D1) on retry; the
  question and gold answer stay fixed}

\psec{Feedback --- regenerate BETTER distractors (wrong answers)}
Your previous degraded-answer set(s) were REJECTED. Do NOT repeat these mistakes:
\begin{plist}
\item Question: \pslot{question}\\
  Rejected because: \pslot{reason}
\end{plist}
Produce a NEW set of four degraded answers (Rating 4$\rightarrow$1) with VISUAL-ONLY
degradations that survive a text-only check but are detectable WITH the video.

\pdivider{\pslot{reason}, by the gate that rejected the set}
\begin{plist}
\item \textbf{Structural checks (D2):} some degraded answers were malformed --- every rating
  (4$\rightarrow$1) must carry a non-empty \texttt{attribute\_degradations} analysis and a proper
  \texttt{response}.
\item \textbf{Parametric detectability (D3):} the degradations were too obvious: BOTH text-only
  judges ranked them correctly WITHOUT the video. Degrade ONLY visual details (colour, spatial
  layout, gesture, on-screen objects/text) that a reader cannot infer from the question or the
  gold answer wording, so the wrong answers are indistinguishable from text alone.
\end{plist}
\end{promptbox}

\subsection{Human evaluation evidence prompt}
\label{app:prompt-ground}

An annotator sees short evidence clips, not the two full segments. To choose them, \qplus{} reads
both segments of a question in full --- each downscaled to 480p with the audio track removed and
encoded at 2\,fps --- together with their complete transcripts, and returns time ranges for two
things: the aspect the question asks about in each segment, and every sentence of the reference
answer. Those ranges become the 30-second chunks in the worker's video panel
(Figure~\ref{fig:mturk-task}). The transcript handed to the model is relabelled to its segment's own
clock, so the \texttt{[MM:SS-MM:SS]} labels on its lines and the timestamps we ask for share one
time base. The pass reads the video itself rather than reusing the timestamps that the generator
recorded when it wrote the question.

\begin{promptbox}{Evidence selection for human evaluation}
\pmeta{Human study \textperiodcentered{} \qplus{} \textperiodcentered{} input: both segment videos in
  full and their transcripts}

You are given, for each of the two hops of a two-hop question, the FULL source video (each label
states its exact length) followed by its full spoken-audio transcript: Video 1 (hop 1) then
Transcript 1, Video 2 (hop 2) then Transcript 2. These are the WHOLE source videos --- the
relevant moment may be anywhere within.

You will ground TWO things against the videos. For each, output the moment(s) it occurs as time
ranges --- \texttt{start\_time} and \texttt{end\_time} in MM:SS (minutes:seconds, relative to the
start of the relevant hop's video; each video is under an hour, so use minutes:seconds only,
e.g.\ 05:15 is 5 min 15 s and 23:40 is 23 min 40 s) plus a short \texttt{event} describing what
actually happens on screen / is said there. The transcript lines are labelled
\texttt{[MM:SS-MM:SS]} in the SAME clock, matching the video's own time.

(A) the two QUESTION HOPS --- the specific aspect the question asks about in each hop. Hop 1 is
grounded in Video 1, Hop 2 in Video 2 (each hop's aspect lives in its own video).

(B) the REFERENCE ANSWER, sentence by sentence --- for each sentence give the hop (1 or 2) that
supports it and the range(s); a sentence may span one hop or both, one range or a few. If a
sentence is a general summary not tied to any specific moment, return an empty ranges list for
it.

Use BOTH modalities to locate every timestamp --- do not rely on the transcript text alone:
\begin{plist}
\item IMAGE (video frames): confirm what is actually SEEN on screen at that moment (objects,
  people, actions, on-screen text, scene) matches.
\item TEXT (transcript): confirm what is SAID at that moment matches.
\end{plist}
A time range is valid only when the visual evidence AND/OR the spoken text at that moment
genuinely support it; cross-check the frames against the transcript and prefer moments where they
agree. Use the smallest ranges that cover the evidence, and make \texttt{event} a concrete
description of that moment.

Question:\\
\pslot{question}

Question hops (ground each in ITS video --- Hop 1 in Video 1, Hop 2 in Video 2):
\begin{plist}
\item Hop 1: \pslot{hop\_1}
\item Hop 2: \pslot{hop\_2}
\end{plist}

Reference answer sentences (ground each in whichever hop supports it):\\
\pslot{numbered}

Respond with a SINGLE valid JSON object and NOTHING else (no prose, no markdown, no code fences),
of this EXACT shape:
\begin{promptcode}
{"hop_groundings": [{"hop": 1, "ranges": [{"start_time": "MM:SS", "end_time": "MM:SS", "event": "<what happens there>"}]}, {"hop": 2, "ranges": [{"start_time": "MM:SS", "end_time": "MM:SS", "event": "<what happens there>"}]}], "groundings": [{"sentence": <int>, "ranges": [{"hop": 1, "start_time": "MM:SS", "end_time": "MM:SS", "event": "<what happens there>"}]}]}
\end{promptcode}

\pdivider{content parts that follow the prompt, once per segment $k \in \{1, 2\}$}
Video $k$ (hop $k$) --- the FULL source video, $\sim$\pslot{minutes} min
(\pslot{duration}s):\\
\quad\textit{[video]}\\
Transcript for Video $k$:\\
\pslot{transcript}
\end{promptbox}

\begin{schemabox}{Evidence selection --- response schema}
\pmeta{Structured output \texttt{grounding}, enforced as a strict JSON schema. It fixes the
  types and the MM:SS pattern only, so the meaning of each field is carried by the prompt above}

\begin{promptcode}
grounding
  hop_groundings : list of        one entry per question hop
    hop            : integer      1 or 2
    ranges         : list of
      start_time     : string     ^\d{1,2}:\d{2}$
      end_time       : string     ^\d{1,2}:\d{2}$
      event          : string
  groundings     : list of        one entry per reference-answer sentence
    sentence       : integer
    ranges         : list of
      hop            : integer    1 or 2
      start_time     : string     ^\d{1,2}:\d{2}$
      end_time       : string     ^\d{1,2}:\d{2}$
      event          : string
\end{promptcode}

\pdivider{field descriptions}

\pfield{hop\_groundings} The two QUESTION HOPS: the specific aspect the question asks about in
each hop, with the moment(s) it occurs. Hop~1 is grounded in Video~1 and hop~2 in Video~2, so a
range here needs no hop of its own.

\pfield{groundings} The REFERENCE ANSWER, sentence by sentence, by the sentence's position in the
numbered list the prompt shows, counting from zero. A sentence may be supported in one hop or
both, by one range or a few; a general summary tied to no specific moment returns an empty
\texttt{ranges} list.

\pfield{hop} Which hop's video the range lies in, for a range that supports an answer sentence.

\pfield{start\_time, end\_time} The range, in MM:SS on the clock of that hop's video --- the same
clock the transcript lines are labelled in. The smallest range that covers the evidence.

\pfield{event} A concrete description of what happens on screen or is said in that range.
\end{schemabox}

\end{document}